\documentclass[10pt,twocolumn,letterpaper]{article}

\usepackage[pagenumbers]{cvpr} 

\newcommand{\PENDING}[1]{\textcolor{orange}{[pending: #1]}}

\usepackage{xspace}
\newcommand{\paper}{FAB\xspace}

\usepackage{graphicx}
\usepackage{booktabs}
\usepackage{placeins}
\usepackage{multirow}
\usepackage{tikz}
\usepackage{colortbl}
\newsavebox{\tabnotesbox}
\newenvironment{tabwrap}{\lrbox{\tabnotesbox}}{\endlrbox}
\newcommand{\tabnotes}[1]{\begin{minipage}{\wd\tabnotesbox}\usebox{\tabnotesbox}\par\vspace{2pt}{\footnotesize\raggedright #1\par}\end{minipage}}
\definecolor{oursbg}{RGB}{232,244,252} 
\usetikzlibrary{arrows.meta,fit,backgrounds,positioning,calc}

\newcommand{\tabfont}{\footnotesize}

\usepackage{iftex}
\ifXeTeX
  \usepackage{fontspec}
\fi
\newif\ifdense
\densetrue
\newif\ifablready
\ablreadytrue
\newif\ifdenseshown
\ifdense\denseshowntrue\fi
\ifablready\denseshowntrue\fi

\definecolor{cvprblue}{rgb}{0.21,0.49,0.74}
\usepackage[pagebackref,breaklinks,colorlinks,allcolors=cvprblue]{hyperref}

\def\paperID{*****} 
\def\confName{arXiv}
\def\confYear{2026}

\title{Fast and Accurate Body Estimation in Pixels, World Space and Scenes}

\author{Jinkun Cao\thanks{This paper is an independent personal project and not related to Jinkun Cao's position at Amazon FAR.}\\
{\tt\small jinkuncao@gmail.com}
}

\begin{document}
\makeatletter\let\fab@maketitle\@maketitle\def \@maketitle{\fab@maketitle\global\let\fab@thanks\@thanks}\makeatother
\twocolumn[{%
\renewcommand\twocolumn[1][]{#1}%
\maketitle
\vspace{-1.2em}
\begin{center}
\makeatletter\def\@captype{figure}\makeatother
\IfFileExists{figures/teaser.pdf}{\includegraphics[width=\textwidth]{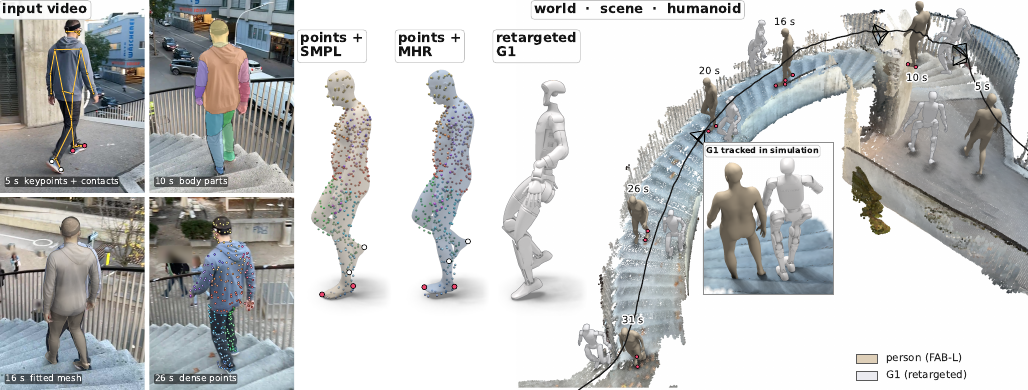}}{%
\fbox{\parbox[c][5.6cm][c]{0.96\textwidth}{\centering\PENDING{teaser}}}}
\caption{\paper from one handheld video (EMDB~\cite{kaufmann2023emdb} P3\_30). Left: four per-frame outputs: 2D keypoints with heel / toe contacts (red: planted heel or toe; white: off), body parts, the SMPL body fitted to its points and the dense 3D points. Middle, the dense points with the fitted SMPL body, the same points with the MHR mesh, and the retargeted Unitree G1 reference. Right: the reconstructed stairs in one metric, gravity-aligned world, with the camera path, the person and the G1 reference at six instants (red dots on the feet: predicted contacts). Inset: the G1 tracked in MuJoCo by a terrain-aware waypoint policy on our stair mesh, beside the person at the same instant.}
\label{fig:teaser}
\end{center}

\vspace{0.8em}
}]
\makeatletter{\def\thefootnote{\fnsymbol{footnote}}\fab@thanks}\makeatother
\ifdense\def\figtworet{the motion retargeted to a Unitree G1 from the dense points (red frames)}\else\def\figtworet{retargeting to a Unitree G1}\fi
\begin{figure*}[t]
\centering
\resizebox{\linewidth}{!}{%
\ifdense\input{sec/m_overview_v3f_dense}\else
\begin{tikzpicture}[x=1mm, y=1mm,
  hd/.style={font=\sffamily\bfseries\footnotesize, anchor=south west, inner sep=0pt},
  lb/.style={font=\sffamily\scriptsize, anchor=base, inner sep=0pt, text=black!80},
  sub/.style={font=\sffamily\scriptsize, anchor=north west, inner sep=0pt, text=black!65},
  arr/.style={-{Stealth[length=2mm, width=1.6mm]}, line width=0.7pt, draw=black!70},
  rail/.style={line width=0.7pt, draw=black!70},
  clb/.style={font=\sffamily\scriptsize\bfseries, text=ctoe!85!black, inner sep=0pt},
  hlb/.style={font=\sffamily\scriptsize\bfseries, text=cheel!85!black, inner sep=0pt, align=left},
  img/.style={inner sep=0pt, anchor=south west},
]
\definecolor{ctoe}{RGB}{239,71,111}
\definecolor{cheel}{RGB}{17,138,178}
\def\H{36}
\def\ks{0.947368}
\def\T{\H+2.2}
\node[img, draw=black!15, line width=0.3pt] at (3.4,7.2) {\includegraphics[height=28.4mm]{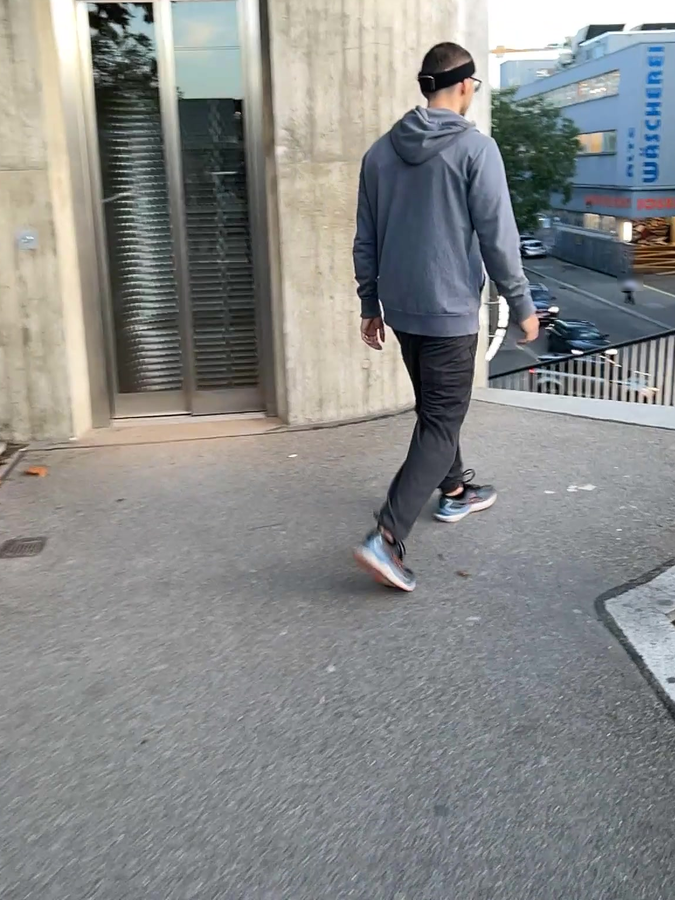}};
\node[img, draw=black!15, line width=0.3pt] at (1.7,5.5) {\includegraphics[height=28.4mm]{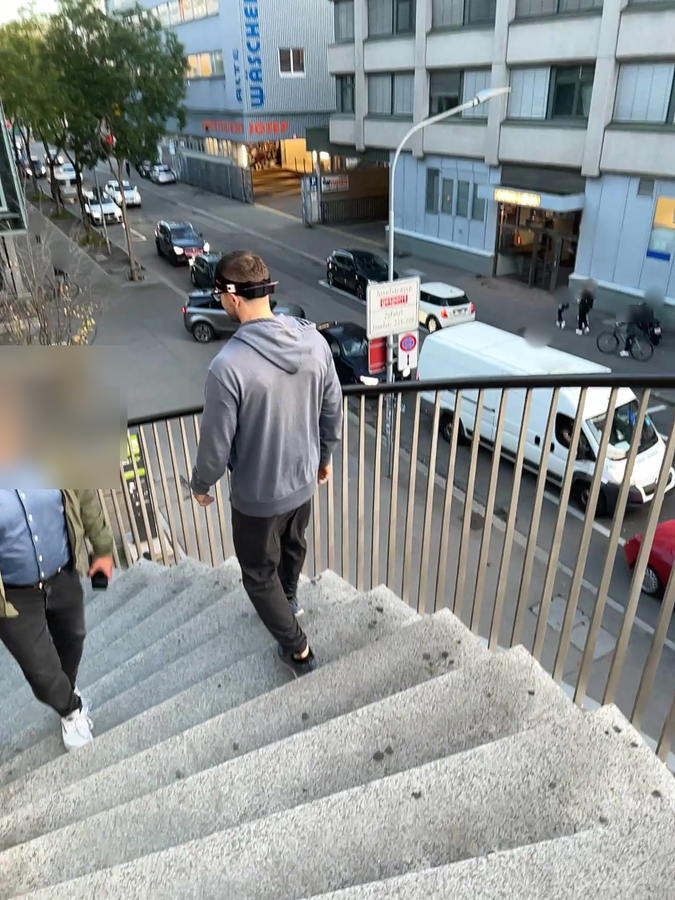}};
\node[img, draw=black!15, line width=0.3pt] (in) at (0,3.8) {\includegraphics[height=28.4mm]{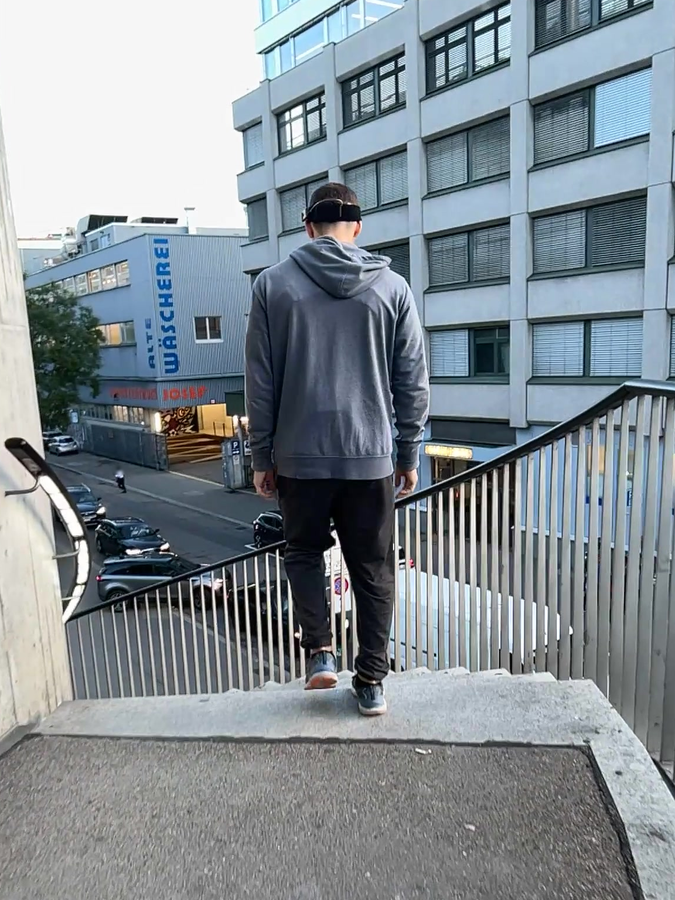}};
\node[hd, anchor=base west] at (0,\T) {Input video};
\draw[arr] (25.6,\H/2) -- (30.4,\H/2);
\filldraw[draw=black!45, fill=black!5, line width=0.5pt, rounded corners=1pt]
  (32,4) -- (32,\H-4) -- (44,\H-10) -- (44,10) -- cycle;
\node[font=\sffamily\scriptsize, align=center, text=black!80] at (38,\H/2) {ViT\\[-1pt]{\tiny DINOv3}\\[-2pt]{\tiny distilled}};
\filldraw[draw=orange!70!black, fill=orange!12, line width=0.5pt, rounded corners=2pt] (46,10) rectangle (55,\H-10);
\node[font=\sffamily\scriptsize, align=center, text=black!80, rotate=90] at (50.5,\H/2) {decoder};
\draw[rail] (44,\H/2) -- (46,\H/2);
\node[img] (kp) at (60,0)  {\includegraphics[height=\H mm]{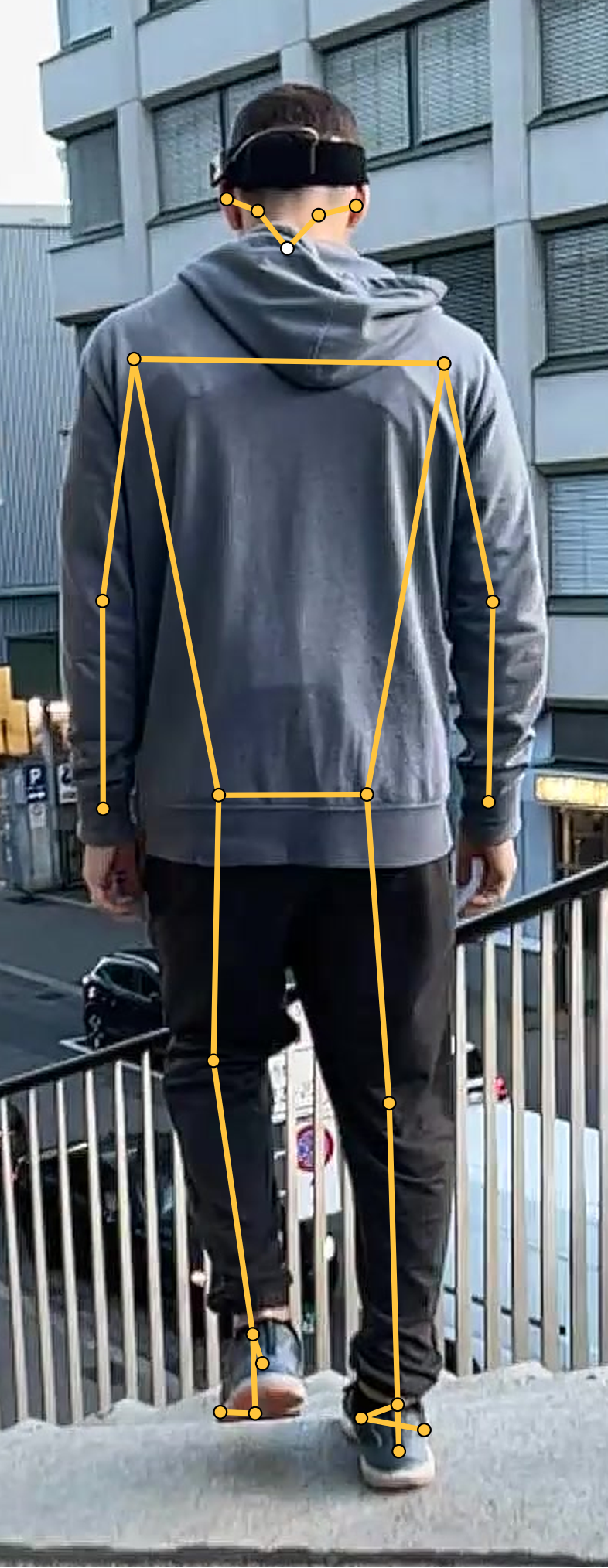}};
\node[img] (pt) at (80,0)  {\includegraphics[height=\H mm]{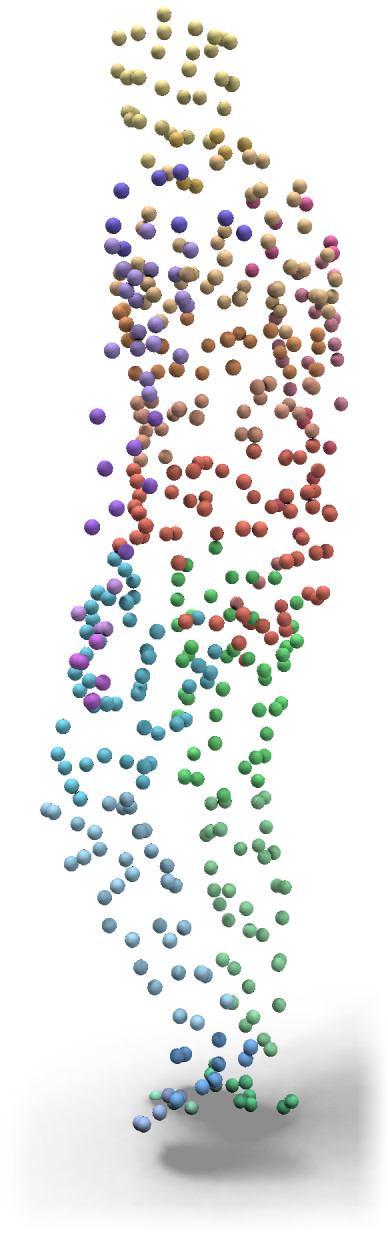}};
\node[img] (sm) at (96,0) {\includegraphics[height=\H mm]{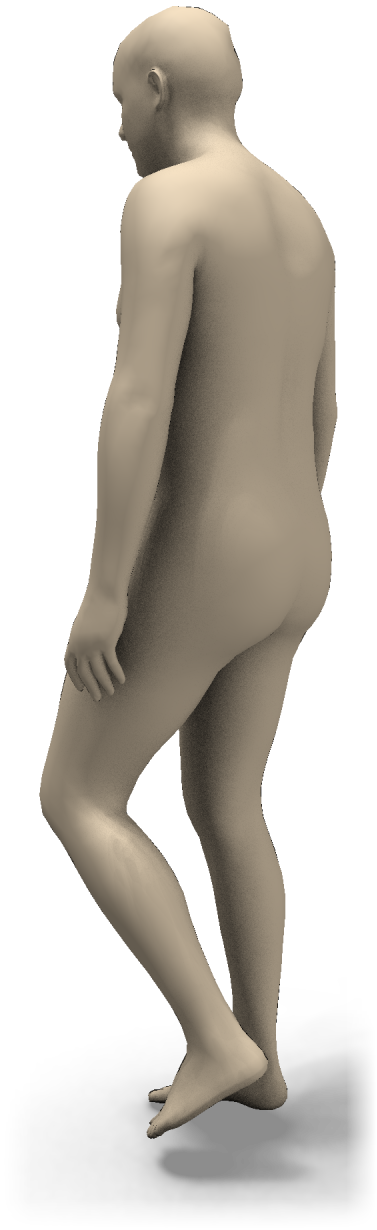}};
\node[lb] at (kp.north |- 0,\T) {2D keypoints};
\node[lb] at (pt.north |- 0,\T) {3D points};
\node[lb] at (sm.north |- 0,\T) {fitted mesh};
\draw[arr] ([xshift=0.5mm]pt.east |- 0,\H/2) -- ([xshift=-0.5mm]sm.west |- 0,\H/2) node[midway, above, font=\sffamily\scriptsize, text=black!70] {fit};
\def\cx{112}
\fill[ctoe!5, rounded corners=2.5pt] (\cx-1.5,-1) rectangle (\cx+29.5*\ks,\H+1);
\node[img] (ct) at (\cx,0) {\includegraphics[height=\H mm]{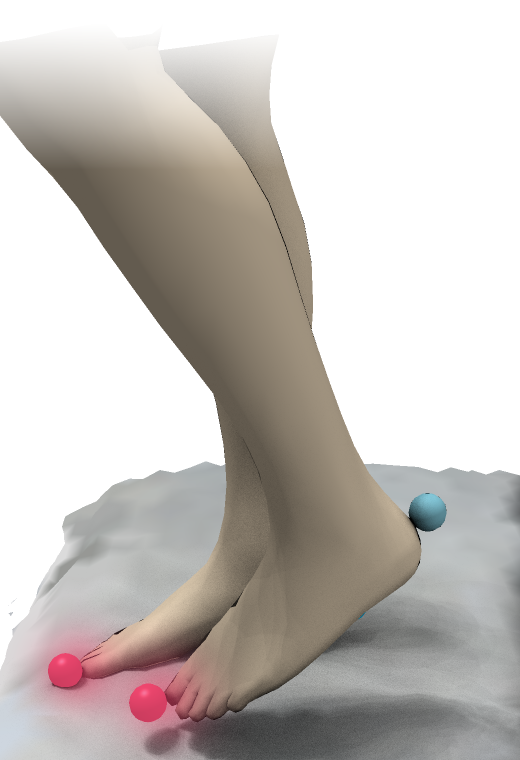}};
\draw[ctoe, line width=1.3pt, rounded corners=2.5pt] (\cx-1.5,-1) rectangle (\cx+29.5*\ks,\H+1);
\filldraw[fill=white, draw=cheel, line width=0.7pt] (\cx+21.45*\ks,12.35*\ks) circle (1.05mm);
\draw[ctoe!85!black, line width=0.4pt] (\cx+8.9*\ks,2.7*\ks) -- (\cx+14.0*\ks,1.3*\ks) node[clb, anchor=west] {toe on};
\draw[cheel!85!black, line width=0.4pt] (\cx+22.1*\ks,11.4*\ks) -- (\cx+22.8*\ks,9.4*\ks) node[hlb, anchor=north west] {heel\\raised};
\node[font=\sffamily\tiny, text=black!70, anchor=north west, align=left, inner sep=0pt] at (\cx+15.6,\H-1.4)
  {\textcolor{ctoe}{$\bullet$} toe\enspace\textcolor{cheel}{$\bullet$} heel\\$\circ$ off\\[1pt]toe on,\\heel raised};
\node[font=\sffamily\scriptsize\bfseries, anchor=base west, inner sep=0pt, text=ctoe!85!black] at (\cx+1,\T) {heel / toe contacts};
\draw[-{Stealth[length=2mm, width=1.6mm]}, line width=0.9pt, draw=ctoe] (\cx+29.9*\ks,16) -- (\cx+33.7,16);
\def\gi{146}
\def\gy{6}
\fill[white] (\gi-0.2,\gy-0.2) rectangle (\gi+18.62,\gy+20.2);
\node[img] at (\gi,\gy) {\includegraphics[height=20mm]{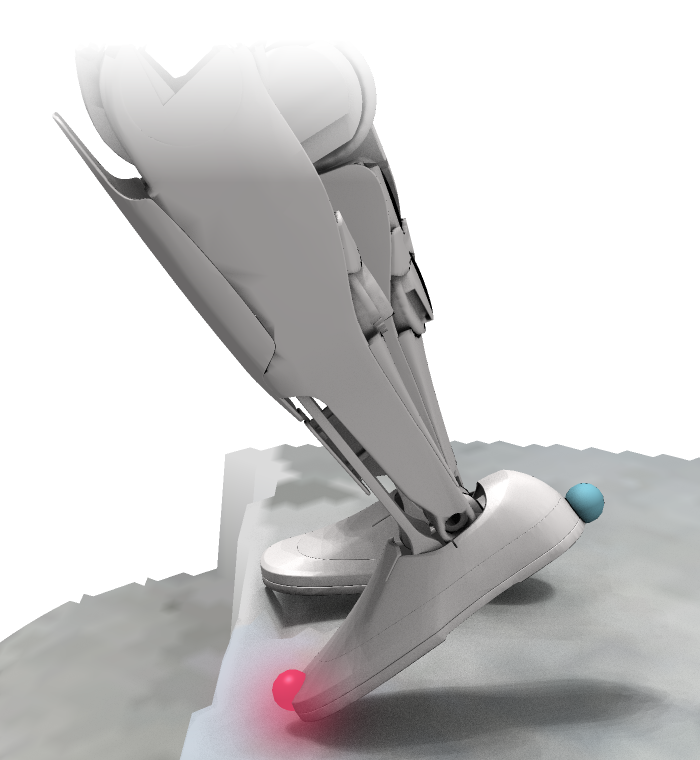}};
\draw[ctoe, line width=0.9pt, rounded corners=1.5pt] (\gi-0.2,\gy-0.2) rectangle (\gi+18.62,\gy+20.2);
\filldraw[fill=white, draw=cheel, line width=0.7pt] (\gi+15.37,\gy+6.29) circle (0.7mm);
\draw[ctoe!85!black, line width=0.4pt] (\gi+7.71,\gy+1.0) -- (\gi+7.71,\gy-0.9) node[clb, anchor=north] {toe on};
\draw[cheel!85!black, line width=0.4pt] (\gi+15.37,\gy+7.0) -- (\gi+15.37,\gy+10.0) node[hlb, anchor=south, align=center] {heel\\raised};
\node[font=\sffamily\scriptsize\bfseries, text=ctoe!85!black, anchor=south, inner sep=0pt, align=center] at (\gi+9.3,\gy+21) {retargeted\\G1 foot};
\node[img] (g1) at (170,0) {\includegraphics[height=34mm]{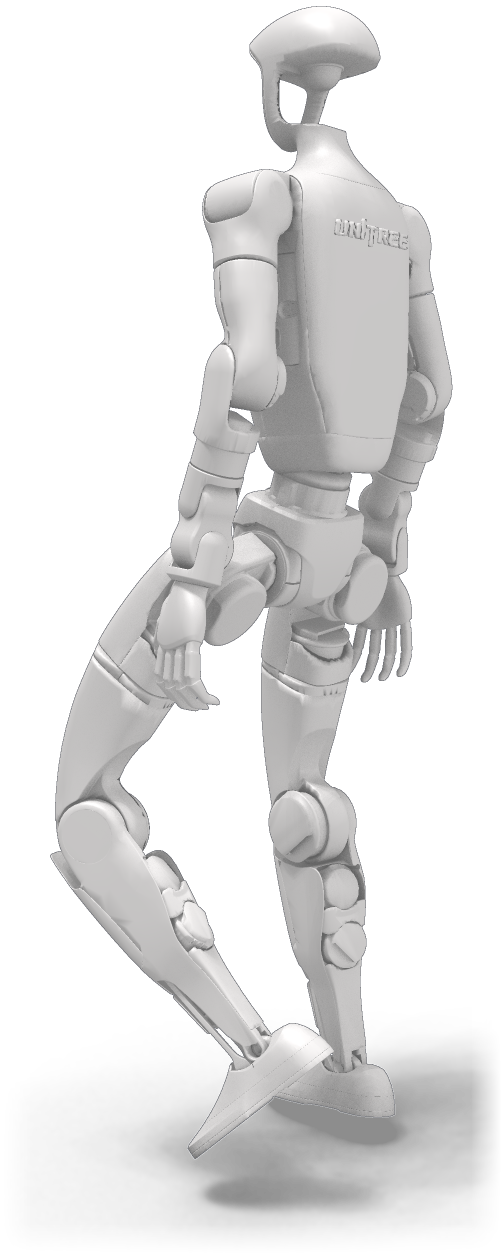}};
\node[lb] at (g1.north |- 0,\T) {G1 joint angles};
\def\R{\H+7.4}
\draw[rail] (55,\H/2) -- (57,\H/2) -- (57,\R) -- (g1.north |- 0,\R);
\foreach \n in {kp,pt} {\draw[arr] (\n.north |- 0,\R) -- ++(0,-2.6);}
\draw[arr] (g1.north |- 0,\R) -- ++(0,-2.6);
\draw[arr] (\cx-0.2,\R) -- (\cx-0.2,\H+1.2);
\def\h{21}
\def\Y{-32.6}
\def\W{54}
\ifdense\def\humsub{retargeting from dense points (or mesh) $\cdot$ G1 in sim}\else\def\humsub{retargeting $\cdot$ G1 tracked in sim}\fi
\fill[black!3, rounded corners=3pt] (-2,\Y-4.0) rectangle (190,\Y+\h+6.0);
\draw[black!12, line width=0.4pt, rounded corners=3pt] (-2,\Y-4.0) rectangle (190,\Y+\h+6.0);
\draw[arr] (93,-1.4) -- (93,\Y+\h+6.2);
\foreach \x/\c/\t/\s in {0/green!45!black/World/{cameras $\cdot$ metric depth $\cdot$ gravity},
                        66.5/orange!80!black/Scenes/{fusion $\cdot$ carving $\cdot$ human--scene opt.},
                        133/red!65!black/Humanoid/{\humsub}} {
  \node[hd, text=\c] at (\x,\Y+\h+2.2) {\t};
  \draw[\c, line width=0.8pt] (\x,\Y+\h+1.2) -- ++(\W,0);
  \node[sub] at (\x,\Y-0.7) {\s};
}
\node[inner sep=0pt] at (0+\W/2,\Y+\h/2) {\includegraphics[width=\W mm, height=\h mm, keepaspectratio]{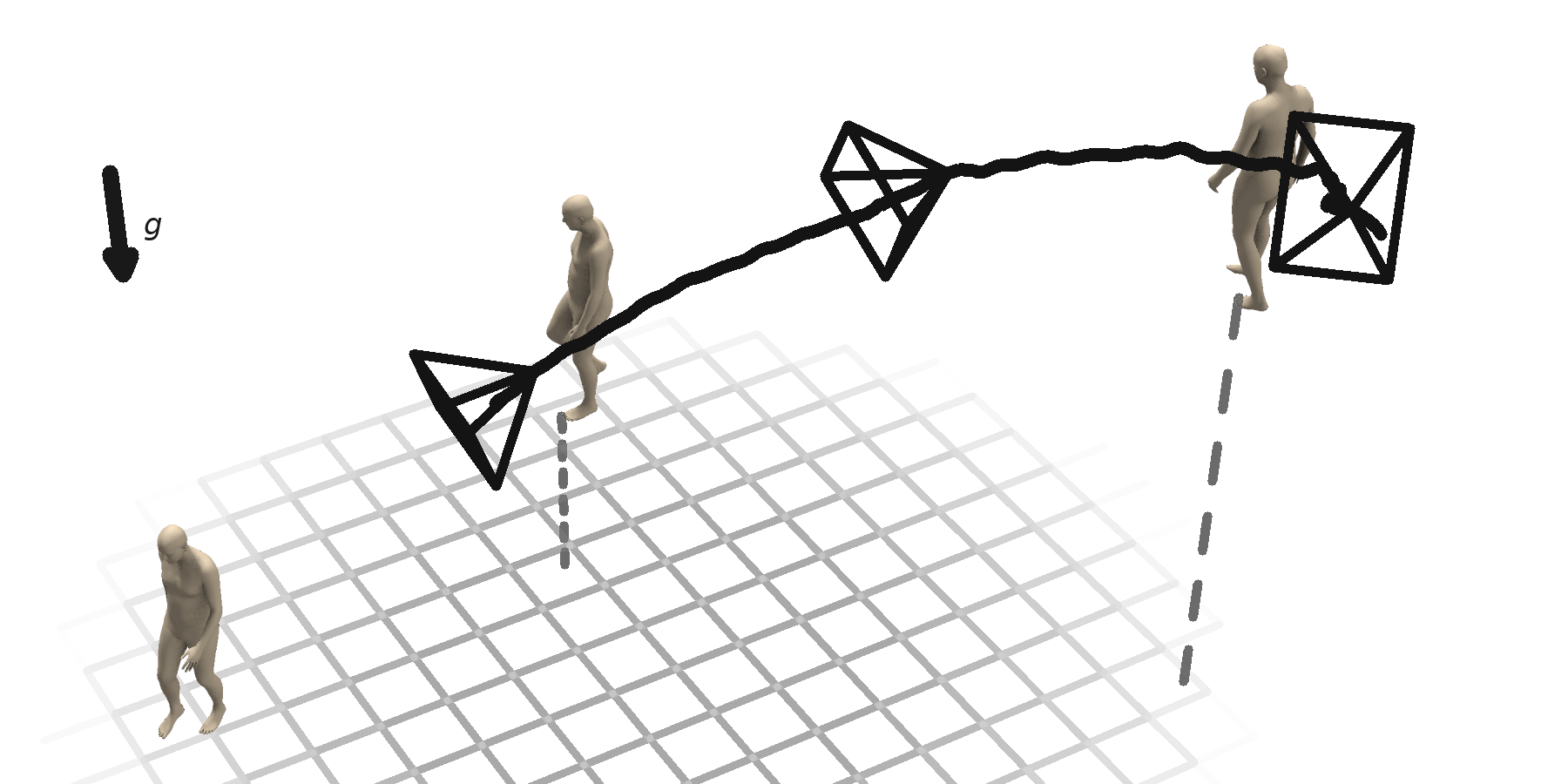}};
\node[inner sep=0pt] at (66.5+\W/2,\Y+\h/2) {\includegraphics[width=\W mm, height=\h mm, keepaspectratio]{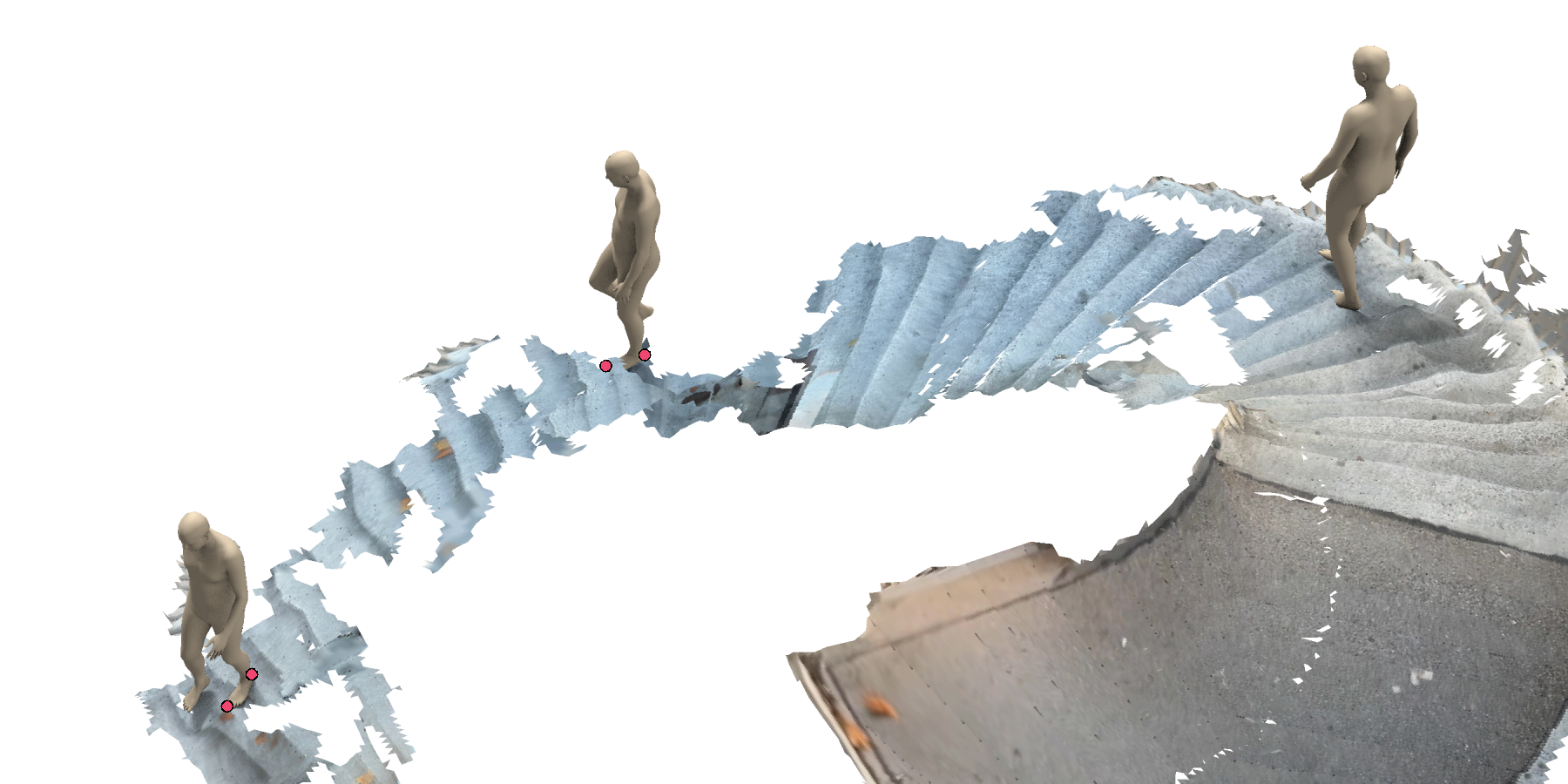}};
\ifdense\node[inner sep=0pt, anchor=east] at (133+\W,\Y+\h/2) {\includegraphics[width=\W mm, height=\h mm, keepaspectratio]{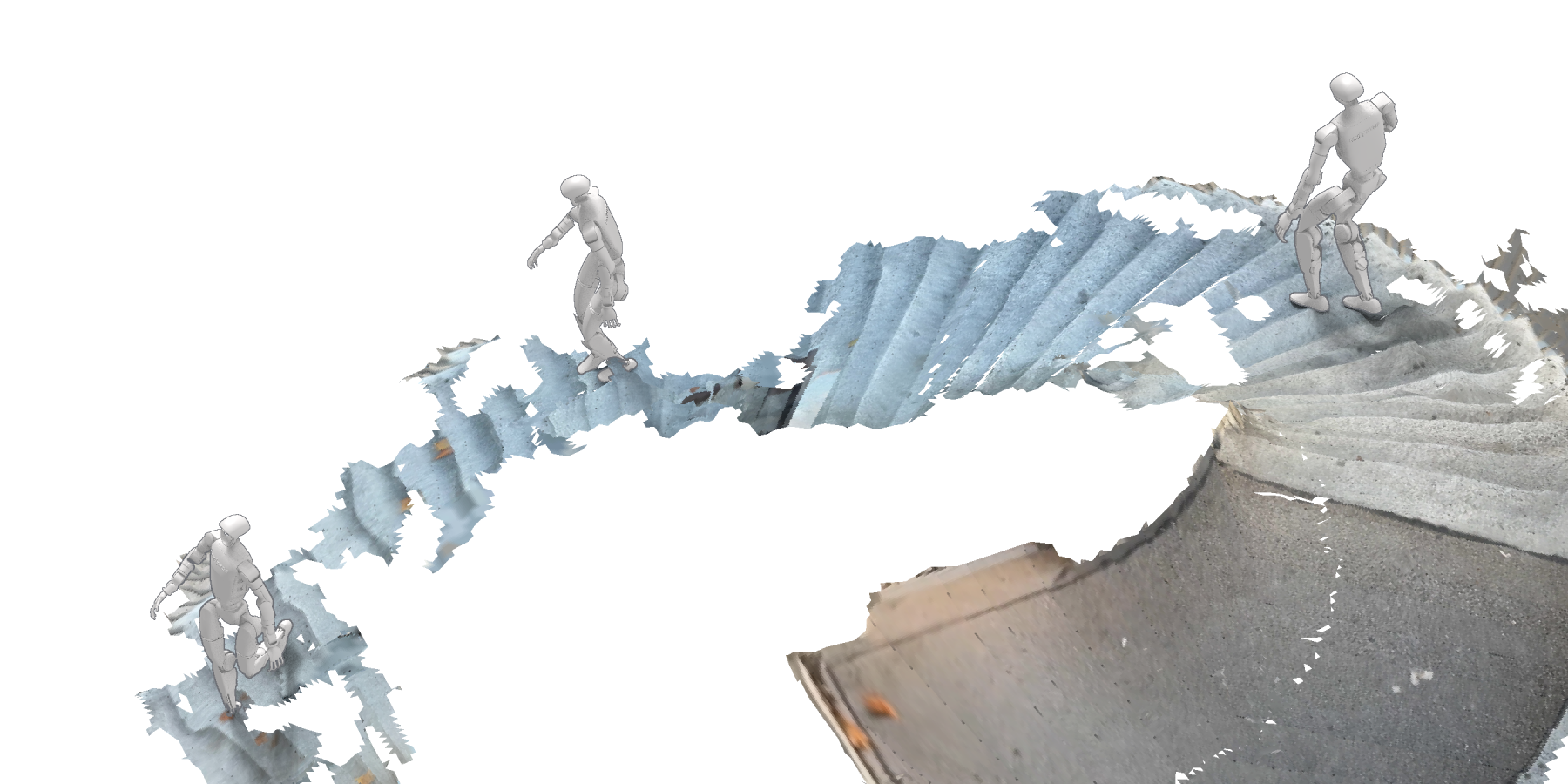}};\else\node[inner sep=0pt] at (133+\W/2,\Y+\h/2) {\includegraphics[width=\W mm, height=\h mm, keepaspectratio]{figures/fig2v2/strip_humanoid.png}};\fi
\draw[arr] (\W+1.5,\Y+\h/2) -- (65,\Y+\h/2);
\draw[arr] (66.5+\W+1.5,\Y+\h/2) -- (131.5,\Y+\h/2);
\ifdense
\def\bz{-3.0}
\draw[red!65!black, line width=1.0pt, -{Stealth[length=2.2mm, width=1.8mm]}] (pt.south |- 0,-0.4) -- (pt.south |- 0,\bz) -- (150.6,\bz) -- (150.6,\Y+\h+6.2);
\draw[red!65!black, line width=0.6pt, dash pattern=on 1.4pt off 1.0pt] (sm.south |- 0,-0.4) -- (sm.south |- 0,\bz);
\node[font=\sffamily\tiny, text=red!65!black, anchor=south west, inner sep=0pt] at ([xshift=0.7mm]sm.south |- 0,\bz+0.5) {or mesh};
\draw[white, line width=2.4pt] (93,\bz+0.9) -- (93,\bz-0.9);
\draw[arr] (93,-1.4) -- (93,\Y+\h+6.2);
\node[img] (cth) at (133.8,\Y+\h-12.6) {\includegraphics[height=12mm]{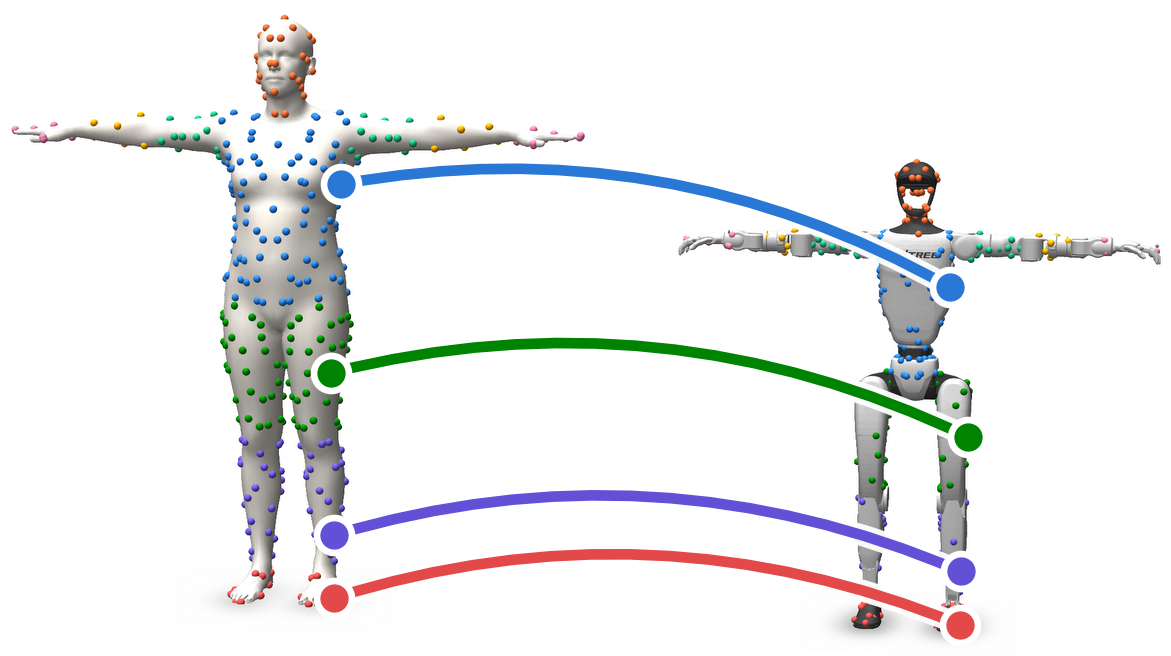}};
\draw[red!65!black, line width=0.6pt, rounded corners=1.5pt] ([shift={(-0.4,-0.4)}]cth.south west) rectangle ([shift={(0.4,0.4)}]cth.north east);
\node[font=\sffamily\tiny, text=red!65!black, anchor=north west, inner sep=0pt] at ([yshift=-1.0mm]cth.south west) {points $\to$ G1};
\fi
\end{tikzpicture}\fi}
\caption{\ifdense One network per person and frame predicts  dense points, keypoints, body parts and heel / toe contacts; SMPL is fitted to the points. Per clip: a metric, gravity-aligned world; the scene placed on the body via contacts; \figtworet; for illustration, a terrain-aware policy tracks it in simulation.\else The \paper pipeline.  One network per person and frame predicts dense points, keypoints, heel / toe contacts and G1 joint angles; SMPL is fitted to the points. Per clip: a metric, gravity-aligned world; the scene placed on the body via contacts; \figtworet; for illustration, a terrain-aware policy tracks it in simulation. Walls cut above 0.5\,m. Images: EMDB.\fi}
\label{fig:overview}
\end{figure*}
\begin{abstract}
We present \paper (\textbf{F}ast and \textbf{A}ccurate \textbf{B}ody estimation), a human body estimator that achieves leading inference efficiency and body recovery faithfulness at the same time.
As a 3D human pose estimator, \paper's core is to predict human skeleton joints and surface dense points directly, rather than only parametric mesh model parameters.
In pixels, \paper predicts dense 2D and 3D body points directly, alongside a parametric (MHR) head, and fits SMPL to them in one learned pass and two Gauss--Newton steps.
Its ViT-L student \paper-L, distilled from a fine-tuned SAM 3D Body, runs in real time for one person on one RTX 4070 Ti SUPER, from person detection to a fitted body, 4.6$\times$ faster than SAM 3D Body's body-only path in eager PyTorch, and is more accurate than SAM 3D Body on most camera-space benchmarks.
The same network predicts heel, toe, hand, and knee contacts, which tie the body to the scene.
In world space, \paper-L is more accurate than many recent methods.
In scenes, \paper fuses the static scene's depth around the visible person and aligns the scene and body through the predicted contacts. 
\paper also brings convenience for the robotics task of human-to-humanoid real-to-sim motion transfer. 
An in-model G1 head and FAB-Retarget, a motion-level retargeter that reads the dense points (or the mesh fitted to them), map the motion to a Unitree G1.
FAB-Retarget's references follow the person more closely than those of the specialized retargeters (SOMA, GMR, UMR, OmniRetarget) and much more closely than Fast SAM 3D Body's body retargeted by our sequence IK. Project page: \mbox{\url{https://www.jinkuncao.com/FAB}}.
\end{abstract}

\section{Introduction}
\label{sec:intro}

Monocular human pose and shape estimation underpins motion capture, digital humans and robot learning from human video.
For robots, the target is no longer a mesh that overlays the image well but motion in a metric, gravity-aligned world, in contact with a reconstructed scene, that a humanoid can track~\cite{he2024omnih2o,fu2024humanplus,allshire2025videomimic,he2025asap}.
Current systems split this into stages built on different assumptions. The strongest per-image models~\cite{yang2026sam3dbody} run 840M-parameter encoders below 10\,fps per person, even with training-free acceleration~\cite{yang2026fastsam3dbody}. Video and world-space methods~\cite{shin2024wham,wang2024tram,shen2024gvhmr,li2025genmo} build on weaker per-frame estimators. Real2sim pipelines~\cite{allshire2025videomimic} reconstruct the scene and the body separately.

These stages' outputs also pull against each other. Keypoints regressed from a parametric mesh, once projected, often disagree with the observed 2D joints~\cite{zhang2021pymaf,joo2020eft}. Much of the 3D supervision is pseudo ground truth that inherits a fitting prior~\cite{bogo2016smplify,moon2020neuralannot}. For video and robots, smooth, physically consistent motion matters more than per-frame millimetres.

In the spirit of achieving these outputs with visual faithfulness and physical feasibility, we present \paper, one system that estimates the body in pixels, in the world and in the scene with a fast per-frame network, and transfers the motion to a humanoid. Dense body points are its central output. The optional SMPL is fitted to them, and \ifdense the G1 motion is retargeted from the points themselves, matched to the G1's surface\else the G1 motion is retargeted from that body\fi. The outputs are coupled only weakly: the 2D points are not projections of the 3D ones, and the G1 head's joints are not inverse kinematics on the mesh. 
We build \paper on distilling the per-frame SAM 3D Body model and aim to achieve real-time inference on consumer GPU devices. We use the ViT-L backbone model \paper-L as the default, and the bigger \paper-H and the smaller \paper-S models are also provided.
We summarize our contributions as:

\textbf{A fast, points-first body estimator.} \paper  predicts dense 2D and 3D body points (keypoints, skeleton joints and surface points) directly, and fits SMPL~\cite{loper2015smpl} to them in one learned pass and two Gauss--Newton steps. It runs at 39.2\,fps for one person on one RTX 4070 Ti SUPER and is more accurate than SAM 3D Body on multiple benchmarks.

\textbf{Contacts as a first-class output}: static and any contact for heels, toes, hands and knees, per image and causally over video; they drive the human--scene alignment and is important for ground-aware humanoid pose retargeting.

\textbf{World-space motion} from one ViPE~\cite{huang2025vipe} run and single-image gravity~\cite{veicht2024geocalib}, with the metric scale taken from the body through the joint human--scene optimization.
\textbf{Human--scene reconstruction} that keeps the person visible and in contact: the static scene is fused around the person, geometry covering the visible body is carved, and a joint optimization scales the scene to the body through contacts.

\textbf{Human-to-humanoid retargeting}: an in-model G1 head and a motion-level retargeter \ifdense that reads the dense points\else on the body fitted to the points\fi{} to achieve state-of-the-art visual alignment\ifdense{} and tracking feasibility\fi. The retargeting can also be bundled with scene reconstruction. 

\section{Related Work}
\label{sec:related}

\paragraph{Body representations and per-image recovery.}
2D keypoint estimators~\cite{cao2018openpose,sun2019hrnet,xu2022vitpose} are accurate and fast. 3D keypoints are lifted from 2D~\cite{martinez2017simple,pavllo2018videopose3d} or regressed as metric heatmaps or localizer fields~\cite{sarandi2020metrabs,sarandi2024nlf}, with skeletons that differ across datasets~\cite{sarandi2022dozens}.
Parametric models (SMPL~\cite{loper2015smpl}, SMPL-X~\cite{pavlakos2019smplx}, MHR~\cite{ferguson2025mhr}) are fitted~\cite{bogo2016smplify} or regressed~\cite{kanazawa2017hmr,kolotouros2019spin,li2022cliff,goel2023humans4d}. Later work improves camera modeling~\cite{patel2024camerahmr}, promptability~\cite{wang2025prompthmr} and scale~\cite{cai2023smplerx}, and realigns the projected keypoints with the image~\cite{zhang2021pymaf}.
SAM 3D Body~\cite{yang2026sam3dbody} combines a DINOv3~\cite{simeoni2025dinov3} encoder, promptable decoders and the Momentum Human Rig. Fast SAM 3D Body~\cite{yang2026fastsam3dbody} accelerates it without retraining. FastHMR~\cite{mehraban2025fasthmr} merges tokens and layers, and one-stage methods recover all people in one pass~\cite{baradel2024multihmr,fiche2026multihmr2,su2024sathmr,lee2026detram}.
\paper predicts dense points, as NLF does~\cite{sarandi2024nlf}, and fits the body model to them with a learned fit~\cite{song2020lgd}. FAH~\cite{cao2026fastaccuratehandestimation} follows the same methodology for hands (points first, then a parametric fit, on a DINOv3 backbone, in pixels, interaction and world space); \paper applies it to the whole body and extends it to scenes and humanoid transfer. Unlike these per-image approaches, \paper uses dense points as the common representation for body recovery, world and scene reconstruction, and humanoid transfer.

\paragraph{Video and world space.}
Video pose estimation requires more temporal smoothness than per-frame alignment, and world space estimation imposes spatial smoothness constraints.
WHAM~\cite{shin2024wham}, TRAM~\cite{wang2024tram}, GVHMR~\cite{shen2024gvhmr}, GENMO~\cite{li2025genmo} and DuoMo~\cite{wang2026duomo} recover temporally consistent, world-grounded motion, typically with visual odometry~\cite{teed2023dpvo}. OnlineHMR~\cite{zhao2026onlinehmr} and Human3R~\cite{chen2025human3r} run causally. Composing camera-space human depth with odometry suffers from a scale mismatch~\cite{chien2026diagnostic}. \paper instead obtains metric scale from the body and resolves body--scene scale jointly through predicted contacts.

\paragraph{Humans in scenes and real2sim.}
Scene constraints~\cite{hassan2019prox} and image contact estimators~\cite{huang2022rich,tripathi2023deco} reduce pose ambiguity compared to pure pixels. JOSH~\cite{liu2025josh} and HSfM~\cite{muller2025hsfm} optimize humans, scenes, and cameras jointly.
VideoMimic~\cite{allshire2025videomimic} reconstructs a person and a scene from a phone video with MegaSaM~\cite{li2025megasam}, retargets the motion to a humanoid and trains a terrain-aware controller on the result. GaussGym~\cite{escontrela2025gaussgym} renders reconstructed scenes for vision-based locomotion. \paper keeps the person visible while fusing the scene and uses the predicted contacts to jointly align the reconstructed body and scene.

\paragraph{Human-to-humanoid motion transfer.}
Humanoid controllers learn to track human motion retargeted to the robot~\cite{luo2023phc,he2024h2o,he2024omnih2o,fu2024humanplus,he2025asap,luo2025sonic}, increasingly from monocular video~\cite{allshire2025videomimic}. Retargeting, usually after pose estimation, uses shape fitting~\cite{he2024h2o}, inverse kinematics~\cite{araujo2025gmr,nvidia2026soma}, learned point correspondences~\cite{cao2026umr} or interaction meshes~\cite{yang2025omniretarget}.\ifdense{} \paper-Retarget directly uses its dense points in one motion-level optimization with contact and terrain constraints.\fi

\section{Method}
\label{sec:method}
The dense body points are the network's central output: SMPL is fitted to them (Sec.~\ref{sec:pixels}),  and the world, scene and retargeting stages build on that body\ifdense; FAB-Retarget's points mode also reads the points\fi{} (Secs.~\ref{sec:world}--\ref{sec:fabretarget}).

\subsection{Pixels: a fast, points-first estimator}
\label{sec:pixels}

\paragraph{Making SAM 3D Body fast.}

SAM 3D Body~\cite{yang2026sam3dbody} encodes a $512\times512$ person crop with an 840M-parameter DINOv3 ViT-H+ and refines MHR parameters with a promptable decoder. We first make it fast without retraining (Sec.~\ref{sec:suppmethod}).
We then fine-tune it lightly on our mixture (\paper-H), with an SMPL-space loss against exact SMPL ground truth (BEDLAM~\cite{black2023bedlam}, 3DPW-train~\cite{vonmarcard2018_3dpw}, MOYO~\cite{tripathi2023moyo}, RICH~\cite{huang2022rich}). \paper-H is thus a small modification of SAM 3D Body: it keeps SAM 3D Body's architecture, starts from its weights, freezes all but the last blocks of its backbone and fine-tunes the rest, and adds our dense-point head, trained on the frozen body (Sec.~\ref{sec:suppmethod}). Unlike \paper-L and \paper-S, whose whole backbone, body and dense heads are trained jointly on our full recipe, \paper-H is a light fine-tune; our compute did not allow full training at its size.

\paragraph{Fast student (\paper-L).}
\paper-L replaces the backbone with a DINOv3 ViT-L/16 at the same input and token grid, projected to SAM 3D Body's embedding, and keeps SAM 3D Body's decoder. It is trained with token-level feature distillation from SAM 3D Body's frozen backbone, output distillation from \paper-H, and human labels where they exist (Sec.~\ref{sec:suppmethod}). Without SMPL ground truth, \paper-H's outputs are the parametric head's targets. \paper-S, the same recipe on a DINOv3 ViT-S+/16, runs 1.63$\times$ as fast as \paper-L.

\paragraph{Dense points first.}
A light head predicts per-point heatmaps on the image tokens, upsampled $4\times$ and fused with a stride-4 pixel stem. Each point is queried by the decoder's pose token plus a learned embedding. A soft-argmax gives its 2D position, and an MLP on the features under its heatmap gives its depth; the point is the 2D position back-projected along its camera ray to that depth and placed by a closed-form root translation. Three point types are each supervised in their own convention: 70 image-defined keypoints by human 2D labels and \paper-H's 3D; the 14 evaluation joints like the surface points; and 468 SMPL surface points by SMPL ground truth or \paper-H's mesh.
The points are predicted directly, not as offsets on the parametric mesh, and heels and toes get direct 3D supervision where ground-truth feet are reliable (BEDLAM, MOYO). The 2D and 3D points are coupled only through reprojection: the 3D points must project onto the 2D labels (keypoints) or onto the head's own detached 2D (mesh points).

\paragraph{One-pass learned fit.}
A learned fitter maps the surface points to SMPL pose and shape: a Kabsch alignment into the template frame, then two learned refinement stages that compare the current fit with the points. Two robust Gauss--Newton steps on the points refine the fit, and a closed-form foot refinement re-solves the ankle and toe rotations onto the points' feet. \paper-H's SMPL by default comes from its MHR mesh, \paper-L's and \paper-S's from their dense points.

\begin{figure}[t]
\centering
\includegraphics[width=\linewidth]{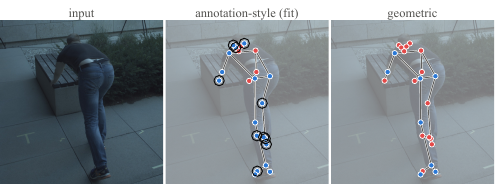}
\caption{Visibility of the 23 body keypoints on RICH: input, annotation-style (trained on human visibility flags; read by the fit) and geometric (trained on occlusion rendered from the target mesh) outputs; blue visible, red occluded, ring: disagrees with the ground-truth mesh.}
\label{fig:vis}
\end{figure}

\paragraph{Body parts and visibility.}
A part head on the frozen body segments 14 body parts and the background in the person's crop, trained on labels rendered from the target mesh and clipped to SAM~3's~\cite{carion2025sam3} person mask. It lowers the one-person rate by 1.2\,fps. A visibility head labels the dense head's keypoints and joints visible, occluded or outside the crop (Fig.~\ref{fig:vis}).

\paragraph{Person detector.}
A one-class YOLO11m~\cite{jocher2024yolo11} with labels distilled from SAM~3 runs at 246\,fps (ultralytics' predict call, host pre- and post-processing); against the stock YOLO11m it trades 2.8 COCO~\cite{lin2014coco} person AP for recall of rotated, low-light and crowded people (Sec.~\ref{sec:suppmethod}).

\subsection{Contacts}
\label{sec:contacts}
A 2-layer MLP reads the pose token and the dense head's features at twelve keypoints (ankles, heels, big and small toes, knees, wrists), plus the camera translation. It predicts two contact types for each heel, toe, hand and knee: \emph{static} contact with a world-fixed surface, and \emph{any} contact, including moving supports such as a skateboard.
The labels come from three sources: the world-frame ground truth of BEDLAM, RICH, MOYO and EgoBody (lowest vertex near the local floor and slow), RICH's human--scene contact annotations, and contacts from simulating the ground-truth motion. The head is trained on the frozen body, so no pose output changes.
A causal 2-layer GRU over a track's head features and keypoints predicts a residual on these logits. Its keypoints are expressed in the world frame, so a planted foot stays static under a moving camera. 

\subsection{World and Scenes}
\label{sec:world_and_scenes}\label{sec:world}\label{sec:scenes}
The world motion is one pipeline, shared with the scene. Each track's per-frame body is smoothed offline with a zero-phase Gaussian. Cameras and metric video depth come from one ViPE~\cite{huang2025vipe} run with UniDepth~\cite{piccinelli2024unidepth}. Gravity comes from GeoCalib~\cite{veicht2024geocalib}. The metric scale of cameras and scene comes from the body itself, through the joint optimization.

\paragraph{Scene geometry.}
The static scene's depth comes from multi-view stereo~\cite{schoenberger2016mvs} on ViPE's fixed cameras where the parallax suffices and from ViPE's metric depth elsewhere (the \emph{MVS scene}); it is fused in a GPU voxel hash. Dynamic objects and the person's body are masked, and free-space carving removes points that the camera sees through. \emph{Human-visibility carving} then removes the scene (mesh vertices and fused points) that projects inside the visible person (the person mask intersected with the body silhouette) and lies more than 5\,cm + 2\% of the body's depth in front of the body, tested on every second frame.
A truncated signed distance function (TSDF)~\cite{curless1996volumetric} gives the scene mesh, plus a heightfield and boxes for collision.

\paragraph{Joint human--scene optimization.}
The body and the scene come from different estimates of depth, and a fixed scale factor between them leaves the feet floating or sunk. We keep \paper's camera-frame body, whose depth is accurate where the intrinsics are right, and bring the scene to it. One optimization per video solves for a single global log-scale of scene and cameras, constant over the video. Its robust terms are contact (on frames with predicted contact, the signed distance of the heel, toe, hand and knee vertices to the scene), a penetration hinge, body--scene depth consistency on the visible person, and root acceleration and foot skate.

\subsection{\ifdense FAB-Retarget\else Humanoid\fi}
\label{sec:fabretarget}\label{sec:humanoid}
\newcommand{\densemethod}{Each dense point (an SMPL template vertex) is matched once to a G1 surface anchor by construction, not by learning as in UMR~\cite{cao2026umr}: in a matched T-pose, it takes the G1 sample of its body part nearest in cylindrical coordinates $(s,\theta)$ about the part's bone (Fig.~\ref{fig:corr}). Per frame, a Kabsch fit of each rigid group's points to their T-pose, turned so that the group's G1 bone lies on the skeleton's, gives a target $R^{*}_{g}$ (swing from the joints, twist from the points). The term averages $\lVert(R_g(q)-R^{*}_{g})\,a_i\rVert^2$ over groups, frames and anchors $a_i$ and is free of limb proportions.}

\paragraph{In-model G1 head.}
Batched GPU inverse kinematics retargets each training body to the 29 joints of the Unitree G1, and a head on the pose token learns to regress these joint angles from images in one pass. The labels are direction-first: the robot's bones follow the human's bone directions, not scaled positions. A \emph{sequence IK} solves the same IK jointly over a whole clip, with one scale per clip and an acceleration penalty.


\ifdense
\paragraph{Motion-level retargeting.}
For whole motions, FAB-Retarget, our motion-level retargeter, maps a human motion to a G1 reference by one optimization over the clip (Sec.~\ref{sec:suppmethod}). Its input is a joint skeleton (SMPL's 24 joints, onto which other skeletons map; \emph{skeleton mode}, FAB-Retarget [skeleton] in the tables), optionally with an SMPL mesh (\emph{mesh mode}, [skeleton + mesh]) and dense points (\emph{points mode}, [skeleton + mesh + points], the default); the source skeleton or mesh does not necessarily come from \paper\ (e.g.\ motion capture). So FAB-Retarget can be used as a separate retargeting method from \paper. The contact is an optional signal to \paper-Retarget to improve ground-aware retargeting consistency and prevent G1 floating.

\paragraph{Objective.}
At one human-to-G1 scale per clip and in one configuration for every clip (Sec.~\ref{sec:suppmethod}), it combines a direction-first pose term on the bone directions and pelvis frame, contact and terrain terms (labeled stance feet stay planted without sliding, and no foot or link penetrates the floor or the reconstructed terrain; hands and knees act as clearance only), and self-collision, acceleration and joint-velocity-limit penalties. A stance pin then fixes each stance foot on the terrain. 

\paragraph{Modes.}
The modes differ only in what the contact labels (feet and seat) and the root's start height are computed from, mesh vertices or joints lowered by the SMPL template's offsets (Sec.~\ref{sec:suppmethod}), and in the point term of the points mode: each dense point is matched once to a constructed G1 surface anchor, not a learned one as in UMR~\cite{cao2026umr}, and the G1's part rotations follow the points'.

\else
\paragraph{Motion-level retargeting.}
For whole motions, a \emph{motion-level retargeter} optimizes the clip jointly on the world body fitted to the dense points. Its pose term is direction-first. Its contact terms keep the human's foot contacts planted without sliding or penetrating the floor or the reconstructed terrain; hands and knees act as clearance only. Stance pinning then fixes each stance foot on the cleaned terrain and re-solves legs and root. Acceleration and joint-velocity-limit penalties respect the robot's limits. Its settings are in Tab.~\ref{tab:retargetobj}; the 25 test clips are not held out from the retargeter's design.
\ifdense\else\ifablready
An ablation also reads the dense points, matched to G1 surface anchors after UMR~\cite{cao2026umr}, and the predicted contacts (Tab.~\ref{tab:retargetinputs}).
\fi\fi
\fi

\begin{figure}[t]
\centering
\includegraphics[width=\linewidth]{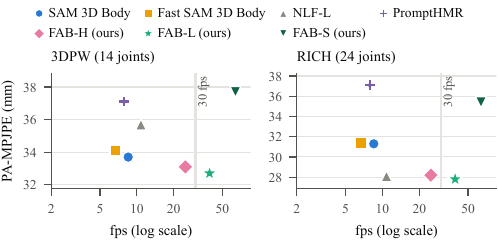}
\caption{Speed against accuracy: PA-MPJPE (mm) of Tab.~\ref{tab:main}'s image methods against end-to-end fps on one RTX 4070 Ti SUPER, person detection included. Grey line: 30\,fps (real time). Lower right is better.}
\label{fig:speed}
\end{figure}
\begin{table}[t]
\centering\tabfont
\caption{Camera space, re-run by us under the standard protocol on the same frames (mm; PA: PA-MPJPE; 3DPW: 14 H36M~\cite{ionescu2014human36m} joints; RICH test: 24 SMPL joints). \paper-L / \paper-S: SMPL fitted to their dense points; \paper-H is also reported with its MHR mesh converted to SMPL, by our fit (its default) and by MHR's official converter (official) (Sec.~\ref{sec:pixels}; timing and inputs per method: Sec.~\ref{sec:suppprotocols}). Bold: best per column; shaded: ours (also in the other tables).}
\label{tab:main}
\setlength{\tabcolsep}{1.2pt}
\begin{tabwrap}
\resizebox{\linewidth}{!}{\begin{tabular}{l cc cc c}
\toprule
& \multicolumn{2}{c}{3DPW} & \multicolumn{2}{c}{RICH} & \\
\cmidrule(lr){2-3}\cmidrule(lr){4-5}
Method & PA & MPJPE & PA & MPJPE & fps \\
\midrule
TRAM~\cite{wang2024tram} & 36.4$^\dagger$ & 58.8 & 40.0 & 68.7 & 1.0 \\
GVHMR~\cite{shen2024gvhmr} & 38.4$^{\diamond\dagger}$ & 58.2$^\diamond$ & 38.0 & 66.6 & 9.2 \\
GEM-X~\cite{nvidia2026gemx} (GENMO) & 37.6 & 59.8 & 35.3 & 56.4 & 2.0 \\
WHAM~\cite{shin2024wham} & 38.3$^\dagger$ & 63.8 & 42.2 & 77.9 & 2.8 \\
DuoMo~\cite{wang2026duomo} & 36.7$^\dagger$ & 59.4 & 35.0 & 49.5 & 1.1$^{\|}$ \\
SAM 3D Body~\cite{yang2026sam3dbody} & 33.7$^\dagger$ & 55.1 & 31.3 & 50.0 & 8.5$^\ddagger$ \\
Fast SAM 3D Body~\cite{yang2026fastsam3dbody} & 34.1$^\dagger$ & 55.5 & 31.4 & 49.8 & 6.7 \\
NLF-L~\cite{sarandi2024nlf} & 35.7$^\dagger$ & 57.7 & 28.1$^\dagger$ & \textbf{45.3} & 10.8 \\
PromptHMR~\cite{wang2025prompthmr} & 37.1$^\dagger$ & 59.1 & 37.1 & 54.5 & 7.9 \\
\rowcolor{oursbg}
\paper-H, MHR mesh $\to$ SMPL, official & \textbf{32.4}$^\dagger$ & \textbf{52.7} & 27.9$^\dagger$ & 46.0 & 0.12 \\
\rowcolor{oursbg}
\paper-H, MHR mesh $\to$ SMPL, our fit & 33.1$^\dagger$ & 53.2 & 28.2$^\dagger$ & 45.6 & 24.9 \\
\rowcolor{oursbg}
\paper-H, dense points $\to$ SMPL & 33.7$^\dagger$ & 54.8 & 29.8$^\dagger$ & 48.2 & 21.8 \\
\rowcolor{oursbg}
\paper-L (ours) & 32.7$^\dagger$ & 53.1 & \textbf{27.8}$^\dagger$ & 45.5 & 39.2 \\
\rowcolor{oursbg}
\paper-S (ours) & 37.7$^\dagger$ & 59.8 & 35.4$^\dagger$ & 56.4 & \textbf{63.8} \\
\bottomrule
\end{tabular}}
\end{tabwrap}
\tabnotes{fps: end to end on one RTX 4070 Ti SUPER (TensorRT body for \paper-L / -S; protocol: Sec.~\ref{sec:latency}). $^\dagger$Trains on this benchmark's training split (NLF-L on RICH: split unstated). $^\diamond$Flip test-time augmentation. $^\ddagger$Body-only mode, eager PyTorch. $^{\|}$Exceeds the card's 16\,GiB.}
\end{table}

\begin{table*}[t]
\centering\tabfont
\caption{World space on the same frames: EMDB-2 (all 25 sequences) and RICH-191. Top: world methods with their released code and own inputs (Sec.~\ref{sec:suppvideoworld}). Middle: per-frame bodies (our models as in Tab.~\ref{tab:main}) through our world pipeline, with the same cameras, boxes and contacts. Bottom: before the joint human--scene step, and with the ground-truth camera motion; last row: \paper-L with the ground-truth boxes.}
\label{tab:worldmain}
\setlength{\tabcolsep}{2.8pt}
\begin{tabwrap}
\begin{tabular}{l c lccccc lccc}
\toprule
& & \multicolumn{6}{c}{EMDB-2 (25 sequences)} & \multicolumn{4}{c}{RICH-191} \\
\cmidrule(lr){3-8}\cmidrule(lr){9-12}
Method & Box & Cam. & WA & W & RTE & J & FS & Cam. & WA & W & RTE \\
\midrule
TRAM~\cite{wang2024tram} & GT & GT K, SLAM & 93.3 & 328.3 & 3.20 & 18.6 & 13.9 & GT K, static & 95.9 & 154.5 & 2.24 \\
GVHMR~\cite{shen2024gvhmr} & GT & est.\ K, DPVO & 109.1 & 272.8 & 1.90 & 16.1 & \textbf{3.4} & GT K, static & 77.1 & 125.1 & 2.34 \\
GEM-X~\cite{nvidia2026gemx} (GENMO) & GT & GT K, GT R & 114.2 & 270.9 & 2.54 & 35.3 & 4.3 & GT K, GT R & 79.3 & 128.5 & 2.58 \\
WHAM~\cite{shin2024wham} & GT$^\ast$ & FOV, GT R & 148.8 & 384.5 & 4.63 & 22.0 & 5.9 & FOV, static & 117.5 & 201.9 & 5.06 \\
DuoMo~\cite{wang2026duomo} & GT & GT K, TRAM SLAM & 67.5 & 167.9 & 1.19 & 9.5 & 3.9 & GT K, static & 55.6 & 83.4 & 1.70 \\
CRISP~\cite{wang2026crisp}$^\S$ (human--scene) & own & MegaSaM & 74.7 & 208.7 & 1.46 & 53.6 & 12.4 & MegaSaM & 91.7 & 158.1 & 3.37 \\
\midrule
\multicolumn{12}{l}{\emph{per-frame bodies, our detector and cameras}} \\
SAM 3D Body &  &  & 56.0 & \textbf{141.0} & 1.06 & 5.2 & 6.3 &  & 73.9 & 144.0 & 5.67 \\
Fast SAM 3D Body &  &  & 56.3 & 141.1 & 1.06 & 5.3 & 6.4 &  & 74.4 & 145.4 & 5.70 \\
NLF-L &  &  & 60.4 & 151.8 & 1.14 & 5.7 & 6.9 &  & 69.7$^\dagger$ & 137.2 & 5.72 \\
\rowcolor{oursbg}
\paper-H, MHR mesh $\to$ SMPL &  &  & \textbf{55.4} & 142.5 & 1.09 & 5.2 & 6.2 &  & 68.8$^\dagger$ & 135.5 & 5.75 \\
\rowcolor{oursbg}
\paper-H, dense points $\to$ SMPL &  &  & 59.0 & 147.3 & 1.04 & 5.5 & 7.7 &  & 75.6$^\dagger$ & 150.2 & 6.01 \\
\rowcolor{oursbg}
\paper-L (ours) &  &  & 57.1 & 147.2 & 1.06 & 5.3 & 6.8 &  & 65.9$^\dagger$ & 132.2 & 5.73 \\
\rowcolor{oursbg}
\paper-S (ours) &  &  & 61.9 & 163.8 & 1.13 & 5.8 & 8.6 &  & 80.1$^\dagger$ & 155.2 & 5.93 \\
\midrule
\rowcolor{oursbg}
\paper-L: ViPE camera at its own scale, smoothed & ours & ViPE scale & 60.9 & 170.4 & 1.55 & 4.7 & 7.6 & ours & 65.9$^\dagger$ & 132.2 & 5.73 \\
\rowcolor{oursbg}
\paper-L, ground-truth camera motion & ours & ours K, GT motion & 51.0 & 137.9 & 0.19 & 4.0 & 6.8 & ours K, GT motion & 65.9$^\dagger$ & 132.5 & 5.73 \\
\midrule
\rowcolor{oursbg}
\paper-L (ours), ground-truth boxes & GT & ours & 55.7 & 143.0 & \textbf{1.03} & \textbf{5.1} & 6.2 & ours & \textbf{50.6}$^\dagger$ & \textbf{77.4} & \textbf{1.16} \\
\bottomrule
\end{tabular}
\end{tabwrap}
\tabnotes{WA / W: WA- / W-MPJPE (mm); RTE (\%); jerk J (10\,m\,s$^{-3}$); foot sliding FS (mm). Box: GT, ours (our detector) or own (its masks). Cam.: intrinsics K (FOV: fixed field of view) and camera motion (own SLAM / DPVO~\cite{teed2023dpvo} / MegaSaM, GT rotation R, or static); ours: our focal rule and ViPE. Bold: best in the top and middle blocks and the last row. $^\ast$RICH-191: box from WHAM's ViTPose keypoints in the GT box. $^\S$Overlapping 600-frame chunks. $^\dagger$Trains on RICH's training split (NLF-L: split unstated). Details: Sec.~\ref{sec:suppvideoworld}.}
\end{table*}

\section{Experiments}
\label{sec:experiments}

\paragraph{Datasets.}
Camera space: 3DPW~\cite{vonmarcard2018_3dpw} test, RICH~\cite{huang2022rich} test, EgoBody~\cite{zhang2022egobody}, Harmony4D~\cite{khirodkar2024harmony4d} and SSP-3D~\cite{sengupta2020ssp3d}. World space: EMDB-2~\cite{kaufmann2023emdb} and RICH-191 (RICH's 191 static-camera test videos, the GVHMR / WHAM protocol). Scenes: RICH-3 (three RICH test clips), EgoBody, SLOPER4D~\cite{dai2023sloper4d} and PROX~\cite{hassan2019prox}, with their scans. Every method is scored on the same EMDB-2 frames.

\paragraph{Metrics.}
(PA-)MPJPE (mm); as in WHAM, W- / WA-MPJPE, RTE, jerk J and foot sliding FS, with world ground-truth joints from the neutral SMPL regressor; for the G1, the mean angle between the robot's and the person's bone directions\ifablready{} and pelvis-relative part rotations\fi. Definitions: Secs.~\ref{sec:suppprotocols} and~\ref{sec:supphumanoid}; curated benchmarks: Sec.~\ref{sec:benchmarks}.

\paragraph{Baselines.}
Every baseline is re-run by us under one protocol~\footnote{This protocol reproduces SAM 3D Body's published 3DPW numbers within 0.3\,mm.} per benchmark: ground-truth boxes, ground-truth intrinsics, and MHR meshes converted to SMPL by MHR's official converter; video methods keep their own camera inputs (Secs.~\ref{sec:suppprotocols} and~\ref{sec:suppvideoworld}).

\subsection{Images and Videos}
\label{sec:exp_pixels}
Of the timed models, \paper-L and \paper-S are the only ones at 30\,fps or more, and \paper-L also has the lowest PA-MPJPE on RICH test (MPJPE 0.2\,mm behind NLF-L's) and is more accurate than SAM 3D Body on 3DPW (Tab.~\ref{tab:main}, Fig.~\ref{fig:speed}). With MHR's official converter from its MHR mesh to SMPL, which is not real-time, \paper-H, our light fine-tune of SAM 3D Body (Sec.~\ref{sec:pixels}), is the most accurate model on 3DPW.
On Harmony4D and EgoBody's egocentric test (Tab.~\ref{tab:morebench}), \paper-L leads every baseline; on Harmony4D it leads at every overlap level, most when people overlap (Sec.~\ref{sec:suppcamera}).

\begin{figure*}[t]
\centering
\includegraphics[width=\linewidth]{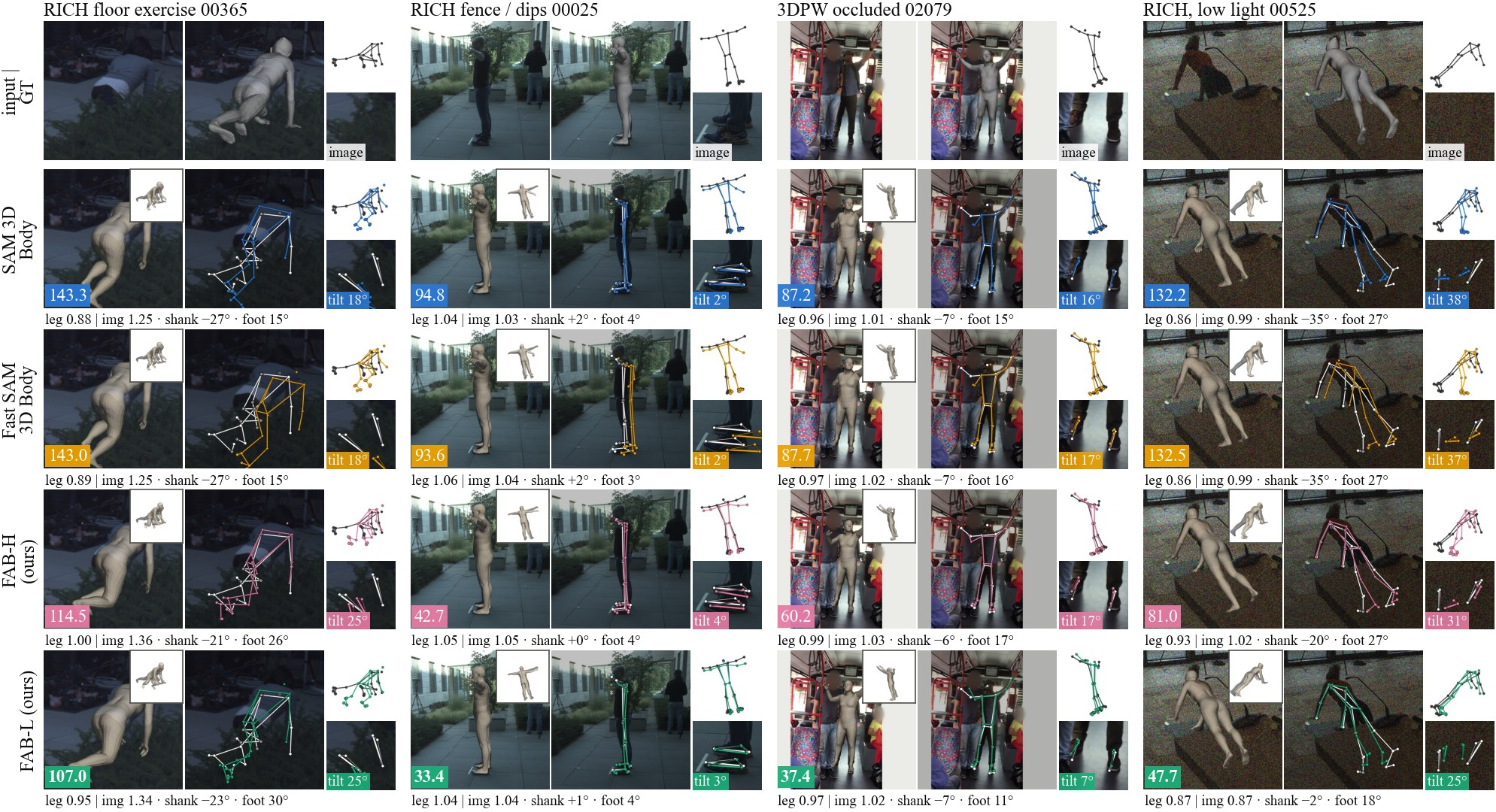}
\caption{Hard frames, one per pool, each picked by rule as the frame with the largest \paper advantage. Each cell: the mesh with its side view (inset; ground truth grey), the keypoints over the ground truth (white), and the 3D keypoints from the side; bottom left: per-frame PA-MPJPE, best in bold. Typical and worst frames: Fig.~\ref{fig:qualtyp}; the numbers below each cell and tilt: Sec.~\ref{sec:suppprotocols}.}
\label{fig:qual}
\end{figure*}
\begin{table}[t]
\centering\tabfont
\caption{More benchmarks under the standard protocol: PA-MPJPE / MPJPE over 24 SMPL joints (mm; our models as in Tab.~\ref{tab:main}) on the test splits (Harmony4D: SAM 3D Body's).}
\label{tab:morebench}
\setlength{\tabcolsep}{3pt}
\begin{tabwrap}
\begin{tabular}{l cc}
\toprule
Method & Harmony4D & EgoBody (egocentric) \\
\midrule
SAM 3D Body & 35.3 / 49.2$^\dagger$ & 40.8 / 58.9 \\
Fast SAM 3D Body & 35.7 / 49.5$^\dagger$ & 41.3 / 59.6 \\
NLF-L & 48.1 / 70.0 & 41.6 / 59.4 \\
\rowcolor{oursbg}
\paper-H, MHR mesh $\to$ SMPL & 37.8 / 54.6$^\dagger$ & 40.3 / 58.6$^\dagger$ \\
\rowcolor{oursbg}
\paper-H, dense points $\to$ SMPL & 36.7 / 49.8$^\dagger$ & 40.4 / 58.7$^\dagger$ \\
\rowcolor{oursbg}
\paper-L (ours) & \textbf{33.2 / 42.1}$^\dagger$ & \textbf{38.8 / 55.6}$^\dagger$ \\
\bottomrule
\end{tabular}
\end{tabwrap}
\tabnotes{$^\dagger$Trains on its training split (EgoBody: images, without its SMPL).}
\end{table}

\begin{table}[t]
\centering\tabfont
\caption{Body shape on SSP-3D (PVE-T-SC: per-vertex error of the neutral-pose body after scale correction, mm).}
\label{tab:shape}
\setlength{\tabcolsep}{4pt}
\begin{tabular}{l c}
\toprule
Method & PVE-T-SC \\
\midrule
SAM 3D Body & 15.8 \\
Fast SAM 3D Body & 15.4 \\
NLF-L & 10.9 \\
\rowcolor{oursbg}
\paper-H, MHR mesh $\to$ SMPL & \textbf{10.0} \\
\rowcolor{oursbg}
\paper-H, dense points $\to$ SMPL & 12.9 \\
\rowcolor{oursbg}
\paper-L, MHR mesh $\to$ SMPL & 10.8 \\
\rowcolor{oursbg}
\paper-L, dense points $\to$ SMPL & 13.3 \\
\bottomrule
\end{tabular}
\end{table}

\begin{table}[t]
\centering\tabfont
\caption{Foot contact F1 against three label sets.}
\label{tab:contactmain}
\setlength{\tabcolsep}{1.9pt}
\begin{tabular}{l cccc}
\toprule
& \multicolumn{2}{c}{static} & mat & HSC \\
\cmidrule(lr){2-3}
Method & EMDB-2 & RICH & MOYO & RICH \\
\midrule
keypoint height / speed rule & 0.46 & 0.62 & 0.55 & 0.65 \\
WHAM (released code) & 0.63 & 0.87 & 0.66 & 0.91 \\
GVHMR (released code) & \textbf{0.71} & \textbf{0.89} & 0.61 & 0.90 \\
\midrule
\rowcolor{oursbg}
ours, image head & 0.67 & 0.88 & \textbf{0.89} & \textbf{0.97} \\

\rowcolor{oursbg}
ours, causal video model (world frame) & 0.69 & \textbf{0.89} & 0.87 & \textbf{0.97} \\
\bottomrule
\end{tabular}
\par\vspace{2pt}{\footnotesize\raggedright Static: world-fixed contact from the ground-truth world motion; mat: MOYO's measured pressure mat; HSC: any contact against RICH's human--scene contact annotations; all on validation splits, which share subjects with training. WHAM and GVHMR output only static contact, scored against every label set. Details: Sec.~\ref{sec:contactlabels}.\par}
\end{table}

\paragraph{Why points first.}
The mesh and the retargeting start from the points, so we test the points themselves. On shape (Tab.~\ref{tab:shape}), each model's MHR mesh $\to$ SMPL fit is closer than its dense points $\to$ SMPL fit, and all four are closer than SAM 3D Body's parametric meshes. The points' gains are in the feet and in 2D: \paper-L's points have legs as long as its mesh's, in 3D and in the image, and feet closer to the ground truth. Against human foot labels on COCO-WholeBody~\cite{jin2020cocowholebody}, the dense head's 2D feet beat SAM 3D Body's (Sec.~\ref{sec:suppcamera}).

\begin{table}[t]
\centering\tabfont
\caption{Robustness on our held-out FAB-Hard benchmark (annotation boxes; MPJPE: 14 joints, mm; 2D: PCK@0.05, \%). More metrics and models, detected boxes, motion blur and low light: Sec.~\ref{sec:suppcamera}.}
\label{tab:robust}
\setlength{\tabcolsep}{2.2pt}
\resizebox{\linewidth}{!}{\begin{tabular}{l cccc cccc}
\toprule
& \multicolumn{4}{c}{MPJPE $\downarrow$} & \multicolumn{4}{c}{2D PCK@0.05 $\uparrow$} \\
\cmidrule(lr){2-5}\cmidrule(lr){6-9}
Method & low light & extreme & acro & occl. & low light & extreme & acro & occl. \\
\midrule
SAM 3D Body & 78.9 & 68.6 & 69.2 & 74.5 & 75.1 & 78.3 & 82.9 & 68.4 \\
Fast SAM 3D Body & 78.6 & 69.7 & 69.8 & 75.2 & 74.6 & 76.1 & 81.1 & 67.8 \\
\rowcolor{oursbg}
\paper-L (ours) & \textbf{63.7} & \textbf{62.9} & \textbf{60.6} & \textbf{66.4} & \textbf{76.6} & \textbf{79.6} & \textbf{85.1} & \textbf{68.9} \\
\bottomrule
\end{tabular}}
\end{table}

\paragraph{Robustness.}
On our held-out FAB-Hard benchmark, \paper-L is the most accurate model in 3D under low light, on acrobatics and martial arts, and under occlusion, and in 2D in every category (Tab.~\ref{tab:robust}; hard frames: Fig.~\ref{fig:qual}). It also degrades least under motion blur; blur, synthetic low light and real low light (ExLPose~\cite{lee2023exlpose}): Sec.~\ref{sec:suppcamera}.

\begin{table*}[t]
\centering\tabfont
\caption{Scenes against three released pipelines, run by us on the same videos, under VideoMimic's protocols where it has them; one measure per column, grouped by dataset (measures, our output per column and clip sets: Sec.~\ref{sec:suppscenes}).}
\label{tab:scene}
\setlength{\tabcolsep}{1.93pt}\scriptsize
\begin{tabwrap}
\begin{tabular}{l ccccccccc ccccc cccccc}
\toprule
 & \multicolumn{9}{c}{scene geometry} & \multicolumn{5}{c}{human in the scene's world} & \multicolumn{6}{c}{human--scene contact} \\
\cmidrule(lr){2-10}\cmidrule(lr){11-15}\cmidrule(lr){16-21}
 & \multicolumn{2}{c}{RICH-3} & \multicolumn{3}{c}{EgoBody} & SLOPER4D & \multicolumn{3}{c}{PROX} & \multicolumn{3}{c}{SLOPER4D} & \multicolumn{2}{c}{EMDB P3\_30} & \multicolumn{4}{c}{RICH-3$^\P$} & \multicolumn{2}{c}{VideoMimic (123), G1} \\
\cmidrule(lr){2-3}\cmidrule(lr){4-6}\cmidrule(lr){7-7}\cmidrule(lr){8-10}\cmidrule(lr){11-13}\cmidrule(lr){14-15}\cmidrule(lr){16-19}\cmidrule(lr){20-21}
Method & rigid & Sim(3) & rigid & Sim(3) & ATE & scene & 1-way & 2-way & non-pen.$\uparrow$ & WA & W & RTE & WA & W & pen. & float & $|s|$ & feet & pen.\ $>$2\,cm & gap \\
\midrule
VideoMimic~\cite{allshire2025videomimic} & 18.4 & 11.7 & 10.9 & 7.6 & 3.0 & \textbf{49.8} & 24.6 & 31.0 & 84.8 & 35.7 & 73.8 & 11.4 & 10.8 & 36.5 & 10.2 & 13.6 & 23.8 & \textbf{4.3} & 57.2 & 5.0 \\
CRISP~\cite{wang2026crisp} & 8.1 & \textbf{6.9} & \textbf{7.6} & \textbf{6.1} & 2.4 & 75.8$^\S$ & \textbf{18.6} & 27.3 & 76.0 & 18.1$^\S$ & 67.2$^\S$ & 3.6$^\S$ & 6.4$^\S$ & 14.9$^\S$ & 7.5 & 11.6 & 19.2 & 6.5 & 0.8$^\ddagger$ & \textbf{1.9}$^\ddagger$ \\
JOSH~\cite{liu2025josh} & 25.3 & 15.2 & 22.2 & 6.9 & 9.9 & 264.6 & 74.5 & 44.2 & 87.3 & 22.2 & 77.7 & 4.6 & 40.8 & 83.2 & 4.9 & \textbf{4.1} & 9.0 & 35.4 & 0.7$^\ddagger$ & 2.6$^\ddagger$ \\
\rowcolor{oursbg}
\paper-L (ours)$^\dagger$ & \textbf{8.0} & 7.5 & 12.5 & 9.4 & \textbf{2.2} & 94.4 & 21.3 & \textbf{17.8} & \textbf{92.0} & \textbf{10.6} & \textbf{40.8} & \textbf{1.1} & \textbf{5.9} & \textbf{14.3} & \textbf{3.5} & 5.4 & \textbf{8.9} & 4.8$^{\ast\ast}$ & \textbf{0.6} & 2.5 \\
\bottomrule
\end{tabular}
\end{tabwrap}
\tabnotes{Units: cm, except non-pen., RTE and pen.\ $>$2\,cm (\%); lower is better except non-pen. $^\S$Overlapping 600-frame chunks. $^{\ast\ast}$After the joint human--scene step (10.5 before it). $^\P$pen.\ / float / $|s|$: camera-frame bodies placed in the scan by the ground-truth camera. $^\dagger$Trains on RICH's and EgoBody's training splits. $^\ddagger$CRISP / JOSH: bodies and scenes through \ifdense FAB-Retarget's mesh mode\else our motion-level retargeter\fi, as ours; JOSH, which outputs no gravity, uses ours.}
\end{table*}

\ifdense\def\tabsevenours{FAB-Retarget [skeleton] / [skeleton + mesh] / [skeleton + mesh + points]: its skeleton / mesh / points mode, pinned in one batch (Sec.~\ref{sec:fabretarget}); }\else\def\tabsevenours{}\fi
\begin{table}[tb]
\centering\tabfont
\caption{G1 references on 25 flat-floor test clips, tracked by SONIC (G1 mode) in one MuJoCo setup, all from the world pipeline's body (Sec.~\ref{sec:supphumanoid}).}
\label{tab:trackersmain}
\label{tab:retargetmain}
\setlength{\tabcolsep}{2.1pt}
\begin{tabular}{l cc c}
\toprule
& \multicolumn{2}{c}{reference} & tracked \\
\cmidrule(lr){2-3}
Reference & bone & slip & falls, bone \\
\midrule
\ifdense
SOMA~\cite{nvidia2026soma} & 15.5 & 0.28 & 5, 23.4 \\
GMR~\cite{araujo2025gmr} & 14.7 & 0.26 & 5, 19.6 \\
UMR~\cite{cao2026umr} & 18.2 & 0.39 & 5, 24.9 \\
\else
SOMA~\cite{nvidia2026soma} & 15.5 & 0.28 & \textbf{5}, 23.4 \\
GMR~\cite{araujo2025gmr} & 14.7 & 0.26 & \textbf{5}, \textbf{19.6} \\
UMR~\cite{cao2026umr} & 18.2 & 0.39 & \textbf{5}, 24.9 \\
\fi
OmniRetarget~\cite{yang2025omniretarget} & 21.7 & 0.31 & 8, 27.4 \\
sequence IK & 24.3 & 0.22 & 6, 25.8 \\
Fast SAM 3D Body & 25.5 & 0.39 & 6, 27.6 \\
\ifdense
\rowcolor{oursbg}
FAB-L G1 head (per frame) & 14.5 & 0.38 & 5, 20.4 \\
\rowcolor{oursbg}
FAB-Retarget [skeleton] & 11.7 & 0.12 & 5, 19.2 \\
\rowcolor{oursbg}
FAB-Retarget [skeleton + mesh] & \textbf{11.6} & \textbf{0.11} & 6, 20.0 \\
\rowcolor{oursbg}
FAB-Retarget [skeleton + mesh + points] & 11.8 & 0.13 & \textbf{4}, \textbf{18.3} \\
\else
\rowcolor{oursbg}
FAB-L G1 head (per frame) & 14.5 & 0.38 & \textbf{5}, 20.4 \\
\rowcolor{oursbg}
ours, motion-level retargeter & 11.6 & \textbf{0.11} & 6, 20.0 \\
\rowcolor{oursbg}
\quad started from SOMA's motion & \textbf{11.5} & 0.12 & 6, 21.6 \\
\fi
\bottomrule
\end{tabular}
\par\vspace{2pt}{\footnotesize\raggedright Reference: bone-direction error ($^\circ$) and sole slip (m\,s$^{-1}$); tracked: falls of 25, bone-direction error before the first fall. FAB-L G1 head (per frame): the in-model G1 head; \tabsevenours Fast SAM 3D Body: its own SMPL through our sequence IK. Details: Sec.~\ref{sec:supphumanoid}.\par}
\end{table}

\begin{figure*}[t]
\centering
\ifdense\includegraphics[width=0.95\linewidth]{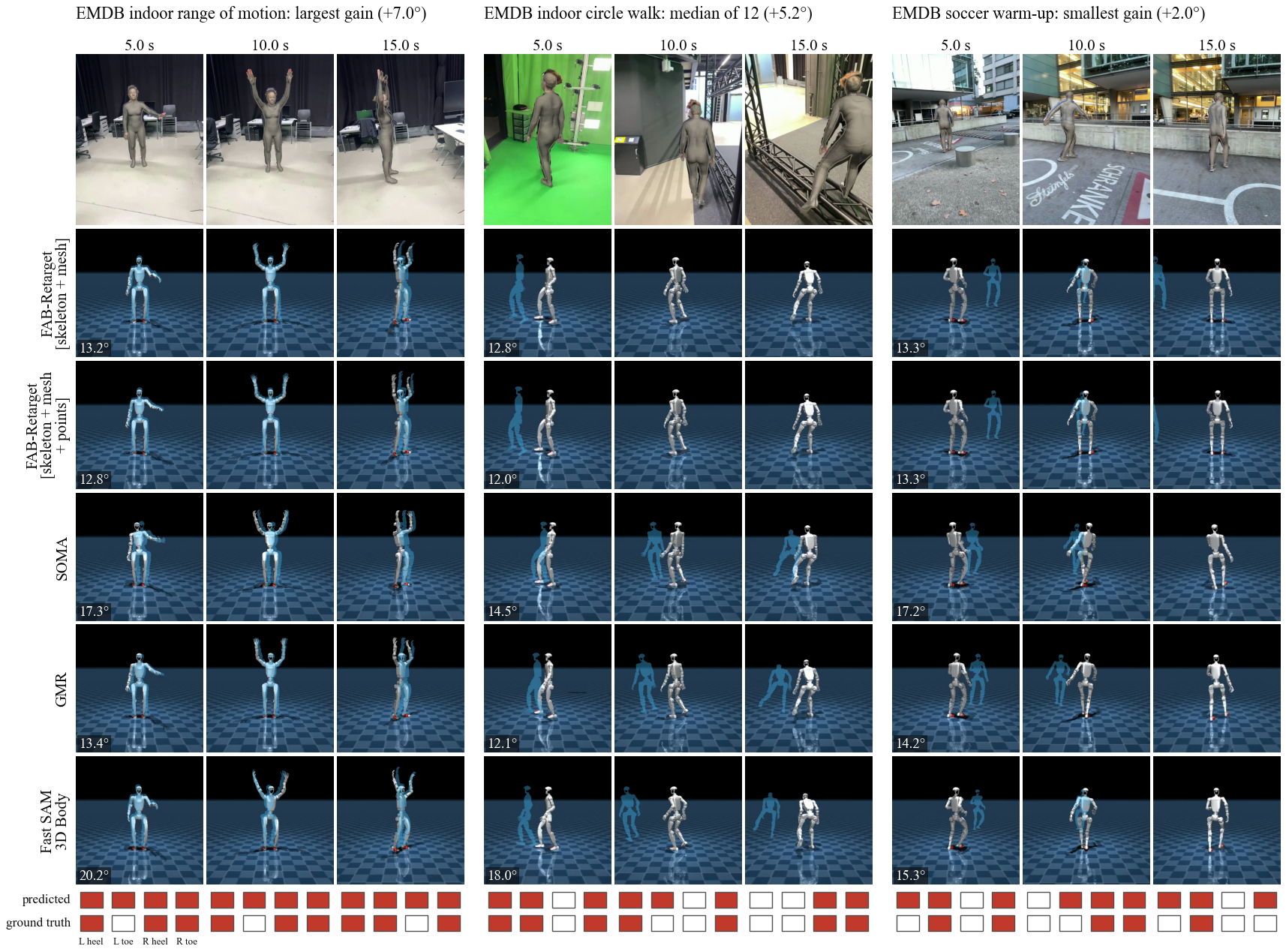}\else\includegraphics[width=0.95\linewidth]{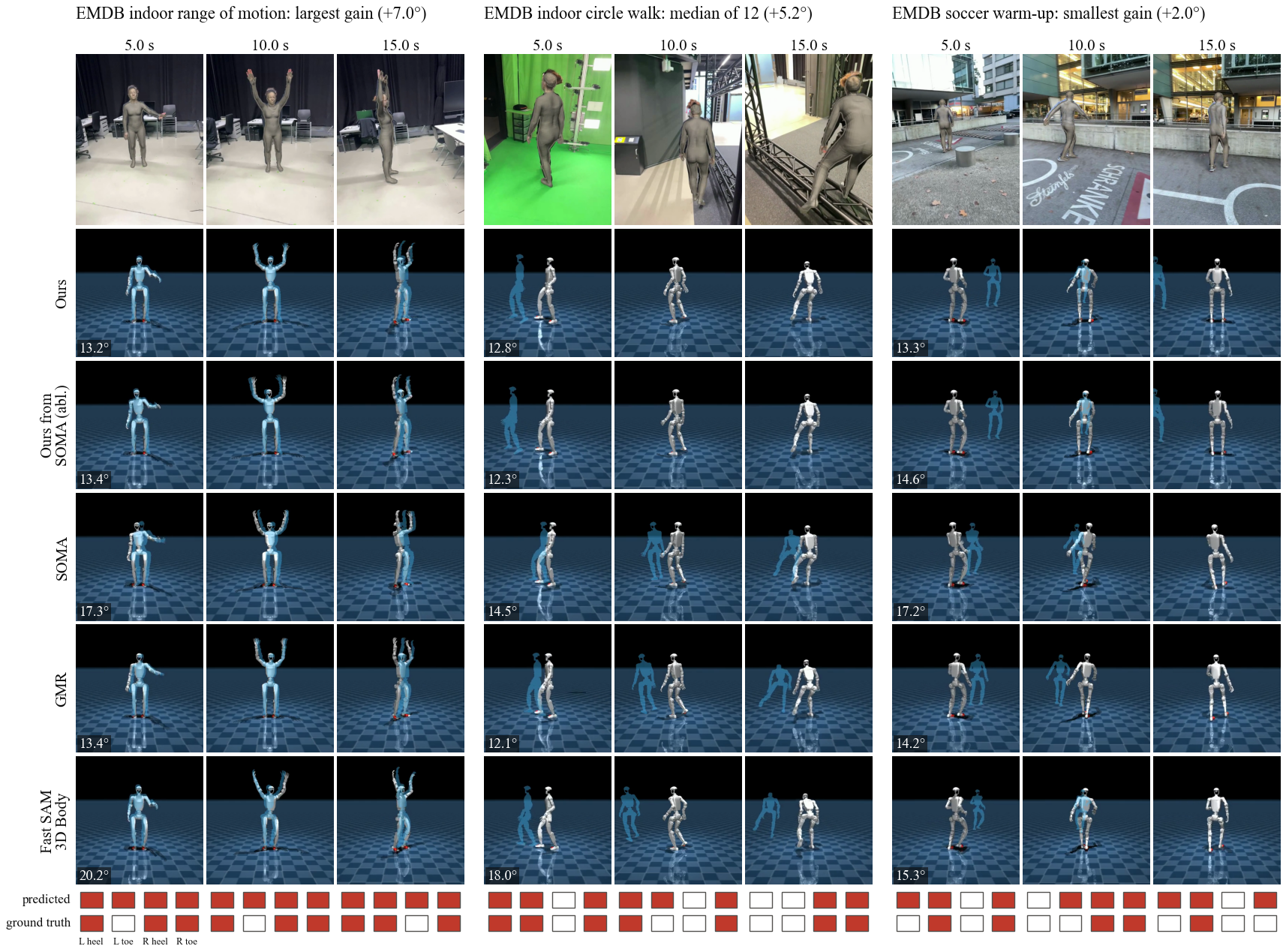}\fi
\caption{Retargeters through SONIC-G1 in our MuJoCo setup (Tab.~\ref{tab:retargetmain}) on \paper-L's world body, and Fast SAM 3D Body's own SMPL through our sequence IK. Ours: \ifdense FAB-Retarget's mesh / points (default) modes\else the motion-level retargeter and its ablation started from SOMA's motion\fi. Top row: input with that world body's SMPL; solid: the SONIC-G1 rollout, which can trail its translucent reference. Numbers: bone-direction error ($^\circ$) to the real person. Bottom rows: feet on the ground (any contact) from the video contact model (predicted; also the red markers on the robots' feet) and from the ground-truth world motion (Sec.~\ref{sec:contactlabels}). Clip picks, title gains, full legend: Sec.~\ref{sec:suppprotocols}.}
\label{fig:g1}
\end{figure*}
\subsection{World}
\label{sec:exp_world}

On EMDB-2, \paper-L is more accurate than every re-run world method (Tab.~\ref{tab:worldmain}). On RICH-191, given the ground-truth boxes like every world method but CRISP, it has the lowest WA-MPJPE, W-MPJPE and RTE; with our own detector it trails DuoMo in WA-MPJPE, and other world methods in W-MPJPE and RTE, because it loses the subject in 23 videos; on the other 168 it remains the most accurate (Tab.~\ref{tab:richsplit}). \paper-L still leads without EMDB-2's skateboard sequence, on which TRAM's, GVHMR's, GEM-X's and WHAM's WA-MPJPE rises to 3--7 times their 25-sequence mean, while DuoMo's, CRISP's and ours rise less, ours least (Sec.~\ref{sec:suppvideoworld}).

The lead comes mainly from the world pipeline: fed the same cameras, boxes and contacts, \paper-H's MHR mesh $\to$ SMPL fit, from our light fine-tune of SAM 3D Body, and both SAM 3D Bodies' per-frame bodies give slightly lower WA- and W-MPJPE than \paper-L's on EMDB-2 and a higher WA-MPJPE on RICH-191. Its joint human--scene optimization, which takes the metric scale from the body, cuts EMDB-2's RTE by about a third; the remaining trajectory error is mostly the camera's (the ground-truth camera motion cuts RTE about fivefold). Without locking any foot, the pipeline keeps every per-frame body's foot sliding well below TRAM's and CRISP's, which also lock no foot (Tab.~\ref{tab:worldmain}); GVHMR, GEM-X and WHAM correct their root or joints with predicted contacts in post-processing.

\subsection{Contacts}
\label{sec:exp_contacts}

On RICH's static contacts the video model is on par with GVHMR's and beats WHAM's; on EMDB-2 it beats WHAM's but trails GVHMR's; on RICH's human--scene contacts and MOYO's mat, both our heads score above both baselines (Tab.~\ref{tab:contactmain}). The world-frame model also separates static from any contact better: on the skateboard sequence, whose feet are world-fixed on only 4\% of the frames, the video model predicts static contact on 31\% of the frames and the image head's static channel on 80\% (Sec.~\ref{sec:suppvideoworld}).

\subsection{Scenes}
\label{sec:exp_scenes}

\noindent\textbf{Geometry.} Per frame, our scene is more accurate than VideoMimic's on RICH-3's static cameras. On EgoBody's egocentric recordings our cameras are more accurate, while VideoMimic's depth is more accurate than our TSDF mesh (Tab.~\ref{tab:scene}). On SLOPER4D's 78\,m walk our scene's Chamfer is about twice VideoMimic's: similar accuracy, much lower completeness.

\noindent\textbf{Contact.} Placed by the ground-truth camera, \paper's camera-frame body lies closer to the scanned contact surfaces than VideoMimic's on all three RICH-3 clips, helped on Gym by the static-camera focal rule. With each method's scene aligned to the scan, the joint optimization brings our feet to within 0.6\,cm of VideoMimic's (Tab.~\ref{tab:scene}).

\noindent\textbf{G1 references.} On VideoMimic's clips, ours penetrate their own terrain far less than VideoMimic's released references, with a smaller contact gap. With its terrain policy on its own meshes, VideoMimic's references track on more clips than ours from either start pose.

\noindent\textbf{Human--scene methods.} CRISP's planar scene is about as accurate as ours per frame on RICH-3 and more accurate on EgoBody's recordings, but its bodies lie farther from the scanned surfaces and its world motion trails ours. JOSH's world motion breaks on SLOPER4D and P3\_30 (Sec.~\ref{sec:suppscenes}).

\subsection{Human-to-Humanoid Retargeting}
\label{sec:exp_humanoid}
\ifdense
In simulation with SONIC's G1 mode (Fig.~\ref{fig:g1}, Tab.~\ref{tab:trackersmain}), the points mode's references are tracked closest to the person's bone directions and fall least (4 of 25), and the per-frame and mesh-mode references track more closely than every baseline but GMR's. Most falls come from the 8 clips outside the trackable 17 (Sec.~\ref{sec:supphumanoid}); on those 17, Fast SAM 3D Body's tracked keypoint error is lower than the per-frame references'.

As references (Tabs.~\ref{tab:retargetmain} and~\ref{tab:retargeters}), the mesh and points modes follow the person's bone directions more closely than every baseline, with less sole slip, jitter and link acceleration than any of them, and are much cleaner than the per-frame G1 head's references. Against the mesh mode (Tab.~\ref{tab:retmodes}), the points mode follows the person's body-part rotations more closely\ifablready{} (slightly more closely still with UMR's learned correspondence in place of ours: Tab.~\ref{tab:retargetinputs})\fi{}, at a slightly higher slip and bone-direction error. With NVIDIA's later SONIC release (v1.1)~\cite{luo2025sonic} in its G1 mode, most references track less closely, and ours still track closer than every baseline (Sec.~\ref{sec:supphumanoid}).
\else
In simulation (Fig.~\ref{fig:g1}), with SONIC's G1 mode our per-frame and motion-level references track more closely than every baseline but GMR's (Tab.~\ref{tab:trackersmain}); on the 17 trackable clips (Sec.~\ref{sec:supphumanoid}), Fast SAM 3D Body's tracked keypoint error is lower than the per-frame references' (103 against 110\,mm).

Among the retargeters, ours follows the person's bone directions more closely than SOMA's, UMR's, OmniRetarget's, GMR's and the sequence IK's (11.6 against 14.7$^\circ$ or more), with the lowest sole slip (Tab.~\ref{tab:retargetmain}), jitter and link acceleration (Tab.~\ref{tab:retargeters}); its references are much cleaner than the per-frame G1 head's (bone 11.6 against 14.5$^\circ$, slip 0.11 against 0.38\,m\,s$^{-1}$, link acceleration 1.38 against 10.99\,m\,s$^{-2}$, jitter 2.4 against 21.0\,rad\,s$^{-2}$), which fall once less with SONIC-G1 (5 against 6). SOMA's references fall once less (5 against 6) but track less closely (23.4 against 20.0$^\circ$). The ablation has a lower reference bone-direction error (11.5 against 11.6$^\circ$) but a higher slip, link acceleration and jitter (0.12 against 0.11\,m\,s$^{-1}$, 1.39 against 1.38\,m\,s$^{-2}$, 2.5 against 2.4\,rad\,s$^{-2}$); it falls as often and tracks less closely (21.6 against 20.0$^\circ$), and needs another retargeter's output to start, so is not part of \paper. Most falls come from 8 hard clips. With NVIDIA's later SONIC release (v1.1)~\cite{luo2025sonic}, most references track less closely, and ours track closer than every baseline (Sec.~\ref{sec:supphumanoid}).
\fi

\paragraph{Tracking on non-flat terrains.} 
We also retarget motions onto non-flat terrain and track them with a terrain-aware policy. 
Our references\ifdense{} from the mesh mode (also Tab.~\ref{tab:scene}'s G1 columns)\fi{} form a G1 terrain dataset of 5{,}330 clips (14.5\,h), 2{,}997 of them cleared for training, each tracked without a fall by VideoMimic's released terrain-aware policy~\cite{allshire2025videomimic} (Sec.~\ref{sec:supphumanoid}).

\FloatBarrier
\section{Conclusion}
\label{sec:conclusion}

We propose \paper for fast yet accurate human pose estimation. Predicting dense points and fitting the body to them gives an estimator above 30\,fps that is more accurate than SAM 3D Body on most benchmarks. The body fitted to the points is the metric anchor of the system: it ties the person to the scene and is retargeted to robot references that barely penetrate their terrain\ifablready
, and retargeting from the points themselves lowers their part-rotation error\fi. \ifdense FAB-Retarget's references are tracked most closely on flat ground in points mode.\else The motion-level retargeter's references penetrate their reconstructed terrain least.\fi

\section{Limitations}
\textbf{Backbone size.} \paper targets fast pose estimation and retargeting, so we favor inference speed over absolute accuracy. Training a larger backbone end to end was beyond our compute: \paper-H trains its dense head on SAM 3D Body's frozen body, so \paper-L is our largest end-to-end model. A larger backbone and more training data are expected to improve accuracy further.

\noindent\textbf{Failure on challenging pose.} We noticed failures on challenging pose, such as yoga or gymnastics. Adding such data to the data engine can potentially improve this. 

\noindent\textbf{No real-robot test.} In simulation, the references of \ifdense\paper-Retarget's default points mode\else the motion-level retargeter\fi{} are tracked more closely than those of the specialized retargeters\ifdense\else{} other than GMR with SONIC's G1 mode\fi, but they have not yet been tested on a physical robot due to lack of resources. 

\noindent\textbf{World-space robustness.} The camera and scene modules we build on can fail: long moving-camera sequences drift (SLOPER4D), a wrong focal length on a static camera puts body and scene at the wrong depth, and one global scale can mis-scale a scene (Sec.~\ref{sec:supplimits}). Training our own camera models is left to future work.

\FloatBarrier
{
    \small
    \bibliographystyle{ieeenat_fullname}
    \bibliography{main}
}

\clearpage
\appendix

\noindent In the appendix tables, where a column is bolded, bold marks its best value (per block where a table has blocks; a caption states any other rule), and shaded rows or columns are ours.

\section{Method details}
\label{sec:suppmethod}

\paragraph{Making SAM 3D Body fast.}
SAM 3D Body~\cite{yang2026sam3dbody} refines MHR~\cite{ferguson2025mhr} parameters layer by layer in a promptable decoder over a $32\times32$ grid of 1280-d tokens, re-projecting its keypoints after every layer; most of its body-path time is the eager MHR rig. Without changing any output, we evaluate the rig only where the 70 keypoints read it at intermediate layers, remove every host synchronisation from the body path so that it is captured as one CUDA graph, store BF16 weights once, use a sparse pose-corrective matrix and compile backbone and decoder. Without the backbone, the decoder path goes from 10.5\,fps per person with the eager stock rig to 19.2\,fps with our rig and 84\,fps as one CUDA graph. The one-pass fit runs at 154\,fps eager (CUDA graph: Sec.~\ref{sec:pixels}).

\paragraph{Accurate model (\paper-H).}
We freeze all but the last 8 blocks of SAM 3D Body's backbone and fine-tune the rest (those blocks, the backbone's final norm, the decoder, the heads, and the prompt and camera encoders) on our mixture (Data, below) with the SMPL-space loss below, the mined hard poses oversampled; EgoBody~\cite{zhang2022egobody} frames enter without their SMPL~\cite{loper2015smpl}. Its dense-point head is trained on the frozen body with \paper-L's dense-head recipe in one run (the same pools, labels, losses and weights; 40K steps, learning rate $5\cdot10^{-4}$, 500 warm-up steps), so its MHR head, its keypoints and the labels it gives the students are unchanged. The 3D point targets are the ground-truth SMPL mesh where a crop has one and elsewhere \paper-H's own MHR mesh, carried into SMPL topology by the barycentric MHR$\to$SMPL transfer, at half weight, with no second teacher; human 2D labels supervise the head's 70 keypoints where they exist.

\paragraph{Fast students (\paper-L, \paper-S).}
The student's backbone is followed by a LayerNorm and a linear projection to SAM 3D Body's 1280-d embedding; \paper-L is trained in one run (Tab.~\ref{tab:recipe}), with \paper-H's labels as the only output-distillation target. The losses are:
\begin{itemize}
\item token-level feature distillation: relative MSE between the projected student tokens and SAM 3D Body's frozen backbone's;
\item output distillation from \paper-H: its 70 keypoints (3D and 2D), a fixed 2048-vertex subset of its MHR mesh, its parameters and its camera translation;
\item human labels where available: the 3D keypoints and MHR fits of SAM 3D Body's training data, a Kabsch-aligned keypoint loss for world-frame labels, and exact SMPL ground truth through an SMPL-space loss computed as the evaluator does (MHR$\to$SMPL barycentric map, 14 H36M~\cite{ionescu2014human36m}-regressor joints, hip-midpoint alignment).
\end{itemize}

\begin{table*}[t]
\centering\tabfont
\caption{\paper-L's training run, from its saved arguments; \paper-S: the same run on DINOv3~\cite{simeoni2025dinov3} ViT-S+/16. Heads, on the frozen body: visibility 16K steps in one run, contact 4K, parts 4K and G1 6K steps.}
\label{tab:recipe}
\setlength{\tabcolsep}{3pt}
\begin{tabular}{l l}
\toprule
init & DINOv3 ViT-L/16; SAM 3D Body's decoder (all 6 layers) and heads \\
data & the 28 pools of the mixture below; FAB-Hard's AthletePose3D~\cite{yeung2025athletepose3d} takes excluded \\
targets & \paper-H's labels (except on crops whose labels come from SAM 3D Body); human labels and ground-truth SMPL \\
& where available; features of SAM 3D Body's frozen backbone \\
schedule & 40K steps, $512\times512$ crops, batch 8 per process; learning rate $5\cdot10^{-4}$ (backbone $5\cdot10^{-5}$, \\
& decoder $2\cdot10^{-5}$), 500 warm-up steps, cosine decay, gradient clip 1.0 \\
losses & all from step 0; the 2.5D-lift terms ramped in linearly over the first 5K steps \\
model & exponential moving average of the run's weights (decay 0.9998) \\
\bottomrule
\end{tabular}
\end{table*}

\paragraph{Dense points and parts.}
Points within 2\% of the crop border keep a directly regressed 3D position: their heatmaps are clamped at the crop edge, so the 2D ray is not observed; the same rule holds in training and inference. The 70 image-defined MHR keypoints are supervised by human 2D labels (strongly), \paper-H's 2D (weakly) and its 3D. The 14 evaluation joints and 468 SMPL template vertices (farthest-point sampled, closed under the left--right mirror map) are supervised by SMPL ground truth where available and otherwise by \paper-H's mesh in SMPL topology at half weight; flipped crops use the mirrored target. Reprojection terms place the predicted 3D points at the target's root depth. Part labels (background and 14 parts from the SMPL skinning weights) are rendered in the trainer from the target mesh with a z-buffer, so flips, rolls and crop jitter stay consistent. When no parameters are needed, a fixed linear map gives all vertices from the points. The learned fit reconstructs held-out bodies to 8.5\,mm per vertex from surface points with 1\,cm noise and runs as one CUDA graph at 1{,}020\,fps per person; the fit's global orientation is taken from the network's MHR mesh head, and pose, shape, placement and the feet come from the points. The contact head lowers the one-person rate by 1.4\,fps, and the causal contact model runs on its own at 4{,}618\,fps.

\begin{figure*}[t]
\centering
\includegraphics[width=\linewidth]{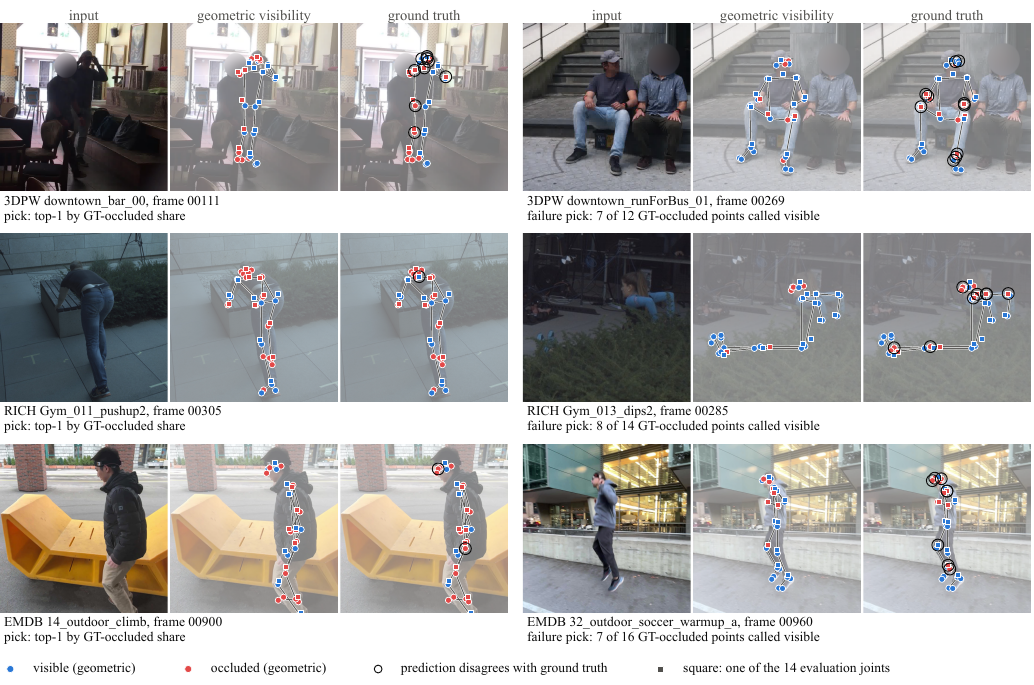}
\caption{The geometric visibility output on occluded people. Rows: 3DPW test people at least 25\,\% covered by a nearer person, RICH test and EMDB-1. Among the people with at least 25 of the 37 labelled points in the image, left: per row, the one with the largest share of ground-truth-occluded points among their in-image labelled points (ties by name; on RICH, the frame of Fig.~\ref{fig:vis}); right: a failure, the person with the most ground-truth-occluded points the geometric output calls visible (7 of 12, 8 of 14 and 7 of 16), from a sequence that none of the pool's other picks (its five largest-share and two random ones) uses; ties go to the larger occluded share, then by name. Per example: input, the 37 labelled points (circles: body keypoints, squares: the 14 joints) coloured by the geometric output, and the ground-truth classes from the mesh's z-buffer (self-occlusion and other people, no object occlusion), ringed where the output disagrees. Faces other than the subject's are blurred.}
\label{fig:visgrid}
\end{figure*}

\paragraph{Visibility.}\label{sec:suppvis}
The visibility head (0.92\,M parameters) reads, per point, the dense head's features at the point, its dense 2D and 3D, a learned point embedding, the decoder's pose token and a position-aware attention to the image tokens, and gives two sets of three-class logits (visible, occluded, outside the crop) for the dense head's 70 keypoints and 14 joints; the 468 surface points have none. The geometric classifier is trained on labels rendered from the crop's target mesh with a z-buffer: a point is occluded when its pixel shows none of its own body parts within a per-keypoint depth tolerance (3--40\,cm), when another person's mesh lies more than 5\,cm in front, or when the bodies call it visible but it falls outside the person's SAM~3 mask; a point outside the training crop is labelled outside, and points the labeller could not judge (inside the image but outside its crop, close-ups without 2D support) are ignored. The annotation-style classifier is trained on human annotators' flags (COCO visible / occluded, COCO-WholeBody and our labelled feet). Both train on the frozen \paper-L body with cross-entropy, unknown labels ignored and the occluded class up-weighted (1.5 geometric, 2 annotation-style; schedule: Tab.~\ref{tab:recipe}). The fit reads the annotation-style output in one place: the foot refinement turns each foot onto the dense 2D heel-to-toe direction in proportion to the lowest visible probability of that foot's heel and big and small toes (zero when any of them lies outside the crop by its 2\,\% margin or the heel-to-toes span is under 6\,px; a foot more than 60$^\circ$ off the dense 2D direction is left unturned). Scored against labels rendered from the ground-truth mesh (self-occlusion and other people; no object occlusion; ground-truth boxes) on the 14 joints of 3DPW, RICH and EMDB-1 (the in-image ones of 8{,}879, 5{,}628 and 12{,}059 people, with 122{,}754, 78{,}762 and 168{,}354 in-image points), the geometric output reaches an occluded AP of 80.8--85.2 and a three-class accuracy of 90.9--93.0\,\% on the three sets, far above a per-point prior in occluded AP and above it in accuracy (examples and failures: Fig.~\ref{fig:visgrid}). The annotation-style output is trained for a different task, the annotators' convention. Scored against the same geometric labels, it reaches 47.2--60.7 occluded AP and 83.6--87.2\,\% accuracy, and it calls 81.1--92.4\,\% of the self-occluded body keypoints visible, about eight to eleven times the geometric output's share.

\paragraph{In-model retargeting.}
The IK of Sec.~\ref{sec:humanoid} is direction-first and per frame, with URDF forward kinematics in PyTorch: the G1's link-bone directions and pelvis frame follow the human's, and its pelvis-relative link positions follow the SMPL joints, scaled by one human-to-robot factor per sample, as a weak term. Links are matched to SMPL joints by the SMPL-H~\cite{romero2017mano}$\to$G1 joint correspondence of Holosoma~\cite{amazonfar2025holosoma}, and a sigmoid parameterisation keeps every joint within its limits; on BEDLAM's ground truth the links lie 6.3\,cm from their scaled SMPL joints on average. The head also regresses the root orientation (6D, camera frame) and trains with forward-kinematics link-position and joint-angle losses.

\ifdense\def\denserowpts{dense points & 1{,}000 & part rotations at the G1 anchors (points mode only) \\}\def\denserowsole{}\else\def\denserowpts{}\def\denserowsole{}\fi
\begin{table*}[t]
\centering\tabfont
\caption{Objective of the motion-level retargeter: weights on squared residuals against the human motion, at one human$\to$G1 scale per clip.}
\label{tab:retargetobj}
\setlength{\tabcolsep}{3pt}
\begin{tabular}{l r l}
\toprule
Term & Weight & Residual \\
\midrule
directions & 1 & limb bones: legs, arms, feet, pelvis--shoulders \\
pelvis orientation & 1 & against the human pelvis frame \\
positions & 0.1 & pelvis-relative links (end effectors $\times$2) \\
\denserowpts
velocities & 0.05 / 0.05 & hands, ankles, toes / pelvis \\
contact & 50 & labelled feet on the floor or terrain \\
skate & 50 & no slide within a contact segment \\
\denserowsole
floor & 1{,}000 & no support point below the floor or terrain \\
link clearance & 1{,}000 & no link closer than 4\,cm above the terrain (a foot exempt while in contact) \\
seated pelvis & 50 & labelled seat on its support \\
self-collision & 1{,}000 & arm--body, leg--leg and hand--hand capsules, 1\,cm margin \\
acceleration & 120 & joint and root acceleration \\
velocity limit & 0.01 & hinge above 0.8$\times$ URDF joint speed \\
root priors & 0.01 / 0.1 & scaled human pelvis path / height \\
\bottomrule
\end{tabular}
\end{table*}

\paragraph{Motion-level retargeting.}
One configuration (Tab.~\ref{tab:retargetobj}) runs on every clip; the 25 test clips are not held out from the retargeter's design.
\begin{itemize}
\item \textbf{Contacts and seat.} A foot is in contact when its lowest vertex is within 4\,cm of the floor or terrain and slower than 0.15\,m\,s$^{-1}$; hands and knees are not contact targets and keep the floor and clearance terms. The pelvis is seated when the buttocks are within 5\,cm of a support, slower than 0.5\,m\,s$^{-1}$, with the pelvis less than 25\,cm above it.\ifdense{} In skeleton mode (Sec.~\ref{sec:fabretarget}), a foot's test takes the lowest of the ankle, the toe base and a toe tip carried by the leg's joints, each lowered by its height above the sole in the SMPL template (scaled to the input's leg length), with the ankle--toe-base midpoint's ground-plane speed; its foot labels agree with the mesh rule's at a median F1 of 0.96 / 0.95 (left / right foot) over the 24 test clips with foot contact.\fi
\item \textbf{Smoothing.} The input is smoothed only by the world pipeline (Sec.~\ref{sec:world}); the retargeter smooths the joint positions it fits with a zero-phase Gaussian of 8 frames at 50\,fps (0.16\,s), the same for every input; every other output is per frame.
\item \textbf{Optimization.} Adam~\cite{kingma2015adam} from the sequence IK (Sec.~\ref{sec:suppsettings}), with root priors from the scaled human pelvis. The start height adds, per frame, how far the human's deeper foot lies below the terrain under its center while the knee is above that surface (one-sided, smoothed like the input). The human pelvis frame's up axis is turned forward about the hip axis by the SMPL template's back-lean before the pelvis-orientation term compares against it. Terrain gradients come from a blurred heightfield, which removes single-frame root spikes at scan risers (EgoBody clips: 1{,}017 $\to$ 4). The SOMA-initialized ablation runs the same objective from SOMA's~\cite{nvidia2026soma} solution.
\item \textbf{Stance pinning} fixes each stance foot over the stillest part of its contact run and re-solves legs and root with a sole-footprint term, a small lift and a cap on the root step (Sec.~\ref{sec:suppsettings}).
\end{itemize}

\ifdense\else\ifablready
\paragraph{Dense-point ablation.}
The ablation of Tab.~\ref{tab:retargetinputs} adds two \paper outputs to the skeleton, the dense points and the predicted contacts. \densemethod{} The video model's heel / toe contacts pin the feet and flatten the sole when both touch.
\fi\fi

\ifdenseshown
\paragraph{Dense-point term.}
\ifdense\densemethod{} \fi Groups: pelvis, chest and, per side, upper arm, forearm, hand, thigh, calf and foot (the head points are unused: the G1 head is fixed to the torso). $R^{*}_{g}$ is the Kabsch rotation of the group's points from the T-pose, left-multiplied by the smallest rotation that turns the group's G1 bone onto the skeleton's bone (none for the pelvis) and composed with the link's T-pose orientation; $a_i$ are the anchors in the link's frame. \ifablready In the contact rows of the retargeter-input ablation (Sec.~\ref{sec:supphumanoid}), a heel or toe is pinned when the video model's static probability is at least 0.5 and the point moves slower than 0.15\,m\,s$^{-1}$, and touches when its any-contact probability is at least 0.5 at the same speed, both with gaps under 0.1\,s filled and runs under 0.2\,s dropped; pinned points enter the contact, skate and stance-pin terms, and while heel and toe both touch the sole's normal is turned onto the terrain's. \fi{}The dense points are registered per frame onto the world SMPL body at the same template vertices by a similarity, and their residual to it is smoothed like the body.

\ifdenseshown
\begin{figure*}[t]
\centering
\includegraphics[width=\linewidth]{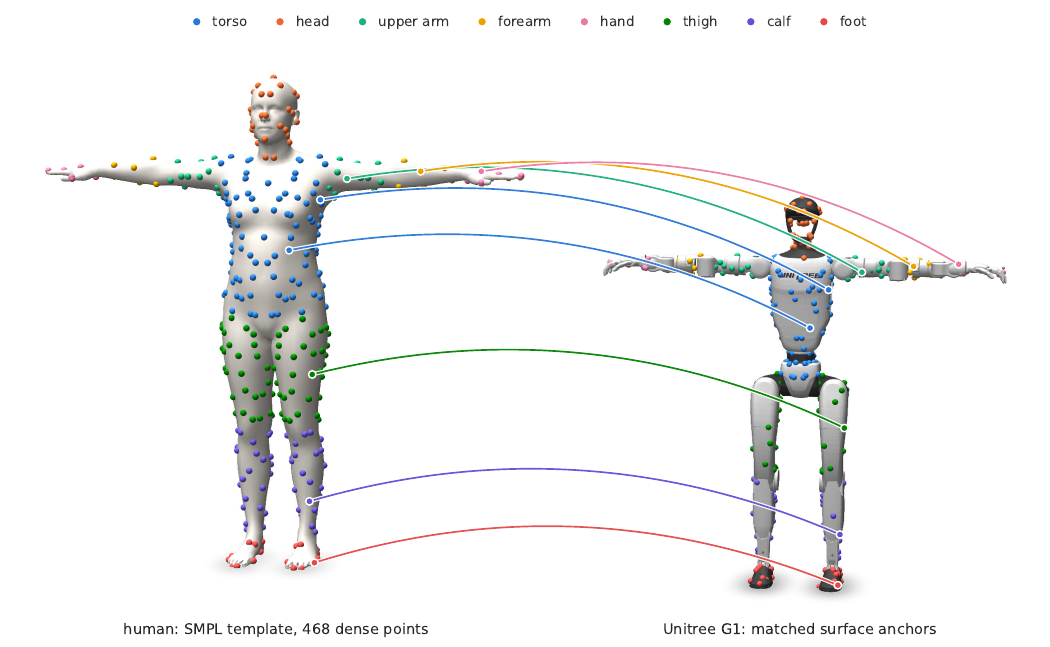}
\caption{Dense-point correspondence. Left: \paper's 468 dense points on the SMPL template in its T-pose. Right: their surface anchors on the Unitree G1 in the matched T-pose, at metric scale. Colours give the body part (this figure's own palette); lines join a few matched pairs. Each point is matched within its part by its cylindrical coordinates $(s,\theta)$ about the part's bone.}
\label{fig:corr}
\end{figure*}
\fi
\fi

\paragraph{Data.}
The mixture:
\begin{itemize}
\item SAM 3D Body's released training annotations for COCO~\cite{lin2014coco}, MPII~\cite{andriluka2014mpii}, 3DPW, AthletePose3D~\cite{yeung2025athletepose3d} and AIST++~\cite{li2021aistpp}, with COCO-WholeBody's~\cite{jin2020cocowholebody} human 2D for body and hands where present; COCO-2014 and AGORA~\cite{patel2021agora} images with YOLO11~\cite{jocher2024yolo11} boxes; ImageNet~\cite{deng2009imagenet} and SA-1B~\cite{kirillov2023sam} person images; Human-Art's~\cite{ju2023humanart} training images (12.5K crops, human 2D).
\item BEDLAM (875K crops, exact SMPL) and 3DPW-train (gendered SMPL in the camera frame).
\item Harmony4D~\cite{khirodkar2024harmony4d} (298K crops), with its multi-view SMPL fits in each exo camera's frame, undistorted as in SAM 3D Body; fits that drift from the triangulated keypoints are dropped, and the \paper-H terms are off where \paper-H contradicts the ground-truth mesh.
\item MOYO and RICH with SMPL in each camera's frame, EgoBody's exo and egocentric images without its SMPL (its fits have short legs relative to the torso), and Embody 3D~\cite{mclean2025embody3d} subsets.
\item VideoSphere, our web-video frames, with \paper-H's labels (all frames kept).
\item Hard poses mined from all pools (torso more than $60^\circ$ from upright, feet above the hips, or the top 5\% of limb spread per source).
\end{itemize}
Foot losses are off on Harmony4D, whose ground-truth heel-to-toe direction differs from two image models' 2D feet by more than 20$^\circ$ on 21\% of feet (their noise floor on COCO-WholeBody: 11\%), and on EgoBody's egocentric images (within that floor, 8.0\%, but sharing EgoBody's fits, which stand toes-up by about 2.5\,cm on static standing frames); Human-Art's foot 2D labels are not used. 211K hard crops, mined from every pool by per-crop error, enter the mixture. A ground-truth 2D check drops a crop when ViTPose~\cite{xu2022vitpose} finds fewer than 6 joints or its projected joints lie more than 8\% of the crop from both the 2D reference and \paper-H. Public datasets are used under their licenses. The one in-house clip shows a consenting team member and was recorded with informed consent; no IRB review was required for this non-interventional recording. The VideoSphere frames come from publicly available web videos; they are used only to train the model, are not redistributed or shown in the paper, and no person in them is identified, so individual consent was not sought.

\paragraph{Video and world space.}
\begin{itemize}
\item \textbf{Video crops.} The person detector runs every 10th frame; in between, each crop is the box of the person's previous-frame keypoints, slightly enlarged, with tracks matched by box IoU (Sec.~\ref{sec:suppsettings}). The subject is the detection that overlaps the previous subject box, otherwise the tallest person (the largest projected keypoint height). Every world and scene result runs this detector and subject selection, with no dataset boxes, except Tab.~\ref{tab:worldmain}'s ground-truth-box row and, for identity only, two multi-person 3DPW G1 clips (Sec.~\ref{sec:supphumanoid}).
\item \textbf{Gravity.} GeoCalib~\cite{veicht2024geocalib} on every 10th frame (0.9$^\circ$ mean error on all 25 EMDB-2~\cite{kaufmann2023emdb} sequences; 0.3$^\circ$ without P0\_09, whose ground-truth world is tilted about 12$^\circ$ from the gravity its walls imply).
\item \textbf{Focal length.} SLAM observes the focal only through parallax. A clip is static when the 90th percentile of its camera centres' distance from their median is below 0.5\% of the median static scene depth (static cameras stay below 0.04\% on the robustness set and reach 0.22\% on our static evaluation clips; the slowest moving camera reaches 1.1\%); static clips take the median of ViPE's~\cite{huang2025vipe}, GeoCalib's and MoGe-2's~\cite{wang2025moge2} focal (the latter two over 12 frames) and re-run ViPE with it fixed. The rule lowers the static-camera focal error from 17.3 to 7.1\% on 21 robustness clips but only from 18.2 to 15.6\% on our 15 static evaluation clips, and it can hurt (3DPW fencing: 0.94 $\to$ 0.82 of the true focal).
\item \textbf{Person detector.} A one-class YOLO11m, trained with rotated copies (tight boxes of the rotated masks), exhaustively labelled MOYO, RICH and EgoBody frames and SAM~3~\cite{carion2025sam3} labels on 45K hard images (crowds, small, truncated, blurred and bystander people) with low-light, blurred and rotated copies, kept where SAM~3 agrees with the ground truth or the detector or after a visual check; SAM~3's partial duplicates are removed with its masks, and confident detector boxes SAM~3 rejects stay as hard negatives. Tab.~\ref{tab:detector} compares it.
\end{itemize}

\paragraph{Entry points.}
\texttt{fab}, \texttt{fab-video}, \texttt{fab-world} and \texttt{fab-retarget} run the models with the paper's settings as their defaults (\texttt{fab-build-trt} builds the TensorRT engines); the benchmarks add only their protocols' boxes or intrinsics.
\begin{itemize}
\item \texttt{fab} (an image or an image folder; \paper-L by default, \paper-S or \paper-H by option): per person and frame, the 2D and 3D keypoints, the dense surface points, the MHR mesh, the SMPL fitted to the points, the heel / toe / hand / knee contacts, the body parts and the G1 joint angles, in the camera frame.
\item \texttt{fab-video} (a video): the same outputs per track, with causal smoothing, the video contact model and an overlay render.
\item \texttt{fab-world} (a video): the world body in a metric, gravity-aligned frame, with the cameras, the video contacts, the scene mesh and its heightfield, through the world and scene pipeline of Sec.~\ref{sec:world}.
\item \texttt{fab-retarget} (a \texttt{fab-world} output): the G1 reference from the motion-level retargeter on that scene's terrain.
\end{itemize}

\begin{table*}[t]
\centering\tabfont
\caption{Person detectors (person AP; recall at IoU 0.5; false positives: share of person-free images with a detection). The YOLO11m models run at the same speed; SAM~3 is 38$\times$ slower.}
\label{tab:detector}
\setlength{\tabcolsep}{2.5pt}
\begin{tabular}{l cccccc}
\toprule
Detector & COCO AP & AP, 180$^\circ$ & crowd recall & false pos. & 3DPW-dark & MOYO val \\
\midrule
YOLO11m (stock) & 62.1 & 27.1 & 67.5\% & \textbf{3.6\%} & 91.2\% & 95.2\% \\
SAM~3 & \textbf{65.6} & 51.6 & \textbf{91.6\%} & 11.3\% & \textbf{99.1\%} & \textbf{100\%} \\
\rowcolor{oursbg}
ours & 59.3 & \textbf{56.2} & 74.8\% & 10.9\% & 97.9\% & \textbf{100\%} \\
\bottomrule
\end{tabular}
\end{table*}

\section{Benchmarks and protocols}
\label{sec:benchmarks}

\paragraph{Curated benchmarks.}
\begin{itemize}
\item \textbf{Low light.} Every 3DPW and EMDB-1 frame goes through a deterministic sensor model, seeded per sample: unprocessing to linear raw, 10--50$\times$ under-exposure, Poisson-Gaussian noise and re-processing to sRGB; ground truth is unchanged.
\item \textbf{Hard poses.} The EMDB-1 handstand, cartwheel, indoor dancing, big stairs down and outdoor workout sequences: 6{,}644 frames with large torso tilts, inversions and fast limb motion.
\item \textbf{FAB-Hard}, our held-out benchmark: 1{,}721 items with 3D ground truth and 1{,}114 with human 2D labels, selected by data properties only (light level, torso tilt, sport, occlusion), never by any method's error, and without AthletePose3D, which is in SAM 3D Body's training data. Low light: synthetic under-exposure of held-out frames from 5 sources; extreme pose: held-out MOYO sequences, RICH test and EMDB; acrobatics / martial arts: Harmony4D test (3D) and LSPET~\cite{johnson2011lspet} (2D); occlusion: held-out multi-person and COCO crops.
\end{itemize}

\paragraph{Model selection.}
Camera-space settings are chosen on 29 EMDB sequences outside EMDB-1 (35{,}589 frames; on them the PA-MPJPE ranking of Tab.~\ref{tab:emdb1}'s methods matches EMDB-1's), MOYO's validation sessions (7{,}765 crops, 20\% with the torso past 90$^\circ$) and the EgoBody and RICH validation splits.

\section{Additional results}
\label{sec:suppresults}

\subsection{Protocols of the tables and figures in Secs.~\ref{sec:method}--\ref{sec:experiments}}
\label{sec:suppprotocols}
\paragraph{Protocol summary.}
All camera-space methods use the same 3DPW and RICH frames, ground-truth boxes and intrinsics; joint sets follow SAM 3D Body. \paper-L and \paper-S use one fixed inference path. \paper-H is reported both through its MHR mesh and through its dense points. Ours and NLF-L are timed end to end with our TensorRT detector; published systems otherwise use their released inference paths. Detailed commands, crop conventions, model revisions and conversion maps are released with the code.

\paragraph{\paper-H's two routes.}
The two routes compare as follows: \paper-H's MHR mesh $\to$ SMPL fit is closer on shape and on most benchmarks, its dense points $\to$ SMPL fit on Harmony4D and, in MPJPE and PVE with annotation boxes, in every FAB-Hard category (Tabs.~\ref{tab:shape}, \ref{tab:morebench} and~\ref{tab:fabhard}). Under the bf16 autocast both models are scored with, \paper-H's dense points are less stable than \paper-L's: a $\pm$1 grey-level change of the image moves \paper-H's largest point by 63.2\,mm on 3DPW and 60.0\,mm on RICH (median over 100 frames each), three to four times as far as \paper-L's, and the fit to them follows. In fp32 the same change moves \paper-H's largest point by 3.8\,mm on 3DPW and 7.4\,mm on RICH.

\paragraph{Tabs.~\ref{tab:morebench} and~\ref{tab:shape}.}
Standard protocol. Harmony4D: SAM 3D Body's 39-sequence test split (11{,}190 people); EgoBody: the egocentric test recordings. SSP-3D~\cite{sengupta2020ssp3d}: 311 images, PVE-T-SC as in the dataset's evaluation; \paper-L and \paper-H fit SMPL with free shape to their own surface points, or through the one-pass fitter to their MHR mesh (as in Tab.~\ref{tab:main}).

\paragraph{Figures.}
Fig.~\ref{fig:g1} shows representative clips chosen by the largest, median and smallest gain over Fast SAM 3D Body among trackable sequences; Fig.~\ref{fig:qual} shows hard examples selected by the largest joint-error gain. Complete selection rules and per-frame measurements accompany the release.

\subsection{Latency}
\label{sec:latency}
Rates are end-to-end on one RTX 4070 Ti SUPER (16,GB), including detection, cropping, the body path, dense points, SMPL fitting and auxiliary heads. We time warm-up-stabilized 3DPW validation and COCO frames across one to four detected people. Ours and image baselines use matched sessions; video methods use their released chains on 300-frame clips. TensorRT preserves \paper-L and \paper-S accuracy to below 0.001,mm and matches the PyTorch detector within 0.02 COCO person AP.

\begin{table}[t]
\centering\tabfont
\caption{End-to-end fps by the number of people in the frame; p95: the 95th-percentile frame time as fps; $<$30: share of one-person frames slower than 30\,fps. Body: TensorRT for \paper-L and \paper-S, compiled PyTorch for \paper-H.}
\label{tab:fps}
\setlength{\tabcolsep}{2.85pt}
\begin{tabular}{l cccc cc}
\toprule
Model, person detector & 1 & 2 & 3 & 4 & p95 & $<$30 \\
\midrule
\rowcolor{oursbg}
\paper-L, TensorRT & 39.2 & 25.7 & 19.6 & 15.4 & 35.0 & \textbf{0.0\%} \\
\rowcolor{oursbg}
\paper-L, PyTorch & 32.6 & 22.9 & 17.8 & 14.2 & 29.0 & 8.8\% \\
\rowcolor{oursbg}
\paper-S, TensorRT & \textbf{63.8} & \textbf{48.6} & \textbf{41.3} & \textbf{33.8} & \textbf{48.7} & \textbf{0.0\%} \\
\rowcolor{oursbg}
\paper-S, PyTorch & 47.2 & 37.8 & 33.4 & 28.2 & 34.0 & 0.6\% \\
\rowcolor{oursbg}
\paper-H, MHR mesh $\to$ SMPL & 24.9 & 14.8 & 10.2 & 7.8 & 24.3 & 100\% \\
\rowcolor{oursbg}
\paper-H, dense points $\to$ SMPL & 21.8 & 13.1 & 9.4 & 7.4 & 20.7 & 100\% \\
\bottomrule
\end{tabular}
\end{table}

\subsection{Camera space and robustness}
\label{sec:suppcamera}

\begin{table*}[t]
\centering\tabfont
\caption{FAB-Hard (Sec.~\ref{sec:benchmarks}): PA-MPJPE / MPJPE / PVE (the per-vertex error; 14 joints, mm) and 2D PCK@0.05 (the share of labelled keypoints within 0.05 of the ground-truth box's longer side, \%). Our models as in Tab.~\ref{tab:main} (2D: the dense heads, and \paper-H's MHR head for its MHR mesh $\to$ SMPL row); other MHR models: the official converter. Bottom: our person detector's boxes instead of the annotation boxes (only the item's own person is scored, so compare within a block).}
\label{tab:fabhard}
\setlength{\tabcolsep}{2.5pt}
\begin{tabular}{l cccc cccc}
\toprule
& \multicolumn{4}{c}{3D} & \multicolumn{4}{c}{2D PCK@0.05} \\
\cmidrule(lr){2-5}\cmidrule(lr){6-9}
Method & low light & extreme pose & acro / martial & occlusion & low light & extreme & acro & occl. \\
\midrule
SAM 3D Body & 47.1 / 78.9 / 90.7 & 44.1 / 68.6 / 76.3 & 42.5 / 69.2 / 82.7 & 43.0 / 74.5 / 84.8 & 75.1 & 78.3 & 82.9 & 68.4 \\
Fast SAM 3D Body & 47.3 / 78.6 / 90.9 & 44.9 / 69.7 / 77.4 & 42.8 / 69.8 / 83.7 & 43.5 / 75.2 / 85.7 & 74.6 & 76.1 & 81.1 & 67.8 \\
NLF-L & 57.1 / 93.3 / 108.1 & \textbf{32.9 / 52.8 / 60.6} & 56.0 / 101.3 / 117.1 & 53.6 / 86.4 / 102.1 & 62.8 & 68.5 & 77.4 & 53.8 \\
\rowcolor{oursbg}
\paper-H, MHR mesh $\to$ SMPL & 45.5 / 73.3 / 84.2 & 42.7 / 64.9 / 74.9 & 46.4 / 76.9 / 90.5 & 46.7 / 85.5 / 97.7 & 72.8 & 77.6 & 80.7 & 66.2 \\
\rowcolor{oursbg}
\paper-H, dense points $\to$ SMPL & 46.3 / 73.1 / 83.7 & 42.5 / 64.7 / 74.1 & 45.8 / 69.7 / 82.4 & 47.8 / 82.6 / 94.5 & 73.9 & 78.5 & 83.8 & 63.7 \\
\rowcolor{oursbg}
\paper-L (ours) & \textbf{39.8 / 63.7 / 73.2} & 36.3 / 62.9 / 71.5 & \textbf{42.4 / 60.6 / 71.7} & \textbf{41.3 / 66.4 / 77.0} & \textbf{76.6} & \textbf{79.6} & \textbf{85.1} & \textbf{68.9} \\
\midrule
\multicolumn{9}{l}{\emph{detected boxes (our person detector)}} \\
\emph{detector recall (\%)} & 91.4 & 99.4 & 95.8 & 99.6 & 90.9 & 97.6 & 97.5 & 93.1 \\
SAM 3D Body & 42.9 / 73.5 / 84.6 & 43.6 / 67.2 / 75.1 & 57.7 / 125.8 / 143.9 & \textbf{52.9} / 111.0 / 127.2 & 76.4 & 76.2 & 83.2 & 66.4 \\
Fast SAM 3D Body & 43.3 / 74.1 / 85.8 & 44.4 / 68.3 / 76.1 & 58.1 / 126.4 / 144.7 & 53.4 / 111.4 / 127.9 & 75.9 & 74.3 & 81.7 & 66.0 \\
NLF-L & 51.6 / 85.8 / 99.3 & \textbf{32.7 / 52.7 / 60.6} & 65.6 / 142.0 / 161.3 & 60.7 / 114.4 / 134.1 & 65.3 & 68.8 & 77.2 & 53.2 \\
\rowcolor{oursbg}
\paper-H, MHR mesh $\to$ SMPL & 41.1 / 69.6 / 80.0 & 42.3 / 64.2 / 74.3 & 59.1 / 128.2 / 145.8 & 54.1 / 111.4 / 128.3 & 74.4 & 74.8 & 80.9 & 65.0 \\
\rowcolor{oursbg}
\paper-H, dense points $\to$ SMPL & 41.9 / 70.0 / 79.9 & 42.2 / 63.9 / 73.4 & 58.4 / 121.7 / 139.1 & 55.3 / 108.8 / 125.6 & 75.0 & 77.5 & 84.0 & 62.6 \\
\rowcolor{oursbg}
\paper-L (ours) & \textbf{39.2 / 68.3 / 78.7} & 36.0 / 61.3 / 69.9 & \textbf{55.9 / 110.8 / 127.0} & 53.1 / \textbf{107.9 / 124.8} & \textbf{76.7} & \textbf{78.0} & \textbf{85.3} & \textbf{66.5} \\
\bottomrule
\end{tabular}
\end{table*}

\begin{table*}[t]
\centering\tabfont
\caption{Hard poses and low light (Sec.~\ref{sec:benchmarks}; PA-MPJPE / MPJPE / PVE, mm; ground-truth boxes, ground-truth intrinsics, 14 H36M joints, both SAM 3D Bodies through the barycentric MHR$\to$SMPL transfer, ours as in Tab.~\ref{tab:main}; full sets: EMDB-1 24{,}117, 3DPW 35{,}515 frames).}
\label{tab:rerun}
\setlength{\tabcolsep}{2.5pt}
\begin{tabular}{l cccc}
\toprule
& & \multicolumn{3}{c}{low light} \\
\cmidrule(lr){3-5}
Method & EMDB-1 hard & 3DPW & EMDB-1 & EMDB-1 hard \\
\midrule
SAM 3D Body (body) & 38.9 / 67.0 / 77.0 & 38.9 / 61.8 / 74.8 & 39.3 / 65.7 / 77.0 & 44.1 / 77.2 / 88.0 \\
NLF-L & 37.5 / 67.4 / 75.9 & 40.0 / 65.0 / 78.5 & 38.7 / 66.5 / 76.2 & 42.0 / 72.3 / 80.2 \\
Fast SAM 3D Body & 39.7 / 69.0 / 79.9 & 39.8 / 62.9 / 76.2 & 40.2 / 67.8 / 79.8 & 44.8 / 79.3 / 90.4 \\
\rowcolor{oursbg}
\paper-H, MHR mesh $\to$ SMPL & \textbf{37.2 / 62.7 / 71.9} & 37.0 / 58.8 / 71.1 & \textbf{36.8 / 62.2 / 72.3} & \textbf{41.0 / 69.9 / 79.5} \\
\rowcolor{oursbg}
\paper-H, dense points $\to$ SMPL & 40.3 / 65.8 / 75.8 & 37.1 / 59.6 / 70.8 & 39.6 / 64.4 / 74.5 & 44.2 / 73.0 / 83.4 \\
\rowcolor{oursbg}
\paper-L (ours) & 38.9 / 71.1 / 81.3 & \textbf{35.6 / 56.5 / 67.8} & 38.2 / 66.2 / 76.7 & 42.0 / 76.0 / 86.2 \\
\bottomrule
\end{tabular}
\end{table*}

\begin{table}[t]
\centering\tabfont
\caption{MPJPE (14 joints, mm) on clean / motion-blurred (1.5--3.5\% of the longer side) / low-resolution (240--360\,px, upscaled back) copies; standard protocol.}
\label{tab:degrade}
\setlength{\tabcolsep}{2.5pt}
\resizebox{\linewidth}{!}{\begin{tabular}{l ccc}
\toprule
Method & 3DPW & EMDB-1 & RICH test \\
\midrule
SAM 3D Body & 55.1 / 66.9 / 59.6 & 59.3 / 68.6 / 63.9 & 52.1 / 84.9 / 58.0 \\
\rowcolor{oursbg}
\paper-H, MHR mesh $\to$ SMPL & 53.2 / 64.1 / 56.3 & \textbf{57.4} / \textbf{64.2} / \textbf{60.7} & \textbf{47.3} / 75.6 / 53.3 \\
\rowcolor{oursbg}
\paper-H, dense points $\to$ SMPL & 54.8 / 65.1 / 57.3 & 60.1 / 66.5 / 62.9 & 50.3 / 77.3 / 55.4 \\
\rowcolor{oursbg}
\paper-L (ours) & \textbf{53.1 / 61.5 / 55.4} & 63.0 / 66.9 / 63.8 & 47.8 / \textbf{72.7 / 51.8} \\
\bottomrule
\end{tabular}}
\end{table}

\begin{table}[t]
\centering\tabfont
\caption{EMDB-1, PA-MPJPE / MPJPE over 24 SMPL joints (mm), standard protocol on all 17 sequences, every 2nd frame (our models as in Tab.~\ref{tab:main}); \paper-H's two MHR mesh $\to$ SMPL rows (official: MHR's official converter; our fit) are the two most accurate.}
\label{tab:emdb1}
\setlength{\tabcolsep}{3pt}
\begin{tabular}{l c}
\toprule
Method & EMDB-1 \\
\midrule
SAM 3D Body & 37.2 / 57.3 \\
Fast SAM 3D Body & 37.8 / 57.9 \\
NLF-L & 38.3 / 62.3 \\
\rowcolor{oursbg}
\paper-H, MHR mesh $\to$ SMPL, official & \textbf{36.8 / 55.3} \\
\rowcolor{oursbg}
\paper-H, MHR mesh $\to$ SMPL, our fit & 36.9 / 55.5 \\
\rowcolor{oursbg}
\paper-H, dense points $\to$ SMPL & 40.0 / 59.4 \\
\rowcolor{oursbg}
\paper-L (ours) & 39.2 / 61.0 \\
\rowcolor{oursbg}
\paper-S (ours) & 45.9 / 69.7 \\
\bottomrule
\end{tabular}
\end{table}

\begin{table}[t]
\centering\tabfont
\caption{2D keypoints on COCO val2017 (17 keypoints, ground-truth boxes, 6{,}352 people): keypoint AP and AP$_{75}$.}
\label{tab:modalities}
\label{tab:mod2d}
\setlength{\tabcolsep}{4pt}
\begin{tabular}{l cc}
\toprule
Method & AP & AP$_{75}$ \\
\midrule
SAM 3D Body (estimator, projected mesh keypoints) & 62.0 & \textbf{69.8} \\
\rowcolor{oursbg}
\paper-L, projected mesh keypoints & 54.2 & 60.4 \\
\rowcolor{oursbg}
\paper-L, dense-head 2D & \textbf{62.5} & \textbf{69.8} \\
\bottomrule
\end{tabular}
\end{table}

\begin{table}[t]
\centering\tabfont
\caption{2D under real low light (ExLPose~\cite{lee2023exlpose}): OKS AP over its 12 limb joints on its low-light photos and their well-lit pairs.}
\label{tab:modexl}
\setlength{\tabcolsep}{3pt}
\begin{tabular}{l cccc}
\toprule
Method & normal & hard & extreme & well-lit \\
\midrule
ViTPose+-H~\cite{xu2023vitposepp} & 39.9 & 13.2 & 2.0 & \textbf{89.2} \\
SAM 3D Body & 63.1 & 38.5 & 15.1 & 87.1 \\
\rowcolor{oursbg}
\paper-H, projected mesh keypoints & 62.4 & 39.0 & 19.2 & 84.2 \\
\rowcolor{oursbg}
\paper-H, dense-head 2D & 62.4 & 39.5 & 18.7 & 85.3 \\
\rowcolor{oursbg}
\paper-L, projected mesh keypoints & 64.1 & 44.9 & 27.2 & 83.7 \\
\rowcolor{oursbg}
\paper-L, dense-head 2D & \textbf{66.9} & \textbf{49.1} & \textbf{34.0} & 86.0 \\
\bottomrule
\end{tabular}
\end{table}

\begin{table}[t]
\centering\tabfont
\caption{G1 joints and bones (every 10th frame): link MPJPE and PA (mm, robot scale) and joint-angle error (A, $^\circ$) against the ground-truth skeleton retargeted by the same IK, and bone-direction error (B, $^\circ$) against the ground-truth person. SMPL $\to$ IK: the predicted SMPL through the two-stage IK (the IK stage alone runs at 4.6\,fps); G1 head: the in-model head in one pass, trained on direction-first labels.}
\label{tab:modg1}
\setlength{\tabcolsep}{2.2pt}
\begin{tabular}{l cccc cccc}
\toprule
& \multicolumn{4}{c}{3DPW} & \multicolumn{4}{c}{EMDB-1} \\
\cmidrule(lr){2-5}\cmidrule(lr){6-9}
Method & MPJPE & PA & A & B & MPJPE & PA & A & B \\
\midrule
\rowcolor{oursbg}
SMPL $\to$ IK & \textbf{40.6} & \textbf{32.2} & \textbf{6.6} & 13.7 & \textbf{43.6} & \textbf{35.4} & \textbf{7.7} & 12.8 \\
\rowcolor{oursbg}
G1 head & 71.0 & 60.0 & 13.8 & \textbf{8.2} & 72.1 & 62.8 & 13.4 & \textbf{9.6} \\
\bottomrule
\end{tabular}
\end{table}

\begin{table}[t]
\centering\tabfont
\caption{\paper-L's body parts against parts rendered from the ground-truth mesh, as in training (Sec.~\ref{sec:suppmethod}).}
\label{tab:modparts}
\setlength{\tabcolsep}{4pt}
\begin{tabular}{l cc}
\toprule
Set & part mIoU & silhouette IoU \\
\midrule
\rowcolor{oursbg}
3DPW & 0.446 & 0.677 \\
\rowcolor{oursbg}
EMDB-1 & 0.511 & 0.762 \\
\rowcolor{oursbg}
EMDB dev & 0.541 & 0.793 \\
\bottomrule
\end{tabular}
\end{table}

\begin{table}[t]
\centering\tabfont
\caption{2D feet on COCO-WholeBody val (2{,}682 people): foot OKS AP, and the heel$\to$big-toe direction error over 3{,}743 labelled feet (mean, median, share above 30$^\circ$).}
\label{tab:modfeet}
\setlength{\tabcolsep}{3pt}
\begin{tabular}{l cccc}
\toprule
Method & AP & mean & median & $>$30$^\circ$ \\
\midrule
SAM 3D Body & 83.7 & 16.5 & 7.8 & 14.1\% \\
\rowcolor{oursbg}
\paper-H, projected mesh & 83.5 & 16.8 & 8.0 & 14.8\% \\
\rowcolor{oursbg}
\paper-H, dense-head 2D & 85.8 & \textbf{15.5} & 7.0 & \textbf{12.9\%} \\
\rowcolor{oursbg}
\paper-L, projected mesh & 80.2 & 19.0 & 9.6 & 17.0\% \\
\rowcolor{oursbg}
\paper-L, dense-head 2D & \textbf{87.2} & \textbf{15.5} & \textbf{6.8} & 13.3\% \\
\bottomrule
\end{tabular}
\end{table}

\begin{table}[t]
\centering\tabfont
\caption{Legs and feet of \paper-L's outputs (3DPW, EMDB-1, EMDB dev, RICH, MOYO, EgoBody; per-set medians averaged). Leg: predicted / true 3D leg length; short: image leg more than 10\% short; feet: heel$\to$toe / sole-normal error ($^\circ$).}
\label{tab:points}
\setlength{\tabcolsep}{3pt}
\begin{tabular}{l ccc}
\toprule
Output & leg & short & feet \\
\midrule
\rowcolor{oursbg}
MHR mesh (auxiliary) & 0.985 & 3.3\% & 12.8 / 11.3 \\
\rowcolor{oursbg}
dense points & 0.995 & 2.6\% & 10.8 / 10.5 \\
\rowcolor{oursbg}
SMPL fitted to the points & \textbf{1.001} & \textbf{2.3\%} & \textbf{10.5 / 9.8} \\
\bottomrule
\end{tabular}
\end{table}

\begin{figure*}[t]
\centering
\includegraphics[width=\linewidth]{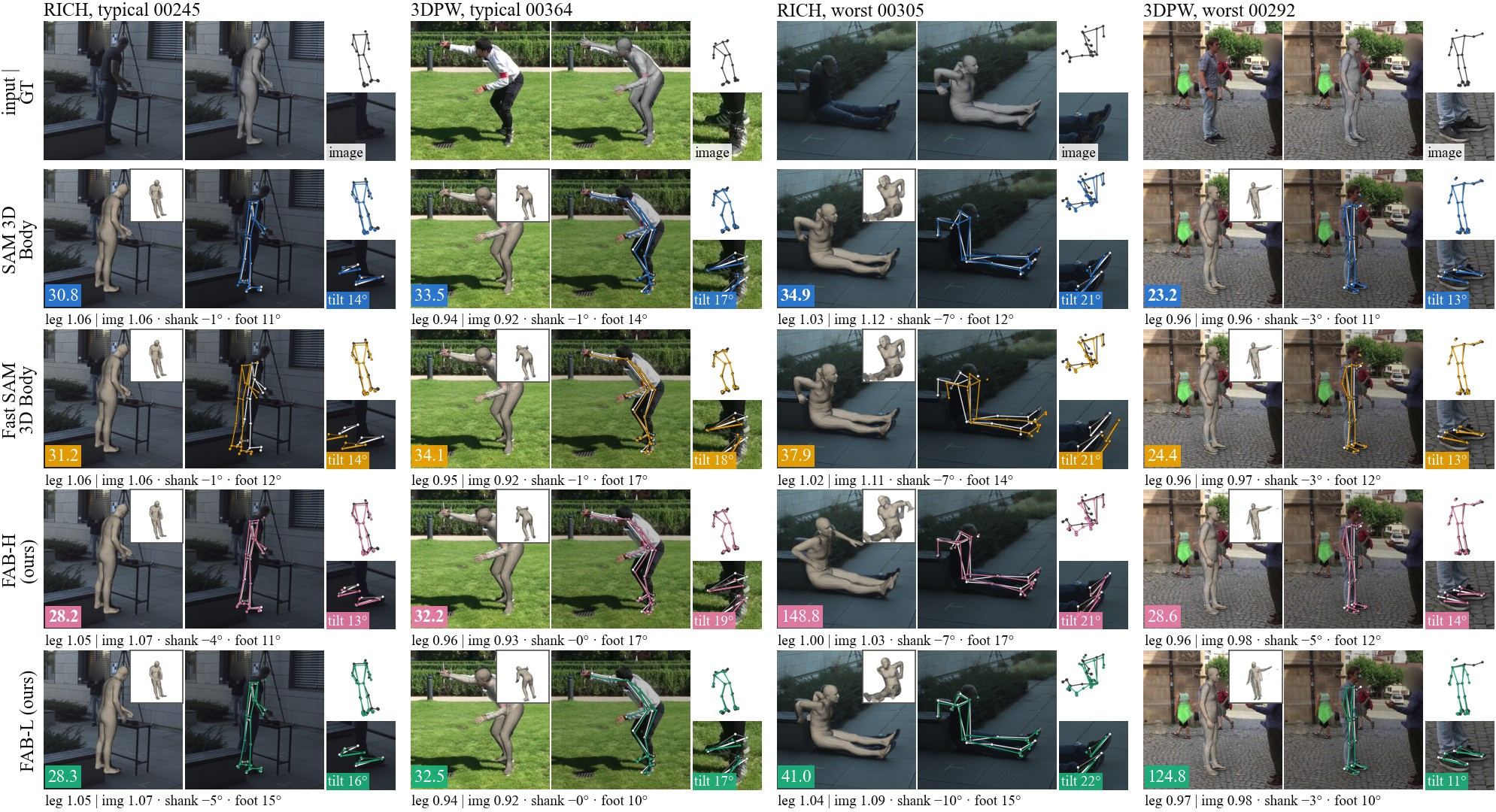}
\caption{Typical frames (errors closest to the set's mean) and the worst frames for \paper (both \paper models lose to both baselines by the largest margin) on RICH and 3DPW; panels as in Fig.~\ref{fig:qual}. Worst on RICH: \paper-H puts an arm forward (148.8\,mm); on 3DPW: \paper-L lowers the arm that the ground truth raises behind the body (124.8\,mm).}
\label{fig:qualtyp}
\end{figure*}

\paragraph{Findings.}
\begin{itemize}
\item \textbf{FAB-Hard.} With annotation boxes \paper-L trails only NLF-L on extreme poses; with our detector's boxes it has the lowest MPJPE in every category but extreme poses (Tab.~\ref{tab:fabhard}).
\item \textbf{Harmony4D (Tab.~\ref{tab:morebench}).} For the 768 people whose box overlaps another person's by IoU $>0.6$, \paper-L's MPJPE$_{24}$ exceeds 150\,mm for 7.4\% of them, under half SAM 3D Body's share (MHR mesh through the barycentric map), and its mean MPJPE$_{24}$ there (70.2\,mm) is far below SAM 3D Body's. These 6.9\% of the people give 29\% of its MPJPE$_{24}$ lead over all people in this comparison.
\item \textbf{Legs and feet.} The points' largest foot gain over the mesh is on MOYO, the clean reference (heel$\to$toe error 7.1$^\circ$ for the points, well below the mesh's, and nearly as low for the SMPL fitted to the points; averages over all sets: Tab.~\ref{tab:points}). In the image every model points the feet roughly right (Tab.~\ref{tab:modfeet}), so the remaining 3D foot error is mostly pitch.
\item \textbf{Low light and degradations.} Under real low light (ExLPose) \paper-L's dense 2D degrades least of all methods and is 2.8--6.8 points ahead of its own projected mesh (Tab.~\ref{tab:modexl}); synthetic low light, blur and low resolution are in Tabs.~\ref{tab:rerun} and~\ref{tab:degrade}.
\end{itemize}

\subsection{World space and contacts}
\label{sec:suppvideoworld}

\begin{table}[t]
\centering\tabfont
\caption{Cause analysis of Tab.~\ref{tab:worldmain}'s RICH-191 row, split after the fact by our detector's failure (WA / W: WA- / W-MPJPE in mm, RTE in \%): the 23 videos on which our detector's box has IoU below 0.5 with the ground-truth box on more than 10\% of the frames, and the other 168. The world methods use the ground-truth boxes in both (CRISP: its own masks). A diagnostic, not a benchmark.}
\label{tab:richsplit}
\setlength{\tabcolsep}{2.5pt}
\begin{tabular}{l ccc ccc}
\toprule
& \multicolumn{3}{c}{subject followed (168)} & \multicolumn{3}{c}{subject lost (23)} \\
\cmidrule(lr){2-4}\cmidrule(lr){5-7}
Method & WA & W & RTE & WA & W & RTE \\
\midrule
TRAM & 87.1 & 138.5 & 1.88 & 164.5 & 279.9 & 5.05 \\
GVHMR & 76.4 & 125.3 & 2.31 & 83.2 & 123.1 & 2.58 \\
GEM-X (GENMO) & 78.0 & 127.6 & 2.52 & 89.7 & 135.5 & 3.12 \\
WHAM & 117.8 & 204.2 & 5.11 & 115.7 & 183.2 & 4.62 \\
DuoMo & 53.1 & 80.3 & 1.62 & \textbf{75.2} & \textbf{107.5} & \textbf{2.31} \\
CRISP & 79.6 & 128.3 & 1.97 & 187.1 & 392.2 & 14.38 \\
\rowcolor{oursbg}
\paper-L (ours) & \textbf{47.3} & \textbf{71.7} & \textbf{1.08} & 212.0 & 607.6 & 42.34 \\
\bottomrule
\end{tabular}
\end{table}
\begin{table*}[t]
\centering\tabfont
\caption{World space on subsets of Tab.~\ref{tab:worldmain}'s 25 EMDB-2 sequences (WA, W in mm; RTE in \%): without the skateboard sequence (P8\_64), the 7 used for selection, and the 18 never used for any choice (pooled over 30{,}902 frames). Ground-truth camera motion: the same bodies, with our intrinsics.}
\label{tab:world}
\label{tab:heldout}
\setlength{\tabcolsep}{4pt}
\begin{tabular}{l ccc ccc ccc}
\toprule
& \multicolumn{3}{c}{without skateboard (24)} & \multicolumn{3}{c}{selection (7)} & \multicolumn{3}{c}{never selected (18)} \\
\cmidrule(lr){2-4}\cmidrule(lr){5-7}\cmidrule(lr){8-10}
Method & WA & W & RTE & WA & W & RTE & WA & W & RTE \\
\midrule
TRAM & 85.7 & 315.0 & 3.06 & 74.7 & 253.0 & 2.09 & 100.7 & 358.0 & 3.64 \\
GEM-X (GENMO) & 85.9 & 217.8 & 2.12 & 81.3 & 201.0 & 2.14 & 127.3 & 298.6 & 2.69 \\
GVHMR & 80.7 & 218.6 & 1.48 & 68.1 & 185.4 & 1.24 & 125.3 & 307.5 & 2.17 \\
WHAM & 121.6 & 336.1 & 4.33 & 115.7 & 328.1 & 3.78 & 161.9 & 406.8 & 4.97 \\
DuoMo & 64.2 & 158.7 & 1.14 & 56.6 & 149.3 & 1.24 & 71.8 & 175.3 & 1.17 \\
CRISP & 73.6 & 204.2 & 1.47 & 68.8 & 187.2 & 0.99 & 77.0 & 217.2 & 1.64 \\
\rowcolor{oursbg}
\paper-L (ours) & \textbf{56.7} & \textbf{144.5} & \textbf{1.08} & \textbf{49.6} & \textbf{123.5} & \textbf{0.91} & \textbf{60.1} & \textbf{156.6} & \textbf{1.13} \\
\rowcolor{oursbg}\multicolumn{10}{l}{}\\[-6pt]
\rowcolor{oursbg}
\paper-L (ours), ground-truth camera motion & 50.6 & 135.7 & 0.19 & 48.9 & 126.1 & 0.15 & 51.8 & 142.5 & 0.21 \\
\bottomrule
\end{tabular}
\end{table*}

\begin{table}[t]
\centering\tabfont
\caption{RICH-191 with the dataset's intrinsics in place of the pipeline's estimate: every body through our world pipeline with the same boxes, cameras, depth and contacts. Columns and our models as in Tab.~\ref{tab:worldmain}; SAM 3D Bodies through MHR's official converter.}
\label{tab:richk}
\setlength{\tabcolsep}{4pt}
\begin{tabular}{l ccccc}
\toprule
Method & WA & W & RTE & J & FS \\
\midrule
SAM 3D Body & 74.6 & 146.6 & \textbf{5.81} & 3.56 & \textbf{6.99} \\
Fast SAM 3D Body & 75.1 & 148.2 & 5.85 & \textbf{3.41} & 7.06 \\
NLF-L & 70.4 & 139.4 & 5.90 & 4.01 & 8.22 \\
\rowcolor{oursbg}
\paper-H, MHR mesh $\to$ SMPL & 69.3 & 137.8 & 5.91 & 3.45 & 7.02 \\
\rowcolor{oursbg}
\paper-H, dense points $\to$ SMPL & 76.0 & 152.0 & 6.16 & 4.93 & 9.63 \\
\rowcolor{oursbg}
\paper-L (ours) & \textbf{66.3} & \textbf{134.4} & 5.90 & 3.86 & 7.40 \\
\rowcolor{oursbg}
\paper-S (ours) & 80.4 & 157.4 & 6.08 & 4.78 & 9.57 \\
\bottomrule
\end{tabular}
\end{table}

\paragraph{Protocols.}
The world methods of Tab.~\ref{tab:worldmain} are re-run with their released code and settings on all 25 EMDB-2 sequences and on RICH-191 (GVHMR's and WHAM's protocol: 106{,}446 frames, 24 joints); GVHMR's own evaluation reproduces its published numbers.
\begin{itemize}
\item \textbf{Metrics} as in WHAM (foot sliding on its vertex definition), frame-weighted over the 25 tracks (foot sliding: over contact frames). On EMDB-2 the ground-truth joints are the neutral SMPL joint regressor applied to the gendered ground-truth mesh for every row, as in GVHMR's EMDB evaluation and in GVHMR's and WHAM's RICH evaluations; WHAM's own EMDB evaluation takes the gendered model's joints, which would raise every EMDB-2 WA-MPJPE in Tab.~\ref{tab:worldmain} by 0.4--1.0\,mm.
\item \textbf{Inputs} as in Tab.~\ref{tab:worldmain}'s Box and Cam.\ columns, plus: TRAM's SLAM masks are SAM masks of the ground-truth boxes; WHAM also takes the ground-truth first frame and flip augmentation on EMDB-2; on RICH-191, TRAM, GVHMR, GEM-X and WHAM take the ground-truth boxes, and CRISP runs as on EMDB-2. TRAM's SLAM is not run-to-run deterministic; we report its per-sequence medians of three runs on EMDB-2.
\item \textbf{World bodies} on RICH-191: TRAM's camera-frame body placed by the static camera, then its trajectory filter; GVHMR's global output with its test-time post-processing; GEM-X's after its contact / IK post-processing; WHAM's demo path, whose world root starts from its own first-frame estimate (its released RICH evaluation takes it from the ground truth).
\item \textbf{DuoMo}~\cite{wang2026duomo} (release-v1 at cfc1cdc): its released world pipeline with its released code and its bundled cuDNN, the ground-truth boxes in place of its detector, the ground-truth intrinsics in place of its estimate and its own gravity estimate; camera motion: static on RICH-191, as in its RICH protocol, and on EMDB-2 the TRAM SLAM poses its release provides.
\item \textbf{CRISP}~\cite{wang2026crisp} (released code at 6cdc755) runs its full video pipeline on every frame: its own person mask (Grounding DINO~\cite{liu2024groundingdino} and SAM~2~\cite{ravi2025sam2}, no ground-truth boxes), camera and focal (MegaSaM~\cite{li2025megasam}), its GVHMR fork on that camera, planar scene primitives and its post-processing (GeoCalib up axis, grounding); we score its world SMPL output (24 joints).
\item \textbf{Ours.} In the middle block both SAM 3D Bodies go through MHR's official converter, within 0.9\,mm WA-MPJPE of each other per sequence. The step rows run on the same tracks and bodies: SMPL fitted to the points, placed by ViPE's camera at its own scale and smoothed offline, before the joint human--scene optimization, and the same with the ground-truth camera motion. On RICH-191 \paper-L runs with its own intrinsics (the static-camera focal rule; the dataset's: Tab.~\ref{tab:richk}) and ViPE camera, as on EMDB-2, and its own detector (Sec.~\ref{sec:suppmethod}); the 23 videos on which it loses the subject (Tab.~\ref{tab:richsplit}) hold 12{,}016 frames, mostly distant people doing push-ups and dips on Gym's cam\_05.
\item Videos over 600 frames run in overlapping chunks (Sec.~\ref{sec:suppscenes}; on RICH-191, 65 of the 191). 3DPW gets no world-camera metric: one homography explains its matched static points to 1.1--2.5\,px on 10 of 11 clips, so its cameras only rotate.
\end{itemize}

\begin{figure*}[t]
\centering
\includegraphics[width=\linewidth]{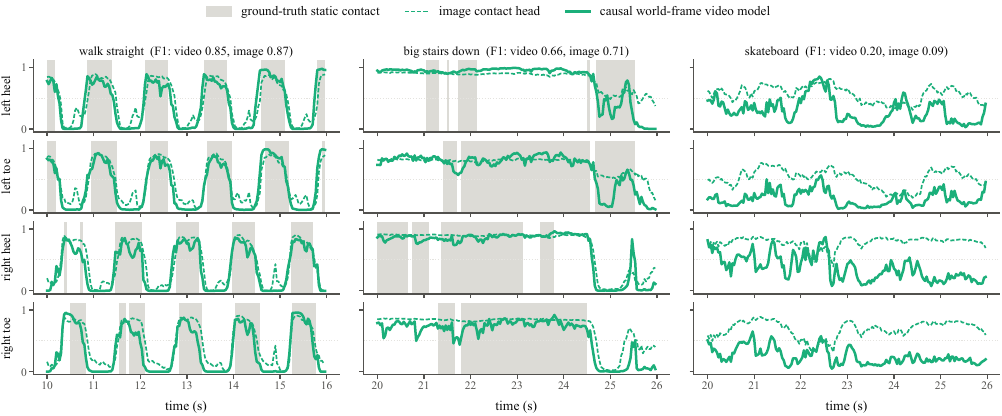}
\caption{Static foot contact over 6\,s of three EMDB-2 sequences: a walk, stairs and a moving support (the skateboard). Grey: ground-truth static contact (Sec.~\ref{sec:contactlabels}); dashed: the image head; solid: the causal world-frame video model; dotted: the 0.5 threshold. F1 in the titles pools the four foot channels at 0.5 over the whole sequence.}
\label{fig:contacts}
\end{figure*}

\begin{figure*}[t]
\centering
\includegraphics[width=\linewidth]{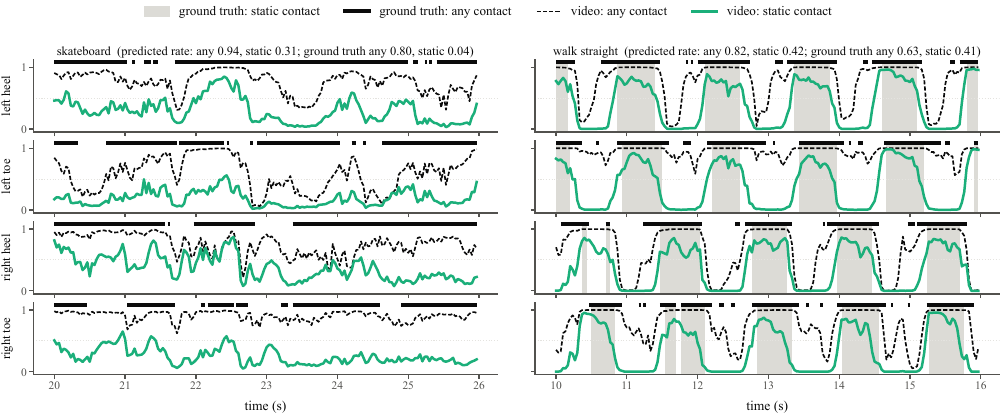}
\caption{Static against any contact on the skateboard and on a walk. Grey / black bar: ground-truth static / any contact (Sec.~\ref{sec:contactlabels}); solid / dashed: the world-frame video model's static / any-contact output; dotted: the 0.5 threshold. On the skateboard the feet touch the board (any contact) on 80\% of the frames (94\% predicted) but are rarely world-fixed (Sec.~\ref{sec:exp_contacts}); while walking the two coincide.}
\label{fig:contactany}
\end{figure*}

\paragraph{Contacts.}\label{sec:contactlabels}
Evaluation contact labels come from the ground-truth world motion: a heel or toe channel is in static contact when its lowest vertex is less than 4\,cm above the local floor and moves slower than 0.3\,m\,s$^{-1}$, and in any contact on the height test alone; the local floor is the 10th percentile of the heel and toe heights within about 15\,cm of the point and 5\,s of the frame, or within 15\,cm over the whole sequence where fewer than 8 heights fall in that window (1.8\,\% of EMDB's and 0.4\,\% of RICH's foot channel-frames), so a ground-truth trajectory that drifts vertically over a long sequence is not given another pass's floor. Contacts against RICH's human--scene annotations reach a hand F1 of 0.88 for the left hand and 0.82 for the right; the causal video models time contact onsets within 79\,ms on RICH, and on EMDB-2 the video model runs on the world pipeline's cameras.

\subsection{Humanoid}
\label{sec:supphumanoid}
\begin{table*}[t]
\centering\tabfont
\caption{Keypoint error (mm) of the tracked G1 against the real person before the first fall, for the references of Tab.~\ref{tab:trackersmain}. ours: the FAB-L G1 head (per frame) references; IK: the sequence IK on our SMPL fit.}
\label{tab:trackers}
\setlength{\tabcolsep}{3pt}
\begin{tabular}{l cc>{\columncolor{oursbg}}c cc>{\columncolor{oursbg}}c}
\toprule
& \multicolumn{3}{c}{all 25 clips} & \multicolumn{3}{c}{trackable 17} \\
\cmidrule(lr){2-4}\cmidrule(lr){5-7}
Tracker & IK & Fast SAM 3D Body & ours & IK & Fast SAM 3D Body & ours \\
\midrule
SONIC-SMPL & \textbf{120} & 146 & 127 & \textbf{99} & 111 & 105 \\
SONIC-G1 & 143 & 154 & \textbf{131} & 105 & \textbf{103} & 110 \\
\bottomrule
\end{tabular}
\end{table*}
\begin{table*}[t]
\centering\tabfont
\caption{References of Tab.~\ref{tab:trackersmain} beyond its cells: the reference itself (keypoint error, mm; link acceleration: the mean pelvis-relative link acceleration, m\,s$^{-2}$; joint jitter: the mean absolute second difference of the joint angles, rad\,s$^{-2}$; sole slip: the mean horizontal sole speed while the sole is within 2\,cm of the floor, m\,s$^{-1}$; the same statistics in Tabs.~\ref{tab:trackersmain}\ifdense, \ref{tab:retmodes}\fi{} and~\ref{tab:retargetinputs}) and tracking on the trackable 17 (falls and bone-direction error, $^\circ$; the per-frame references fall on RICH ParkingLot1\_004's push-up and, with SONIC-SMPL, on MOYO's warrior III pose, \ifdense the mesh mode's\else the motion-level retargeter's\fi{} also on warrior III with SONIC-G1).}
\label{tab:retargeters}
\setlength{\tabcolsep}{3pt}
\begin{tabular}{l ccc cccc cccc}
\toprule
& \multicolumn{3}{c}{all 25: reference} & \multicolumn{4}{c}{trackable 17: tracked} & \multicolumn{4}{c}{trackable 17: reference} \\
\cmidrule(lr){2-4}\cmidrule(lr){5-8}\cmidrule(lr){9-12}
& & & & \multicolumn{2}{c}{SONIC-SMPL} & \multicolumn{2}{c}{SONIC-G1} & & & & \\
\cmidrule(lr){5-6}\cmidrule(lr){7-8}
& kp & accel & jitter & falls & bone & falls & bone & bone / kp & slip & accel & jitter \\
\midrule
SOMA & \textbf{92} & 2.24 & 5.2 & \textbf{1} & 16.4 & \textbf{1} & 18.3 & 14.2 / \textbf{83} & 0.36 & 2.07 & 5.1 \\
\ifdense
GMR & 98 & 4.05 & 6.7 & \textbf{1} & 16.3 & 2 & 15.7 & 13.0 / 85 & 0.33 & 2.94 & 6.4 \\
\else
GMR & 98 & 4.05 & 6.7 & \textbf{1} & 16.3 & 2 & \textbf{15.7} & 13.0 / 85 & 0.33 & 2.94 & 6.4 \\
\fi
\ifdense
UMR & 124 & 5.45 & 5.8 & \textbf{1} & \textbf{15.9} & 2 & 16.1 & 12.8 / 84 & 0.50 & 3.25 & 5.8 \\
\else
UMR & 124 & 5.45 & 5.8 & \textbf{1} & \textbf{15.9} & 2 & 16.1 & 12.8 / 84 & 0.50 & 3.25 & 5.8 \\
\fi
OmniRetarget & 131 & 8.56 & 9.0 & \textbf{1} & 16.2 & 2 & 17.4 & 14.4 / 84 & 0.42 & 5.51 & 8.8 \\
sequence IK & 140 & 3.72 & 4.6 & \textbf{1} & 16.8 & 2 & 20.0 & 18.0 / 97 & 0.28 & 2.37 & 4.9 \\
Fast SAM 3D Body & 140 & 6.20 & 6.1 & 2 & 19.7 & 3 & 19.9 & 20.0 / 106 & 0.39 & 4.66 & 6.3 \\
\ifdense
\rowcolor{oursbg}
FAB-L G1 head (per frame) & 108 & 10.99 & 21.0 & 2 & 18.2 & \textbf{1} & 17.2 & 11.4 / 90 & 0.41 & 10.34 & 21.8 \\
\rowcolor{oursbg}
FAB-Retarget [skeleton + mesh] & 93 & 1.38 & 2.4 & \textbf{1} & 16.9 & 2 & 15.8 & 11.2 / 85 & \textbf{0.14} & 1.08 & 2.4 \\
\rowcolor{oursbg}
\quad started from SOMA's motion & 93 & 1.39 & 2.5 & \textbf{1} & 17.0 & \textbf{1} & 16.9 & \textbf{11.0} / 85 & 0.15 & 1.18 & 2.5 \\
\rowcolor{oursbg}
FAB-Retarget [skeleton + mesh + points] & \textbf{92} & \textbf{1.34} & \textbf{2.2} & \textbf{1} & 16.0 & \textbf{1} & \textbf{15.3} & 11.4 / 84 & 0.17 & \textbf{1.07} & \textbf{2.1} \\
\rowcolor{oursbg}
FAB-Retarget [skeleton + mesh], other pinning run & 93 & 1.38 & 2.4 & \textbf{1} & 16.9 & 2 & 15.8 & 11.2 / 85 & \textbf{0.14} & 1.08 & 2.4 \\
\else
\rowcolor{oursbg}
FAB-L G1 head (per frame) & 108 & 10.99 & 21.0 & 2 & 18.2 & \textbf{1} & 17.2 & 11.4 / 90 & 0.41 & 10.34 & 21.8 \\
\rowcolor{oursbg}
ours, motion-level retargeter & 93 & \textbf{1.38} & \textbf{2.4} & \textbf{1} & 16.9 & 2 & 15.8 & 11.2 / 85 & \textbf{0.14} & \textbf{1.08} & \textbf{2.4} \\
\rowcolor{oursbg}
\quad started from SOMA's motion & 93 & 1.39 & 2.5 & \textbf{1} & 17.0 & \textbf{1} & 16.9 & \textbf{11.0} / 85 & 0.15 & 1.18 & 2.5 \\
\fi
\bottomrule
\end{tabular}
\end{table*}
\ifdense
\begin{table}[t]
\centering\tabfont
\caption{FAB-Retarget's three modes on the 25 test clips of Tab.~\ref{tab:trackersmain}, pinned in one batch and tracked by SONIC (G1 mode); points is the default (pitch: bold closest to 0).}
\label{tab:retmodes}
\setlength{\tabcolsep}{1.6pt}
\resizebox{\linewidth}{!}{\begin{tabular}{l cccccc c}
\toprule
& \multicolumn{6}{c}{reference} & tracked \\
\cmidrule(lr){2-7}
FAB-Retarget & bone & part rot. & slip & accel & jitter & pitch & falls, bone \\
\midrule
\rowcolor{oursbg}
{[skeleton]} & 11.7 & 33.6 & 0.12 & 1.39 & 2.4 & \textbf{\boldmath$-$12.5} & 5, 19.2 \\
\rowcolor{oursbg}
{[skeleton + mesh]} & \textbf{11.6} & 33.6 & \textbf{0.11} & 1.38 & 2.4 & \textbf{\boldmath$-$12.5} & 6, 20.0 \\
\rowcolor{oursbg}
{[skeleton + mesh + points]} & 11.8 & \textbf{24.7} & 0.13 & \textbf{1.34} & \textbf{2.2} & $-$13.0 & \textbf{4}, \textbf{18.3} \\
\bottomrule
\end{tabular}}
\par\vspace{2pt}{\footnotesize\raggedright Columns as in Tabs.~\ref{tab:trackersmain} and~\ref{tab:retargetinputs}; pitch: the G1 stance sole's pitch minus the person's ($^\circ$, signed, over the 24 clips with stance frames; Sec.~\ref{sec:supphumanoid}).\par}
\end{table}
\fi
\paragraph{Tracking in simulation.}
All references are tracked by SONIC in one MuJoCo setup with matched G1 model, flat ground, initial state, fall rule and scorer. The 25 clips span 3DPW, EMDB, RICH and MOYO; the trackable subset excludes acrobatics and torso inversions. Every retargeter receives the same world-space SMPL motion, except Fast SAM 3D Body, which retains its released per-frame SMPL and smoother. We compare SOMA, UMR, OmniRetarget and GMR under their released defaults; their full mappings and simulator settings are released with the evaluation code.

\ifablready
\ifdense\def\skrow{mesh}\else\def\skrow{skeleton}\fi
\begin{table*}[t]
\centering\tabfont
\caption{Inputs of the motion-level retargeter on the 25 test clips of Tab.~\ref{tab:trackersmain}, tracked in the same MuJoCo setup. Reference: bone-direction error and part-rotation error (Eq.~\ref{eq:partrot}) against the real person ($^\circ$), sole slip (m\,s$^{-1}$), link acceleration (m\,s$^{-2}$) and joint jitter (rad\,s$^{-2}$). Tracked: falls of 25, bone-direction error before the first fall ($^\circ$). \ifdense Mesh: FAB-Retarget's mesh mode, with\else Skeleton:\fi{} bone directions, pelvis frame and the body's mesh-rule foot contacts (Sec.~\ref{sec:suppmethod}); contacts: the video model's static heel / toe contacts replace them, pin the feet and flatten the sole when both touch; points: the dense-point term. Last row: \skrow{} + points with UMR's learned correspondence~\cite{cao2026umr} in place of ours. \ifdense The mesh and mesh + points rows are FAB-Retarget's mesh and points modes (ours, mesh / points).\fi}
\label{tab:retargetinputs}
\setlength{\tabcolsep}{3pt}
\begin{tabular}{l ccccc cc cc}
\toprule
& \multicolumn{5}{c}{reference} & \multicolumn{2}{c}{SONIC-SMPL} & \multicolumn{2}{c}{SONIC-G1} \\
\cmidrule(lr){2-6}\cmidrule(lr){7-8}\cmidrule(lr){9-10}
Inputs & bone & part rot. & slip & link acc. & jitter & falls & bone & falls & bone \\
\midrule
\rowcolor{oursbg}
\skrow{} & \textbf{11.6} & 33.6 & \textbf{0.11} & 1.38 & 2.4 & 6 & 19.3 & 6 & 20.0 \\
\rowcolor{oursbg}
\skrow{} + contacts & 13.1 & 34.8 & 0.14 & 2.26 & 3.1 & 5 & 20.1 & 6 & 20.4 \\
\rowcolor{oursbg}
\skrow{} + points & 11.8 & 24.7 & 0.13 & 1.34 & \textbf{2.2} & \textbf{4} & 17.9 & \textbf{4} & \textbf{18.3} \\
\rowcolor{oursbg}
\skrow{} + contacts + points & 13.0 & 25.7 & 0.16 & 2.68 & 3.0 & \textbf{4} & \textbf{17.8} & 5 & \textbf{18.3} \\
\rowcolor{oursbg}
\ifdense\quad ours with UMR's correspondence\else skeleton + points, UMR's correspondence\fi & 11.8 & \textbf{24.3} & 0.14 & \textbf{1.33} & \textbf{2.2} & \textbf{4} & 17.9 & \textbf{4} & \textbf{18.3} \\
\bottomrule
\end{tabular}
\end{table*}
\fi
\begin{table}[t]
\centering\tabfont
\caption{Tab.~\ref{tab:trackersmain}'s references on its 25 clips, tracked by SONIC as in Tab.~\ref{tab:trackersmain} and by NVIDIA's later SONIC release (v1.1)~\cite{luo2025sonic} in the same MuJoCo setup, with v1.1's own observation definition and default gains: falls of 25, bone-direction error before the first fall ($^\circ$).}
\label{tab:sonicv11}
\setlength{\tabcolsep}{1.7pt}
\begin{tabular}{l cc cc}
\toprule
& \multicolumn{2}{c}{SONIC} & \multicolumn{2}{c}{SONIC v1.1} \\
\cmidrule(lr){2-3}\cmidrule(lr){4-5}
Reference & SMPL & G1 & SMPL & G1 \\
\midrule
\ifdense
SOMA~\cite{nvidia2026soma} & \textbf{4}, 19.1 & 5, 23.4 & \textbf{4}, 22.5 & 6, 22.9 \\
GMR~\cite{araujo2025gmr} & \textbf{4}, 19.8 & 5, 19.6 & \textbf{4}, 22.4 & \textbf{4}, 22.4 \\
UMR~\cite{cao2026umr} & \textbf{4}, 19.7 & 5, 24.9 & \textbf{4}, \textbf{21.1} & 6, 25.3 \\
OmniRetarget~\cite{yang2025omniretarget} & 5, 19.7 & 8, 27.4 & 5, 21.7 & 8, 30.3 \\
sequence IK & 5, 19.4 & 6, 25.8 & 5, 22.6 & 6, 30.5 \\
Fast SAM 3D Body & 6, 24.7 & 6, 27.6 & 7, 25.7 & 8, 27.5 \\
\rowcolor{oursbg}
FAB-L G1 head (per frame) & 6, 20.7 & 5, 20.4 & \textbf{4}, 21.7 & \textbf{4}, 20.5 \\
\rowcolor{oursbg}
FAB-Retarget [skeleton + mesh] & 6, 19.3 & 6, 20.0 & \textbf{4}, 22.4 & 5, 21.6 \\
\rowcolor{oursbg}
FAB-Retarget [skeleton + mesh + points] & \textbf{4}, \textbf{17.9} & \textbf{4}, \textbf{18.3} & \textbf{4}, 22.6 & 5, \textbf{19.9} \\
\else
SOMA~\cite{nvidia2026soma} & \textbf{4}, \textbf{19.1} & \textbf{5}, 23.4 & \textbf{4}, 22.5 & 6, 22.9 \\
GMR~\cite{araujo2025gmr} & \textbf{4}, 19.8 & \textbf{5}, \textbf{19.6} & \textbf{4}, 22.4 & \textbf{4}, 22.4 \\
UMR~\cite{cao2026umr} & \textbf{4}, 19.7 & \textbf{5}, 24.9 & \textbf{4}, \textbf{21.1} & 6, 25.3 \\
OmniRetarget~\cite{yang2025omniretarget} & 5, 19.7 & 8, 27.4 & 5, 21.7 & 8, 30.3 \\
sequence IK & 5, 19.4 & 6, 25.8 & 5, 22.6 & 6, 30.5 \\
Fast SAM 3D Body & 6, 24.7 & 6, 27.6 & 7, 25.7 & 8, 27.5 \\
\rowcolor{oursbg}
FAB-L G1 head (per frame) & 6, 20.7 & \textbf{5}, 20.4 & \textbf{4}, 21.7 & \textbf{4}, \textbf{20.5} \\
\rowcolor{oursbg}
ours, motion-level retargeter & 6, 19.3 & 6, 20.0 & \textbf{4}, 22.4 & 5, 21.6 \\
\rowcolor{oursbg}
\quad started from SOMA's motion & 5, \textbf{19.1} & 6, 21.6 & 5, 21.9 & 5, 20.9 \\
\fi
\bottomrule
\end{tabular}
\end{table}
\paragraph{G1 terrain dataset.}
Its training clips come from Embody 3D~\cite{mclean2025embody3d}, AIST++~\cite{li2021aistpp}, EgoBody, MOYO and RICH (1{,}586, 749, 508, 124 and 30 clips) and pass bounds on foot slide, penetration and root steps. No clip from our evaluation sets trains. Of the 5{,}330 clips, 2{,}997 are cleared for training; of the other 2{,}333, the 15 scene-reconstruction clips and 1 lab clip are kept for evaluation only and 2{,}317 fail the tracking or bound checks. Every clip is retargeted with the foot-contact rule of Sec.~\ref{sec:suppmethod}; where \paper ran on video, its heel / toe contacts ship as an annotation the solver does not read.

\ifablready
\paragraph{Retargeter inputs.}
Tab.~\ref{tab:retargetinputs} adds the predicted contacts and the dense points to the \ifdense mesh mode\else skeleton\fi.\ifdense\ifablready{} With the points, the predicted contacts raise the bone-direction error (Tab.~\ref{tab:retargetinputs}), so neither mode uses them.\fi\fi{} Part-rotation error is the mean angle, over the $T$ frames and the 13 non-pelvis groups $\mathcal G$, between each group's G1 link rotation $R_{g,t}$ and the ground-truth body's $\hat R_{g,t}$, both relative to the pelvis $p$:
\begin{equation}
e_{\mathrm{rot}}=\frac{1}{T\,|\mathcal G|}\sum_{t=1}^{T}\sum_{g\in\mathcal G}\angle\!\left(R_{p,t}^{\top}R_{g,t},\,\hat R_{p,t}^{\top}\hat R_{g,t}\right),
\label{eq:partrot}
\end{equation}
where $\angle$ is the geodesic angle, $\arccos\frac{1}{2}(\operatorname{tr}A^{\top}B-1)$, and $\hat R_{g,t}=K(\bar X_g,\hat X_{g,t})\,R^0_g$ is the Kabsch rotation from the group's T-pose points $\bar X_g$ to the ground-truth SMPL vertices $\hat X_{g,t}$ at the same dense-point ids, composed with the link's T-pose orientation $R^0_g$.

\fi

\section{Scene reconstruction details}
\label{sec:suppscenes}

Tab.~\ref{tab:scene} compares our scenes, our bodies in them and our G1 references with three released pipelines, all run by us on the same clips and scorers; below are their settings, ours and each measure's protocol, then findings beyond the table.

\paragraph{Baselines.}
\begin{itemize}
\item \textbf{VideoMimic}~\cite{allshire2025videomimic}: its real2sim pipeline (stages 0--3, estimated shape: SAM~2 tracks, ViTPose, VIMO~\cite{wang2024tram} and BSTRO~\cite{huang2022rich} bodies, MegaSaM~\cite{li2025megasam} cameras and depth, the MegaHunter~\cite{allshire2025videomimic} human--scene scale, GeoCalib, NKSR~\cite{huang2023nksr} meshes) on its 300-frame block grid, joined by its own scripts.
\item \textbf{CRISP} (as in Sec.~\ref{sec:suppvideoworld}): its direct reconstruction, before the 0.2\,m lift its simulation step adds; clips over 600 frames in 600-frame chunks overlapping by 60, world motion joined by a rigid fit on the overlap with a cross-fade, scene geometry through the cameras; scene geometry and feet from its planar primitives. The G1 columns instead take its own robot-bound body (its floor correction and penetration optimization after the scene step, without the 0.2\,m spawn lift); the scene-geometry rows keep its direct output.
\item \textbf{JOSH}~\cite{liu2025josh} (released code at 5ef27ad): Pi3X~\cite{wang2025pi3} initialization, 100-frame chunks of 21 keyframes concatenated without a global adjustment; six release bugs fixed (cache, empty tracks, border crops, degenerate detection boxes, a missing entry script, an output directory written before creation); RICH frames cropped to MASt3R's~\cite{leroy2024mast3r} patch grid (Sec.~\ref{sec:suppsettings}); SAM~3 on 500-frame segments of clips over 1{,}000 frames; no mesh, so its whole-scene and feet scores use its point cloud; it outputs no gravity, so its G1 columns use ours, on terrain from the per-cell median height of its points (meshes: the per-cell top surface). Its own focal is 1.23--1.42$\times$ the true one on RICH-3; on Gym's static camera its camera moves by up to 346.6\,cm, where CRISP's barely moves, and along EMDB P3\_30 its 100-frame windows vary in scale from 0.09 to 2.06, ours only from 0.98 to 1.10, with a camera ATE of 234.4\,cm, many times ours.
\item \textbf{K fixed} (Tab.~\ref{tab:focalmatch}): CRISP's fixed-focal MegaSaM path and JOSH's fixed initial focal, switched on by us. Neither repository has a licence file; we run them for evaluation only.
\end{itemize}

\paragraph{Pipeline.}
\begin{itemize}
\item \textbf{MVS}: COLMAP PatchMatch~\cite{schoenberger2016mvs} with geometric consistency on ViPE's fixed cameras, keeping a pixel's depth where nearby views see the same surface with enough baseline; people are masked. Elsewhere a frame uses ViPE's depth, scaled by its median MVS / ViPE ratio where MVS covers enough of it, so a static camera keeps ViPE's depth (Sec.~\ref{sec:suppsettings}).
\item \textbf{Effect}: on EMDB-2, MVS lowers W-MPJPE by 3.6\,mm and WA-MPJPE by 1.0\,mm with foot sliding about level, from 150.8 and 58.1\,mm with ViPE's depth alone (the same chain otherwise); it raises RTE by more than 0.05\,\% on 7 sequences, most on P9\_80, P8\_65 and P6\_48, and lowers it by more than 0.05\,\% on 11.
\item \textbf{Fusion}: a fused point must be seen in several frames, lie away from depth edges and outside dynamic, person, sky and body masks; free-space carving drops a point the camera sees past; human-visibility carving as in Sec.~\ref{sec:scenes}.
\item \textbf{Texture}: a frame counts for a texel when the mesh's ray-cast depth agrees with the frame's depth and the view is not grazing; people are skipped, and colours blend by view angle and depth into one atlas (2.5--4.6\,mm per texel). On the 118 registered VideoMimic clips the median held-out reprojection PSNR is 23.5\,dB (17.7\,dB at the 10th percentile, 32.1\,dB at the 90th).
\end{itemize}

\paragraph{Joint human--scene optimization.}
Terms as in Sec.~\ref{sec:scenes}, with contact on predicted-contact frames and only in-view vertices, depth consistency as a robust log ratio, and a prior on $\log s$ whose width grows with the number of in-view foot-contact frames. A coarse search over $\log s$ precedes the descent, because the robust Geman--McClure terms saturate far from the solution (Sec.~\ref{sec:suppsettings}).

\paragraph{Tab.~\ref{tab:scene}, measure by measure.}
Clips: EgoBody, 3 egocentric recordings; SLOPER4D, seq.~7; PROX~\cite{hassan2019prox}, 12 recordings; EMDB P3\_30, stairs over 26.1\,m. RICH-3 and EgoBody Chamfer: per frame (per clip: Tab.~\ref{tab:scenegeo}). Our output per column, all from the one pipeline on \paper-L's bodies: the TSDF mesh for the scene-geometry columns, ViPE's cameras for the camera ATE, the world output for the human columns, \paper-L's camera-frame body placed by the ground-truth camera for body to scan (pooled over contact vertices), our body through our own scene before and after the joint human--scene optimization for feet to scan, the pinned motion-level retargeter for the G1 columns. VideoMimic's RICH-3 and EgoBody geometry is its per-frame depth.

\paragraph{Protocols.}
\begin{itemize}
\item \textbf{Per-frame Chamfer} (VideoMimic's protocol, RICH-3 and EgoBody): predicted depth and scan ray-cast from the ground-truth camera at the same pixels, rigid or Sim(3) Umeyama on the correspondences, Chamfer of 2\,cm voxel clouds (symmetric mean, cm), every 10th frame. RICH: ParkingLot2\_009 cam\_00, LectureHall\_019 cam\_00, Gym\_010 cam\_01 (45, 98 and 78 frames); EgoBody: the HoloLens PV camera of three recordings (96--120 frames each).
\item \textbf{SLOPER4D} (VideoMimic's scripts): predicted world points at the ground-truth LiDAR pixels of every frame (our TSDF mesh ray-cast through each frame's camera; CRISP's planar primitives; JOSH's depth), one rigid Umeyama fit, 2\,cm voxels, Chamfer as the mean of accuracy and completeness; human W-MPJPE / WA-MPJPE on 100-frame chunks; seq007\_garden\_001 (2{,}400 frames, 77.6\,m), the authors' official release.
\item \textbf{Body to scan.} Each camera-frame body is placed in the scan frame by the ground-truth camera, with nothing else aligned; per ground-truth contact vertex (RICH's official test annotations), the signed distance to the scan; penetration and float are the means of the negative and positive parts, $|s|$ the mean absolute distance.
\item \textbf{Feet to scan via own scene}: each body is placed through its own scene, which is aligned to the scan by ICP; mean sole-vertex distance on ground-truth contact frames.
\item \textbf{G1 references against their own terrain}: each method's own scene heightfield (4\,cm cells); penetration: the deepest of every 7th body vertex below the terrain, reported as the share of frames above 2\,cm; contact gap: the lowest foot vertex's distance on contact frames.
\item \textbf{VideoMimic's clips}: all 123; for comparisons in its frame or on its mesh, the 118 that register to our reconstruction: root RMSE at most 0.25\,m after a yaw-Sim(3) of VideoMimic's G1 root onto ours, with the median per-clip scale (0.933) as the fallback for 3 clips whose fitted scale lies outside 0.7--1.4.
\item \textbf{PROX} (CRISP's table; our settings where it leaves a choice open): 12 recordings, one per scene, on the same 5{,}499 frames for every method, each with its own intrinsics. The scans cover 37--91\% of each view, and reconstructed points outside the covered pixels (a ray cast of the scan from PROX's colour camera) are not scored. Chamfer is the mean nearest-neighbour distance, one-way from the reconstruction to the scan, both ways as the mean of the two directions, inside PROX's camera frustum; non-penetration follows PLACE~\cite{zhang2020place} on PROX's scan SDF (per frame, 1 without penetration, else the share of SMPL vertices outside), averaged over frames. Rigid: each method's own world placed in the scan by its camera and PROX's camera pose, at its own scale. On MPH1Library\_00034\_01 the seated body starts deep inside the 2.5D scene, where the joint step's scene scale is ill-conditioned: small input changes move $s$ between 1.07 and 1.37 (within 0.02 on the other 11 recordings), and this recording sets our rigid mean apart from CRISP's (Tab.~\ref{tab:scene}).
\item \textbf{Tracked in sim}: no fall and a median link error below 0.5\,m, with VideoMimic's terrain policy in our MuJoCo port of its harness; our meshes are the scene pipeline's terrain after the G1 dataset's terrain cleaning (heightfield spikes removed, floor bumps flattened); the G1's Model 15 joint limits as hard contacts (Sec.~\ref{sec:suppsettings}), no observation noise; every row on the same clips.
\end{itemize}

\begin{table*}[t]
\centering\tabfont
\caption{Per-clip scene geometry (per-frame Chamfer, cm; rigid and Sim(3)) on RICH-3 and EgoBody's egocentric recordings.}
\label{tab:scenegeo}
\setlength{\tabcolsep}{2.5pt}
\begin{tabular}{l cc cc cc cc cc cc}
\toprule
& \multicolumn{6}{c}{RICH-3} & \multicolumn{6}{c}{EgoBody ego} \\
\cmidrule(lr){2-7}\cmidrule(lr){8-13}
& \multicolumn{2}{c}{Gym} & \multicolumn{2}{c}{LectureHall} & \multicolumn{2}{c}{ParkingLot2} & \multicolumn{2}{c}{S21} & \multicolumn{2}{c}{S23} & \multicolumn{2}{c}{S03} \\
\cmidrule(lr){2-3}\cmidrule(lr){4-5}\cmidrule(lr){6-7}\cmidrule(lr){8-9}\cmidrule(lr){10-11}\cmidrule(lr){12-13}
Method & rigid & Sim(3) & rigid & Sim(3) & rigid & Sim(3) & rigid & Sim(3) & rigid & Sim(3) & rigid & Sim(3) \\
\midrule
VideoMimic depth & 26.7 & 7.3 & 15.7 & 16.2 & 12.9 & 11.6 & \textbf{6.1} & \textbf{4.0} & 11.8 & 7.9 & \textbf{14.8} & 10.9 \\
VideoMimic mesh & 16.6 & 15.8 & 32.5 & 32.7 & 30.3 & 20.7 & 45.0 & 23.3 & 18.3 & 16.9 & 19.4 & 18.4 \\
ViPE depth & 11.5 & 11.0 & 14.1 & 14.8 & 17.2 & 15.8 & 7.6 & 7.2 & 8.7 & 8.5 & 15.7 & 10.1 \\
\rowcolor{oursbg}
ours, TSDF mesh & \textbf{7.4} & \textbf{6.7} & \textbf{6.9} & \textbf{6.3} & \textbf{9.7} & \textbf{9.7} & 11.2 & 10.7 & \textbf{7.8} & \textbf{7.7} & 18.5 & \textbf{9.8} \\
\bottomrule
\end{tabular}
\end{table*}

\begin{table*}[t]
\centering\tabfont
\caption{Body to scan per RICH-3 clip: penetration (pen.), float and $|s|$ (cm), and the median predicted / true body depth (ratio). ViPE's focal is 31\% long on Gym; the static-camera focal rule (Sec.~\ref{sec:suppmethod}) brings it to $-6$\%.}
\label{tab:t1}
\setlength{\tabcolsep}{2.5pt}
\begin{tabular}{l ccc ccc ccc ccc}
\toprule
& \multicolumn{3}{c}{Gym} & \multicolumn{3}{c}{LectureHall} & \multicolumn{3}{c}{ParkingLot2} & \multicolumn{3}{c}{body depth} \\
\cmidrule(lr){2-4}\cmidrule(lr){5-7}\cmidrule(lr){8-10}\cmidrule(lr){11-13}
Method & pen. & float & $|s|$ & pen. & float & $|s|$ & pen. & float & $|s|$ & Gym & Lect. & Park. \\
\midrule
VideoMimic & 23.3 & 0.3 & 23.6 & 6.4 & 22.7 & 29.1 & 6.6 & 6.8 & 13.3 & 1.28 & 0.89 & 1.12 \\
\rowcolor{oursbg}
ours, ViPE's focal & 27.0 & 0.5 & 27.5 & \textbf{1.6} & \textbf{7.2} & \textbf{8.8} & \textbf{1.8} & \textbf{3.4} & \textbf{5.2} & 1.44 & 0.97 & 1.03 \\
\rowcolor{oursbg}
ours, rule's focal & \textbf{8.7} & \textbf{0.1} & \textbf{8.8} & 2.1 & 8.6 & 10.7 & \textbf{1.8} & \textbf{3.4} & \textbf{5.2} & 1.04 & 0.96 & 1.03 \\
\bottomrule
\end{tabular}
\end{table*}

\begin{table*}[t]
\centering\tabfont
\caption{Matched focal lengths on the RICH-3 clips (means, all in cm; K fixed: Sec.~\ref{sec:suppscenes}): per-frame Chamfer (rigid, Sim(3)), whole-scene accuracy and completeness after ICP, feet to scan, and body to scan (penetration, float, $|s|$).}
\label{tab:focalmatch}
\setlength{\tabcolsep}{2.5pt}
\begin{tabular}{l cc cc c ccc}
\toprule
& \multicolumn{2}{c}{per frame} & \multicolumn{2}{c}{whole scene} & feet & \multicolumn{3}{c}{body} \\
\cmidrule(lr){2-3}\cmidrule(lr){4-5}\cmidrule(lr){6-6}\cmidrule(lr){7-9}
Method, focal & rigid & Sim(3) & acc. & comp. & to scan & pen. & float & $|s|$ \\
\midrule
VideoMimic, K fixed to our rule's focal & 15.0 & 11.1 & 15.0 & 60.0 & 4.17 & 6.8 & 3.0 & 9.8 \\
VideoMimic, K fixed to the ground-truth focal & 15.9 & 10.6 & 15.1 & 58.0 & \textbf{4.15} & 9.5 & 3.5 & 13.0 \\
CRISP, K fixed to the ground-truth focal & \textbf{7.1} & \textbf{6.2} & \textbf{5.1} & 34.8 & 5.78 & 6.2 & \textbf{2.1} & 8.4 \\
JOSH, K fixed to the ground-truth focal & 16.5 & 9.5 & 98.8 & \textbf{3.8} & 16.08 & 5.8 & 3.4 & 9.2 \\
\rowcolor{oursbg}
ours, ViPE's focal & 10.4 & 7.6 & 17.9 & 26.2 & 4.44 & 7.3 & 4.7 & 12.0 \\
\rowcolor{oursbg}
ours, ground-truth focal & 8.1 & 7.2 & 17.9 & 26.8 & 5.14 & \textbf{3.7} & 3.7 & \textbf{7.4} \\
\bottomrule
\end{tabular}
\end{table*}

\begin{table}[t]
\centering\tabfont
\caption{VideoMimic's terrain policy on the 118 registered clips by start state: tracked in sim (\%) without observation noise and with noise (seeds 1 and 2); both methods' references on VideoMimic's meshes, with the same policy, plant and terrain, all on one machine type. Port protocol: each reference reset into its own first frame, as in our port of VideoMimic's harness. Default pose: reset into the policy's default joint pose, heading-only root, zero velocity, the lower ankle kept at its height.}
\label{tab:vmreset}
\setlength{\tabcolsep}{3pt}
\begin{tabular}{l ccc >{\columncolor{oursbg}}c>{\columncolor{oursbg}}c>{\columncolor{oursbg}}c}
\toprule
& \multicolumn{3}{c}{VideoMimic} & \multicolumn{3}{c}{ours} \\
\cmidrule(lr){2-4}\cmidrule(lr){5-7}
Start & no noise & seed 1 & seed 2 & no noise & seed 1 & seed 2 \\
\midrule
port protocol & 73.7 & 73.7 & 74.6 & 61.0 & 57.6 & 59.3 \\
default pose & 74.6 & 75.4 & 74.6 & 61.0 & 62.7 & 65.3 \\
\bottomrule
\end{tabular}
\end{table}

\paragraph{Findings.}
\begin{itemize}
\item \textbf{Whole scene} (rigid ICP): compared with VideoMimic, we lose accuracy on LectureHall to a depth-scale error and win completeness, as VideoMimic keeps only a ring around the person; CRISP is the most accurate and JOSH the most complete but by far the least accurate (accuracy / completeness, cm: VideoMimic 16.2 / 59.8, CRISP 5.5 / 32.9, JOSH 102.6 / 15.3, ours 16.9 / 25.7). Held at the true focal, CRISP's bodies stop floating above the scan and JOSH's scene and feet improve most (Tab.~\ref{tab:focalmatch}).
\item \textbf{SLOPER4D}: our scene's accuracy (65.8\,cm) is about VideoMimic's, but its completeness (123.0\,cm) is much worse; Sim(3) alignment does not help, and the camera drifts (ATE 77.6\,cm Sim(3), 96.9\,cm rigid over the walk). On the 16 chunks VideoMimic's output covers, our WA-MPJPE and W-MPJPE (9.0 and 26.0\,cm) are well below VideoMimic's. W-MPJPE is dominated by two of the 23 100-frame chunks the protocol scores (2{,}300 of the 2{,}400 frames), each aligned on its first two frames: chunk 1 (1{,}903\,mm) and chunk 5 (2{,}459\,mm). In both, our box is off the subject at the chunk's start (frames 0--9 and 400--428 of the sequence), so the two-frame alignment is fitted on the wrong body and the error grows over the chunk (in chunk 5 over a 4.9\,m walk); aligned on the whole chunk, their WA-MPJPE is 299 and 495\,mm, and over the other 21 chunks W-MPJPE is 239.1\,mm.
\item \textbf{G1 references on VideoMimic's 123 clips}: compared with VideoMimic's references, ours penetrate their terrain far less (penetration p95 4.1\,mm) but float on more frames (11.4\%), about half of it on the 28 clips with low supports (26.8\%), where the stance pin lifts the seated G1 off the seat, and jitter more (2.1\,rad\,s$^{-2}$). On all 123 clips, our references on our meshes track on 59.3\% (one clip without observed terrain cells keeps the solve on the level-snapped heightfield), fewer than VideoMimic's.
\item \textbf{Start pose} (Tab.~\ref{tab:vmreset}). Most of our falls on VideoMimic's meshes, like VideoMimic's own, are height falls after the first second.
\item \textbf{Feet on our own scene.} With the joint optimization, the share of contact frames with the feet within 3\,cm of our own scene rises from 3 to 74\% on RICH-3 and from 14 to 39\% on EMDB-2. The joint step fits the scene to these contacts, so this checks consistency, not accuracy; the scan-based feet-to-scan column of Tab.~\ref{tab:scene} measures accuracy.
\item \textbf{Visibility QA} (visible-body pixels covered by scene): 5.8 $\to$ 1.0\% on EMDB-2 and 1.6 $\to$ 0.23\% on RICH with carving.
\end{itemize}

\section{Limitations}
\label{sec:supplimits}
\begin{itemize}
\item \textbf{Inverted and seated bodies.} On EMDB-1, \paper-L's PA-MPJPE gap to SAM 3D Body (Tab.~\ref{tab:emdb1}) is largest on the sitting sequence (+7.6\,mm) and the handstand (+3.9\,mm), and 1.0--3.4\,mm on the climbing, soccer, dancing, cartwheel, skateboard, rock-chair and two workout sequences.
\item \textbf{Dance.} On AIST++ dance clips, \paper-L's MPJPE is 64.4\,mm, well above NLF-L's, though below that of \paper-H's MHR mesh $\to$ SMPL (MHR's official converter) and of SAM 3D Body.
\item \textbf{Foot pitch.} Out-of-plane foot pitch is the weakest foot measure: on MOYO the fitted foot's heel$\to$toe error is larger in pitch (6.9$^\circ$) than in the image plane (6.1$^\circ$), with a mean bias of $+1.8^\circ$.
\item \textbf{Contacts.} Knee contact over-fires when the knees hover just above the floor (plank-like floor exercises), where measured knee contact is rare, and static contact on a moving support (the skateboard) is the hardest case for the video model.
\item \textbf{Overlapping people.} Identity swaps between closely overlapping people remain (Harmony4D), and the world pipeline follows the tallest person, which switches on two 3DPW G1 clips (runForBus frames 381--599, crossStreets 0--53).
\item \textbf{Intrinsics.} For static cameras the focal is a single-image estimate. In our robustness set the rule fixes EgoBody's exocentric cameras (ViPE $+238$ / $+111$\% $\to$ $+32$ / $+0.2$\%; the third stays at $+13.8$\%), but MOYO's focal stays $+38$ / $+57$\% too long.
\item \textbf{Scene scale.} On EMDB P3\_28 the joint optimization leaves the camera path 4.8\% too large (Sim(3) scale against the ground-truth path), where it was 2.3\% too small before. Such a rescaling can push the shins below step edges.
\item \textbf{Scene height.} On EMDB-2, registered per 10\,s window to the ground-truth cameras, the reconstructed floor lies below the ground-truth body's planted feet, in part because the body has no shoes. On EMDB P3\_30 the MVS depth dips at the stair turn (21--24\,s).
\item \textbf{Stairs.} Near the walking path of EMDB P3\_30 and P3\_29 the heightfield's treads are partly ramps, from the per-frame depth at step edges.
\item \textbf{G1 references.} They are checked kinematically and in simulation (Sec.~\ref{sec:supphumanoid}); dynamic feasibility beyond VideoMimic's released terrain-aware policy~\cite{allshire2025videomimic} and contacts other than the feet are not checked. On video inputs the retargeted feet are less faithful in pitch (heel$\to$toe error 8.3$^\circ$ on 27 clips with \paper's bodies), and on seated motions the G1 pelvis is often off the person's (by more than 10\,cm on 142 of the G1 dataset's 312 seated windows). No tracker uses hand supports: on the incline push-up, where the hands rest on a bike rack, our references fall with both SONIC trackers (Tab.~\ref{tab:retargeters}). 25 of the G1 dataset's 1{,}107 scanned-scene windows keep more than 2$^\circ$ of residual floor tilt.\ifdense{} The stance soles' toe-down pitch (Tab.~\ref{tab:retmodes}) comes mainly from the two rest poses: SMPL's ankle-to-toe bone sits 19--24$^\circ$ below its sole and the G1's 12$^\circ$ below its own, and the retargeter matches the bone directions.\fi
\ifdense\else
\item \textbf{World.} On RICH-191 our detector and subject selection lose the subject in 23 of 191 videos, and there the world root follows another person (Tab.~\ref{tab:richsplit}).
\fi
\item \textbf{2D.} \paper-L's projected mesh is 7.8 COCO body-keypoint AP below SAM 3D Body's.
\item \textbf{Humanoid.} Acrobatic and hand-supported motions are not tracked by either SONIC tracker. \ifdense With SONIC-G1, the robot follows our references over only 47 / 48 / 48\,\% (mesh / points / skeleton) of their root path on the 13 clips that travel more than 3\,m, the least of every row (others 63--75\,\%; SONIC-SMPL 70 / 70 / 70\,\%), as the retargeters' input smoothing slows the swing foot; the points mode is also slightly less smooth on ground-truth bodies (Sec.~\ref{sec:supphumanoid}). On stance frames the retargeted G1's soles pitch toe-down against the person's (a median of 12.5 / 13.0$^\circ$ in mesh / points mode, Tab.~\ref{tab:retmodes}), mainly from the two feet's different rest-pose angles (Sec.~\ref{sec:supplimits}).\else With SONIC-G1, the robot follows our motion-level references over only 47\,\% of their root path on the 13 clips that travel more than 3\,m, the least of every row (63--75\,\% for the others; 70\,\% with SONIC-SMPL), as the retargeter's input smoothing slows the swing foot.\fi
\end{itemize}

\section{Implementation settings}
\label{sec:suppsettings}
Hyperparameters of the stated methods, for reproduction (the reported labels and metrics are defined in the text):

{\footnotesize
\begin{itemize}\setlength{\itemsep}{0pt}\setlength{\parskip}{0pt}
\item \textbf{Video crops}: previous-frame keypoint box enlarged by 1.1; subject kept at IoU above 0.1 with the previous subject box.
\item \textbf{Retargeter}: Adam, 1{,}600 iterations, step 0.02 (2\,mm for the root), cosine decay; terrain gradients from the heightfield blurred over 3.3\,cm.
\item \textbf{Stance pin}: contact runs of at least 0.2\,s, gaps below 0.1\,s filled; at most 8\,cm drift and 15$^\circ$ turn; 600 iterations (1{,}200 for clips still more than 3\,mm off); 1\,cm lift; root step capped at 4.5\,cm per frame.
\item \textbf{MVS}: every 3rd frame a keyframe at 960\,px; sources $\pm$1, 2, 4 and 8 keyframes away; depths 0.3--20\,m; a depth kept if at least two of its $\pm$1 / $\pm$2 neighbours agree within 2\% with a baseline of at least 3\% of the depth (about 1.7$^\circ$ triangulation angle); ViPE's depth scaled to MVS when MVS covers at least 10\% of a frame.
\item \textbf{Fusion}: 1.5\,cm voxels; a point seen in at least 3 frames; free-space carving when the camera sees past a point in more than 8\% of its depth, on every 2nd frame.
\item \textbf{Texture}: depth agreement within $\max(3\,\mathrm{cm}, 1\%)$; views at least 6$^\circ$ off grazing; blend weight $(\cos/\mathrm{depth})^8$; $4096^2$ atlas.
\item \textbf{Human--scene step}: contact on frames of contact probability $\geq 0.5$; $\log s$ prior width from 0.02 to 0.5 over 150 in-view foot-contact frames; coarse $\log s$ search at 81 points in $\pm1.2$ (truncated L1 on penetration, contact, depth consistency and the prior); Geman--McClure $\sigma=5$\,cm.
\item \textbf{JOSH}: RICH frames cropped by 13\,px top and bottom.
\item \textbf{VideoMimic harness}: joint-limit solref 0.01, solimp 0.99 / 0.999.
\end{itemize}}

\end{document}